\documentclass[11pt]{article}

\usepackage{todonotes}
\usepackage[preprint]{acl}

\usepackage{times}
\usepackage{latexsym}

\usepackage[T1]{fontenc}

\usepackage[utf8]{inputenc}

\usepackage{microtype}

\usepackage{inconsolata}

\usepackage{graphicx}
\usepackage{soul}
\usepackage{amssymb}
\usepackage{forest}

\title{A Systematic Review of NLP for Ghanaian Languages: Datasets, Models, and a Research Roadmap}

\author{
\textbf{Sheriff Issaka\textsuperscript{1,2}},
\textbf{Erick Rosas Gonzalez\textsuperscript{2}},
\textbf{Colene Agbo\textsuperscript{2}},
\textbf{Evans Kofi Agyei\textsuperscript{1,3}},
\textbf{Shruti Tyagi\textsuperscript{2}},\\
\textbf{John Emeka Eze\textsuperscript{1}},
\textbf{Enock Appiah Tieku\textsuperscript{4}}, 
\textbf{Junlin Fang\textsuperscript{5}},
\textbf{Thanh Do Nguyen\textsuperscript{5}},
\textbf{Juliet Arthur\textsuperscript{6}}, \\
\textbf{Zhaoyi Zhang\textsuperscript{7}},
\textbf{Mihir Heda\textsuperscript{7}},
\textbf{Keyi Wang\textsuperscript{8}},
\textbf{Yinka Ajibola\textsuperscript{7}},
\textbf{Rebecca Akpanglo-Nartey\textsuperscript{9}}, \\
\textbf{Frank Lawrence Nii Adoquaye Acquaye\textsuperscript{10}}, 
\textbf{Dennis Owusu\textsuperscript{10}},
\textbf{Jerry John Kponyo\textsuperscript{6}}, \\
\textbf{Stephen Moore\textsuperscript{3}},
\textbf{Isaac Wiafe\textsuperscript{6}},
\textbf{Sean Du\textsuperscript{5}}\\[0.3em]
\textsuperscript{1}African Languages Lab
\textsuperscript{2}University of California, Los Angeles
\textsuperscript{3}University of Cape Coast\\
\textsuperscript{4}Max Planck Institute for Evolutionary Anthropology
\textsuperscript{5}Nanyang Technological University\\
\textsuperscript{6}Kwame Nkrumah University of Science and Technology
\textsuperscript{7}University of Wisconsin-Madison\\ 
\textsuperscript{8}Georgia Institute of Technology 
\textsuperscript{9}University of Education, Winneba
\textsuperscript{10}Ashesi University\\[0.4em]
{\small
\textbf{Correspondence:} 
\href{mailto:sheriff@cs.ucla.edu}{sheriff@cs.ucla.edu}
}
}

\begin{document}
\maketitle
\begin{abstract}
Natural Language Processing (NLP) for Ghana's 73 living indigenous languages remains deeply fragmented, under-resourced, and heavily skewed toward a single language. We present the first systematic review of the Ghanaian NLP landscape, screening 17,000+ publications across four academic databases to critically synthesize 36 core studies spanning datasets, model architectures, and evaluation paradigms. Our analysis exposes a severe resource imbalance: Twi-centric NLP has grown modestly, driven largely by religious-text alignment and crowdsourcing, while the remaining 70+ languages remain almost entirely unaddressed, and dataset releases, model checkpoints, and evaluation practices remain inconsistent and rarely shared across the field. We translate these findings into a prioritized roadmap targeting Ghana's acute regional constraints, dialectal variation, non-standardized orthographies, and the absence of shared infrastructure, offering a replicable template for systematic review and research prioritization in other low-resource language settings. \footnote{To promote accessibility, we provide translations of this abstract in 15 languages in Appendix~\ref{app:translations}, using Mansa}
\end{abstract}

\section{Introduction}
\label{sec:intro}
Language is the cornerstone of human identity, culture, and knowledge transmission, making its digital exclusion a global crisis rather than a regional one. Of the roughly 7,000 languages spoken worldwide, only a small fraction are supported by modern Natural Language Processing (NLP) systems, leaving the overwhelming majority of global linguistic diversity out of the generative AI revolution. For the more than 2,000 languages spoken across Africa, representation in NLP remains alarmingly sparse \cite{childs_language_2020, adebara2022afrocentricnlpafricanlanguages, alabi2025chartinglandscapeafricannlp, qin2025survey, issaka2026proxy}.

This disparity is not merely an academic concern: it carries profound societal consequences, accelerating the erosion of languages, cultures, and the communities that carry them \cite{batibo_language_2005, matsinhe01012006, lupke2009margin, unesco_atlas_2010, lupke2019language, essegbey2020language, childs_language_2020}. Crucially, the languages selected for NLP benchmarking and tool development correlate far more strongly with the economic and political power of their speaker communities rather than speaker population, demonstrating that technological neglect tracks historical marginalization rather than linguistic scale \cite{blasi-etal-2022-systematic}.

Ghana illustrates this crisis in sharp relief. Of its 84 documented languages, one has already gone extinct, and eight others are classified as endangered or dying \cite{eberhard_ethnologue_2024}. Ghana's 73 living indigenous languages (including Asante Twi, Ewe, Fante, Ga, and Dagbane) are spoken by millions, yet remain vastly absent from the NLP literature. This neglect is compounded by structural forces: colonial language policies that elevated English at the expense of indigenous tongues, government-sponsored instruction limited to only 11 of those languages, and a media ecosystem in which Akan-Twi dominates over 90\% of non-English programming \cite{akpanglo-nartey_endangered_2012}. The result is a self-reinforcing cycle of under-resourcing: without data, models cannot be built; without models, communities cannot benefit from language technology; and without technology, the documentation and revitalization of these languages stalls. Ghana has an active local research community, government-recognized languages, and growing institutional interest, which makes its persistent underrepresentation all the more diagnostic of the structural barriers facing minority languages worldwide.

Recent progress in NLP offers a realistic opportunity to interrupt this cycle. Massively multilingual machine-translation (MT) resources such as MADLAD-400 \cite{kudugunta2023madlad400}, together with frontier LLMs such as GPT-5.6 \cite{openai2026gpt56}, Opus 5 \cite{anthropic2026opus5}, and Gemini 3.1 Pro \cite{google2026gemini31pro} and open multilingual models such as Aya Expanse \cite{dang2024ayaexpanse} and Tiny Aya \cite{salamanca2026tinyayabridgingscale}, have substantially expanded cross-lingual generation and transfer. More importantly for African and other low-resource languages (LRLs), recent models such as InkubaLM \cite{tonja2024inkubalm}, Serengeti \cite{adebara2023serengetimassivelymultilinguallanguage}, Cheetah \cite{adebara2024cheetahnaturallanguagegeneration}, Lugha-Llama \cite{buzaaba2025lughallamaadaptinglargelanguage}, EMMA-500 \cite{ji-etal-2026-data}, N-ATLaS-LLM \cite{awagptv1_2025}, and AfriqueLLM \cite{yu-etal-2026-afriquellm} demonstrate that continued pre-training, multilingual instruction tuning \cite{devlin_bert_2019, raffel_exploring_2020}, synthetic data, and carefully designed data mixtures can adapt existing foundation models without requiring a separate system to be trained from scratch for every language. These advances make meaningful LRL research increasingly feasible with smaller labeled datasets and more modest computational resources. Nevertheless, recent evaluations such as AfroBench \cite{ojo-etal-2025-afrobench} show that substantial performance gaps between English and many African languages remain, underscoring the continuing need for better data, broader evaluation, and community-grounded model development \cite{issaka-etal-2026-african}. 

Yet despite recent progress, the literature still lacks an up-to-date, granular account of Ghanaian language NLP. Broad surveys of low-resource and African NLP \cite{magueresse_low-resource_2020,haddow_survey_2022,ranathunga_neural_2023,mabokela_multilingual_2023,alabi-etal-2025-charting,hussen2026state} provide valuable continental perspectives but obscure differences among individual Ghanaian languages in resources, tasks, evaluation, and technological readiness. Without such synthesis, research may be duplicated, evaluation practices may remain inconsistent, and urgent language-specific needs may be overlooked.

 We present the first large-scale, region-specific systematic review of NLP research for Ghanaian languages. Our contributions are fourfold:

\begin{enumerate}
    \item \textbf{Comprehensive Survey:} We provide the first exhaustive overview of NLP research conducted on Ghanaian languages, synthesizing findings from across four major databases.
    \item \textbf{Critical Analysis:} We rigorously compare the models, datasets, evaluation metrics, and methodologies employed in prior work, identifying strengths, limitations, and inconsistencies in the existing literature.
    \item \textbf{Actionable Roadmap:} We articulate concrete challenges and prioritized recommendations to guide future research toward maximum impact, including closer integration between linguistics and computation researcher.
    \item \textbf{Methodological Template:} By addressing specific regional challenges, such as Akan dialectical variations and non-standardized orthographies, we offer a replicable framework for LRL research globally. 
\end{enumerate}


\begin{figure}
    \centering
    \includegraphics[width=1\linewidth]{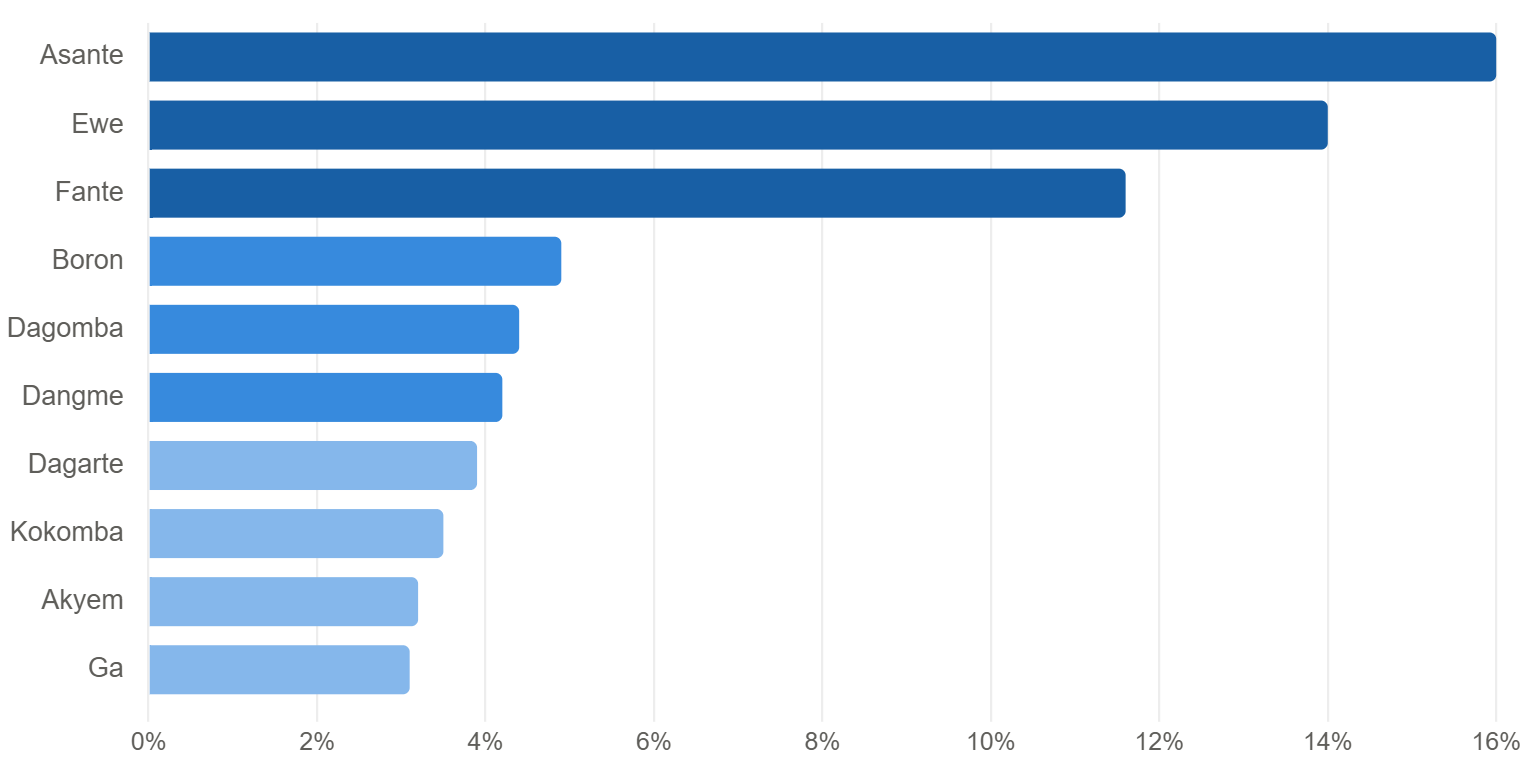}
    \caption{Percentage of speakers for top indigenous Ghanaian languages (Other = 31.2\%).}
    \label{fig:gh_lang_speakers}
\end{figure}

\section{Related Work}
Early work documented NLP initiatives and resources for Ghanaian languages \cite{azunre_nlp_2021}, while other surveys examined West African or LRL NLP more broadly \cite{etim_state---art_2021,magueresse_low-resource_2020,haddow_survey_2022,mabokela_multilingual_2023,ranathunga_neural_2023}. More recent studies provide continental analyses of African NLP and evaluate LLMs across African languages \cite{adebara2025weevaluatingllmperformance,alabi-etal-2025-charting,hussen2026state}, but their broad scope limits country- and language-specific analysis. 

While commendable, none of these works have provided the depth and specificity required to understand the unique characteristics and challenges of Ghanaian languages, a gap this survey is designed to address. 


\begin{figure*}[t]
    \centering
    \includegraphics[width=\textwidth]{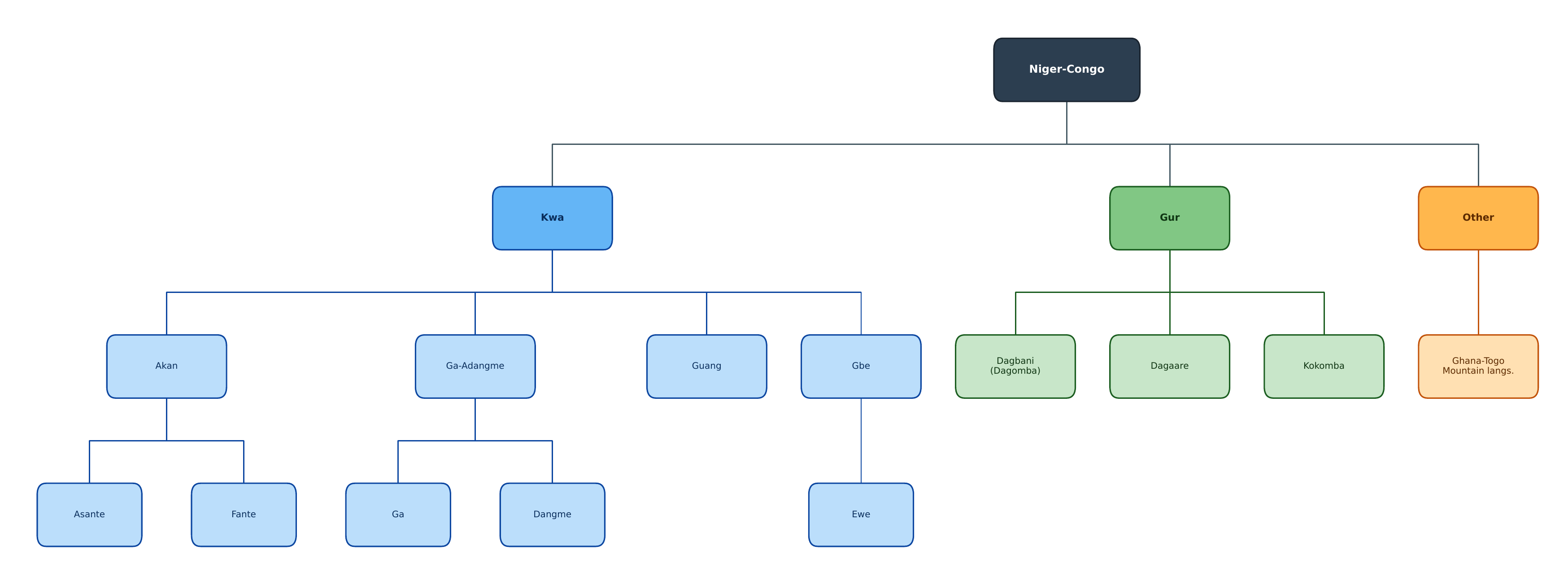}
    \caption{Simplified language family tree for major Ghanaian languages discussed in this survey, showing their placement within the Niger-Congo phylum.}
    \label{fig:lang-family-tree}
\end{figure*}

\section{{Ghanaian Languages}}
Ghana, a West African nation with a population exceeding 35 million, boasts a rich linguistic landscape of 73 living indigenous languages alongside English, its official language, a legacy of British colonial rule \cite{eberhard_ethnologue_2024}. Among these, Asante, Ewe, Fante, Boron, Dagomba, Dangme, Dagaare, Kokomba, Akyem, and Ga are the most widely spoken (Figure \ref{fig:gh_lang_speakers} and Table \ref{tab:percentage-top-speakers}) \cite{ghana_statistical_service_ghana_2023}. Most belong to the Niger-Congo phylum: Akan (Asante and Fante), Ga, Dangme, and Guang fall within the Kwa branch (with Ga-Dangme forming a subgroup and Ewe belonging to the Gbe cluster), while Dagomba (Dagbani), Dagaare, and Kokomba belong to the Gur branch; a smaller number, including some Ghana-Togo Mountain languages, sit outside these major groupings. Typological relatedness within the Kwa and Gur branches directly informs which cross-lingual transfer strategies are likely to succeed (Figure \ref{fig:lang-family-tree}).

Akan is the most widely spoken indigenous language, native to approximately 44\% of the population \cite{osam_introduction_2003}. It comprises two main dialect groups: Fante (including sub-dialects such as Gomua, Ekumfi, Nkusukum, Iguae, Breman, and Agona) and Twi (consisting of Akuapem and Asante). Predominantly spoken in southern Ghana, both Twi dialects are mutually intelligible, used in education, and share identical morphological structures and vowel harmony rules. While Asante Twi enjoys broader daily usage, Akuapem Twi serves as the historical written literary standard \cite{schachter_phonology_1968}.

Government policies mandate using a child's mother tongue (L1) as the medium of instruction from kindergarten through Grade 3 before transitioning to English \cite{usaid_ghana_2020}. However, this policy is restricted to 11 sponsored languages (Akuapem Twi, Asante Twi, Fante, Nzema, Dagaare, Dagbane, Ewe, Dangme, Ga, Gonja, and Kasem) and suffers from low implementation due to shortages of qualified teachers and instructional materials \cite{awedoba_attitudes_2009, adika_english_2012, davis_language_2012, akpanglo-nartey_endangered_2012, owu-ewie_language_2017}. These institutional limitations, combined with Akan’s media dominance (where Twi accounts for 90\% of non-English television programming), have exacerbated the marginalization and potential endangerment of non-sponsored indigenous languages \cite{akpanglo-nartey_endangered_2012}. Coupled with a declining population of fluent elderly speakers, these dynamics have driven researchers to explore technological and NLP-driven solutions to preserve and revitalize Ghana's indigenous linguistic heritage \cite{galla_indigenous_2016}.

\section{{NLP and MT in LRLs Revitalization}}
\label{sec:nlp-and-mt-in-lrls-revitalization}

NLP and MT have emerged as the primary technological approaches for revitalizing African languages. NLP, an interdisciplinary field that combines Linguistics, Computer Science, and AI, enables computers to understand and process human language. MT, a subfield of NLP, focuses on translating text or speech from one language to another. The development of MT systems has evolved through four main paradigms: Rule-Based MT (RBMT), Statistical MT (SMT), Hybrid MT (HMT), and Neural MT (NMT). Details on each paradigm are provided in Appendix~\ref{app:mt-paradigms}.

\subsection{Pre-training and Fine-tuning}

The development of powerful pre-trained language models, such as BERT \cite{devlin_bert_2019}, GPT \cite{radford_improving_2018}, T5 \cite{raffel_exploring_2020}, Serengeti \cite{adebara2023serengetimassivelymultilinguallanguage}, and Cheetah \cite{adebara2024cheetahnaturallanguagegeneration} has transformed the approach to NLP tasks. These models enable researchers to fine-tune pre-trained models for specific applications rather than building new models from scratch. Pre-training has significantly improved the accuracy of NLP models for various tasks, including language translation, sentiment analysis, and text classification \cite{min2023recent, nguefack-etal-2025-pretraining, hussen2026state}. This approach has also facilitated the development of NLP applications for LRLs, which often lack sufficient data for training models from scratch. As the technology continues to advance, pretrained and instruction-tuned multilingual models are expected to become even more powerful, enabling more accurate and nuanced language analysis, particularly for previously underrepresented languages \cite{qin2025survey}.

\subsection{State of NLP in Ghanaian Languages}
\begin{figure*}[t]
  \centering
  \begin{minipage}[t]{0.48\textwidth}
    \centering
    \includegraphics[width=\linewidth]{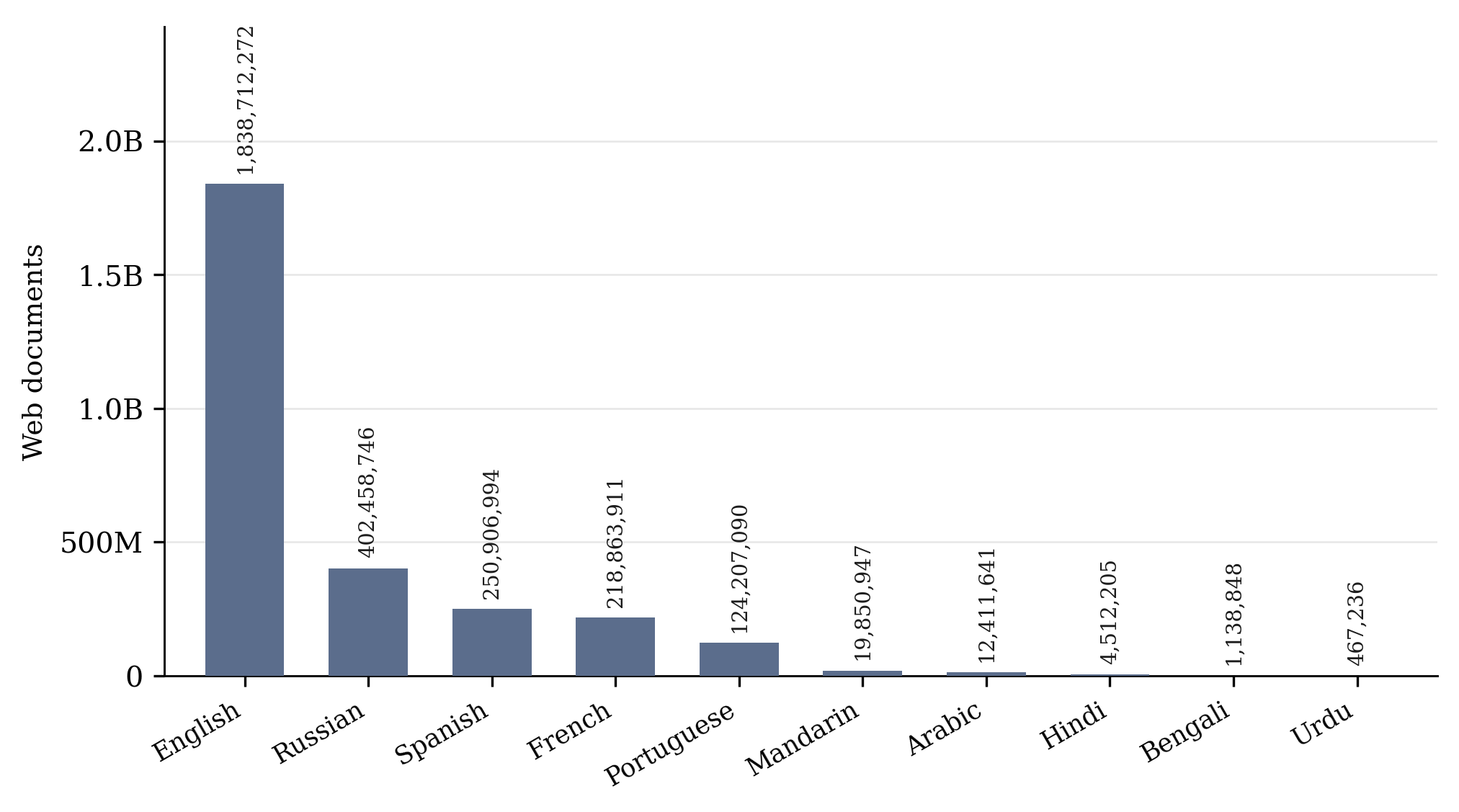}
    \caption{Document counts for world's top ten spoken
    languages in MADLAD-400 \cite{kudugunta2023madlad400}.}
    \label{fig:world-footprint}
  \end{minipage}\hfill
  \begin{minipage}[t]{0.48\textwidth}
    \centering
    \includegraphics[width=\linewidth]{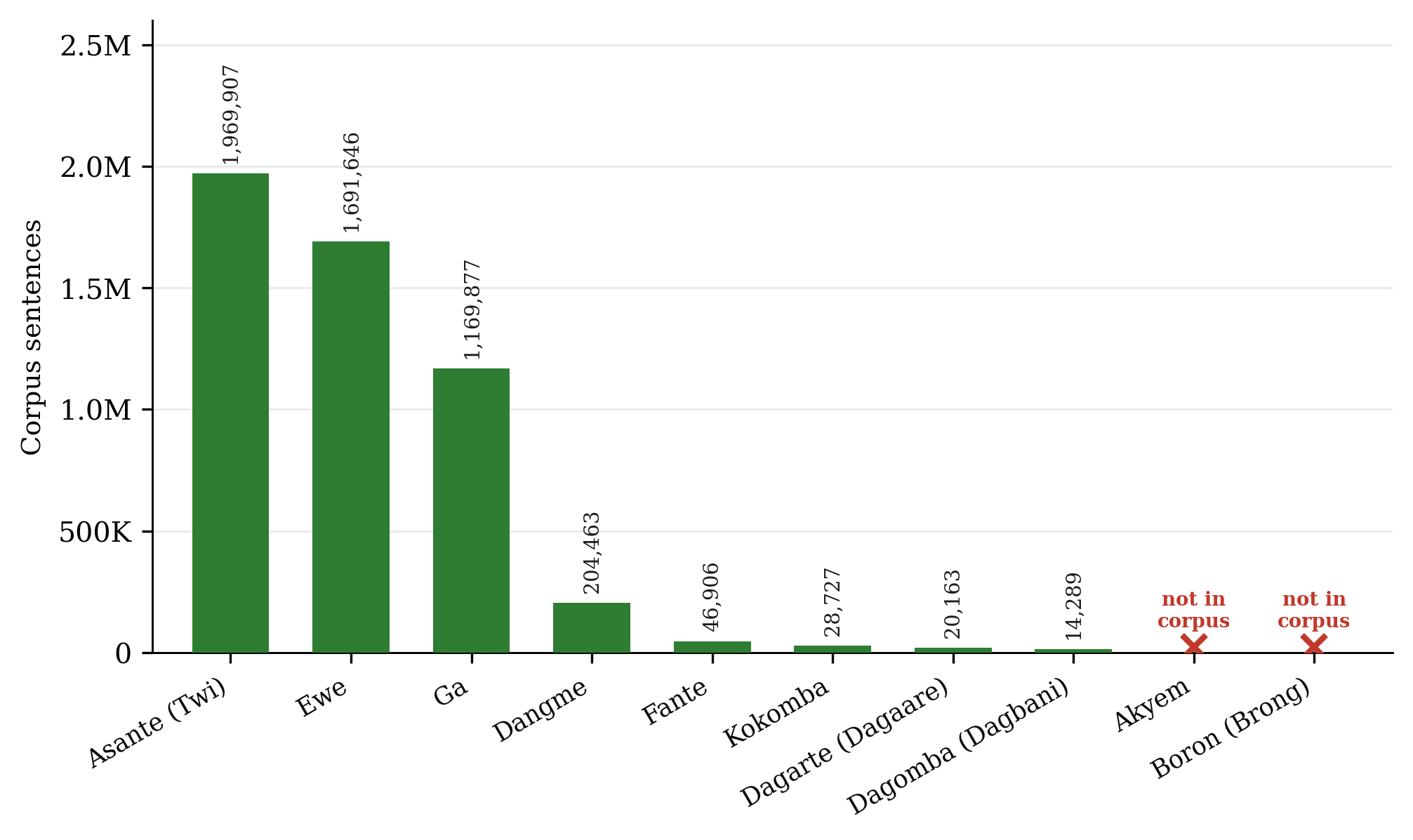}
    \caption{Sentence counts for Ghana's top ten spoken languages
    in the GlotLID corpus \cite{kargaran-etal-2023-glotlid}.}
    \label{fig:ghana-footprint}
  \end{minipage}
\end{figure*}

Like most African languages, all Ghanaian languages are considered LRLs: languages marked by limited publicly accessible corpora, annotated datasets, pretrained models, and NLP tools, minimal digitization, and a lack of privileged status in teaching and resource allocation \cite{magueresse_low-resource_2020, issaka2026proxy, afram_twieng_2022, issaka-etal-2026-african}. This disparity between the linguistic diversity of African languages and their inadequate NLP representation can be attributed to factors such as the opaqueness of available resources and the languages' own linguistic complexity \cite{orife_masakhane_2020}.


To situate Ghanaian languages within this broader digital landscape, Figures~\ref{fig:world-footprint} and~\ref{fig:ghana-footprint} contrast text volume for the world's ten most widely spoken languages against Ghana's ten most widely spoken languages. Because no single web-scale corpus reliably indexes most Ghanaian languages, the two draw on different sources and units: Figure~\ref{fig:world-footprint} reports audited web-document counts from MADLAD-400 \cite{kudugunta2023madlad400}, while Figure~\ref{fig:ghana-footprint} reports sentence counts from GlotLID \cite{kargaran-etal-2023-glotlid}, a corpus built specifically to extend LRL identification; we present them separately rather than normalize them onto a shared axis. The disparity is nonetheless stark: the largest Ghanaian entry, Asante Twi, is measured in the low millions of sentences, while English alone accounts for more than 1.8 billion documents.

Despite ongoing efforts to document and digitize Ghanaian languages, particularly Twi, most available datasets are either religious texts or extremely limited in size, a consequence of these languages being used primarily in colloquial, unofficial settings despite millions of native speakers. This low-resource status exacts high societal and technical costs. Beyond threatening digital language preservation and impeding diaspora education, it obstructs healthcare delivery in remote communities and limits fundamental NLP tasks, such as summarization and classification, vital for critical social and cybersecurity infrastructure \cite{azunre_english-twi_2021, bremang-etal-2026-creating}. The resulting scarcity of large, clean, reliable datasets further compounds the problem, creating a negative feedback loop that continues to slow NLP progress in the region \cite{siminyu_ai4d_2021, issaka2026proxy}. Despite these challenges, recent efforts have produced LLMs trained to support 517 African languages, including Twi \cite{adebara2023serengetimassivelymultilinguallanguage, adebara2024cheetahnaturallanguagegeneration}.

\section{Survey Methodology}

We conducted this review following PRISMA guidelines for systematic reviews, adapted to the reporting conventions of computational linguistics. The subsections below specify the eligibility criteria, information sources, search strategy, selection process, data items, and synthesis methods. Record counts are reported in Section \ref{sec:survery-results} and Table \ref{tab:search-queries}.

\subsection{Eligibility Criteria}
\label{sec:methodology-eligibility-criteria}
We included publications whose primary focus was the development, application, or evaluation of NLP or MT for one or more Ghanaian languages. Records were excluded if they addressed neither NLP nor MT for Ghanaian languages, or if they duplicated a record already retrieved under another query or database. Eligibility was assessed from each record's title, abstract, and keywords. We applied no restriction on publication venue or peer-review status; we record each source's peer-review status as a separate field in our synthesis (Table~\ref{tab:corepapers})

\subsection{Information Sources}
\label{sec:methodology-information-sources}
We searched four bibliographic databases: Scopus, IEEE Xplore, Google Scholar, and the Technology Collection. All searches were run in July 2026, and the counts reported here reflect the state of each database on that date. No additional records were identified through citation chasing, snowballing, or author contact; the review draws exclusively on these four databases.

\subsection{Search Strategy}
Searches used boolean-combined, exact-phrase queries built around the core concepts of this study (NLP, MT, Ghana, Twi, and Language). Each query was run separately against each database. The full set of queries, together with per-database record counts, is reported in Table~\ref{tab:search-queries}.

\subsection{Selection Process}
\label{sec:selection-process}
All screening was performed manually by the authors. Researchers screened the records returned by each query against the eligibility criteria in Section \ref{sec:methodology-eligibility-criteria}, examining each record's title, abstract, and keywords. We extracted the research objectives, the methodology employed, the datasets used, and their construction methods, removed duplicates along with records judged unrelated to the review question, and resolved disagreements over borderline cases by discussion. This process yielded the final set of core papers reported in Section \ref{sec:survery-results}.

\subsection{Data Collection Process and Data Items}
For each included paper, we extracted research objectives, methodology, datasets used and their construction methods, model architecture and training paradigm, evaluation metrics and reported scores, and principal findings. Extracted items are tabulated in Tables \ref{tab:summary-bleu-chrf-ter}, \ref{tab:summary-SpBLEU-rouge-wer-f1}, \ref{tab:modeling-taxonomy}, and \ref{tab:corepapers}. Further details on data synthesis and reproducibility are available in Appendix xxxxxxxxx.

\begin{table*}
  \centering
  \resizebox{\textwidth}{!}{%
  \begin{tabular}{llll}
    \hline
    \textbf{Database} & \textbf{Search Terms} & \textbf{Total Results} & \textbf{Core Papers Retrievable} \\
    \hline
    Scopus                & Machine Translation AND African Languages   & 104   & 3  \\
    Scopus                & Machine Translation AND Twi                 & 10    & 7  \\
    Scopus                & NLP AND Ghanaian Language                   & 2     & 2  \\
    Google Scholar        & Machine Translation AND Ghana Languages     & 14    & 4  \\
    Google Scholar        & Machine Translation AND Twi                 & 1,030 & 29 \\
    Google Scholar        & Machine Translation AND Ewe                 & 808   & 19 \\
    Google Scholar        & NLP AND Twi                                 & 2,040 & 32 \\
    Google Scholar        & LLM AND Twi                                 & 3,160 & 8  \\
    Google Scholar        & LLM AND Ewe                                 & 3,830 & 7  \\
    Google Scholar        & African Languages AND Large Language Model  & 985   & 1  \\
    Google Scholar        & African Languages AND NLP                   & 5,610 & 28 \\
    IEEE Xplore           & Machine Translation AND Twi                 & 3     & 3  \\
    IEEE Xplore           & Machine Translation AND Akan                & 1     & 1  \\
    IEEE Xplore           & NLP AND Twi                                 & 3     & 2  \\
    Technology Collection & Machine Translation AND Ghanaian Language   & 18    & 2  \\
    \hline
  \end{tabular}}
  \caption{Exact-phrase quoted search queries per database, total records returned, and the number of the 36 non-unique core papers (Table~\ref{tab:corepapers}) retrieved by each query.}
  \label{tab:search-queries}
\end{table*}

\section{{Survey Results}}
\label{sec:survery-results}

Our systematic search yielded 17,618 results across four databases as of July 2026 (Table~\ref{tab:search-queries}). Applying the screening and selection process described in Section \ref{sec:selection-process} produced a final set of 36 core papers focusing on NLP for Ghanaian languages (Table ~\ref{tab:corepapers}). All 36 core papers are retrievable via Google Scholar, 10 via Scopus, 4 via IEEE Xplore, and 2 via the Technology Collection, reflecting Google Scholar's indexing of arXiv preprints and regional venues absent from curated databases.

In the following sections, we present a summary of our findings, categorized by datasets, models, and performance. For each category, we discuss the various approaches employed by the reviewed papers, providing a comprehensive overview of the current state of NLP research in the context of Ghanaian languages.

\subsection{Data Collection Approaches}
\label{sec:survery-results-data-collections}

The reviewed papers employed diverse collection approaches to curate datasets for their NLP tasks. We categorize these approaches into religious texts, crowdsourcing, web scraping, and commissioned professional translation.

\subsubsection{Religious Texts}
Nine papers \cite{adjeisah_twi_2020, acheampong_language_2021, yvette_gelr_2021, afram_twieng_2022, agyei2024low, bannerman2024machine, adebara2023serengetimassivelymultilinguallanguage, adebara2024cheetahnaturallanguagegeneration, agyei_akan-english_2021} utilized parallel corpora derived from the English Bible and its corresponding translation. \citealp{acheampong_language_2021} extracted 20,000 raw Akan sentences from the Bible Society of Ghana's 2012 version of YouVersion, which were pre-processed and aligned with their English counterparts. This dataset was directly employed by \citealp{agyei_akan-english_2021} for training and evaluating their MT model. \citealp{adjeisah_twi_2020} adopted a more comprehensive approach, scraping multiple Bible versions in various languages, including English, French, Chinese, German, Spanish, Twi, Greek, and Hebrew, to construct a parallel corpus. In addition to using the Bible derived textual resources, the BibleTTS project demonstrates the use of audio recordings, which were aligned at the verse level to create a massive speech corpus for training high quality text-to-speech for ten languages \cite{meyer2022biblettslargehighfidelitymultilingual}

Two papers \cite{azunre_contextual_2021, hacheme_english2gbe_2021} leveraged the JW300 dataset, a parallel corpus spanning 300 languages with approximately 100,000 sentence pairs per language, derived from the Jehovah's Witness website. Despite its religious origin, this dataset comes primarily from \textit{Watchtower} and \textit{Awake!} magazines and it covers a range of domains \cite{agic_jw300_2019}.

\citealp{agyei2025dynamic} also built a significant portion of their corpus from Seventh-day Adventist Sabbath School quarterly books, which address a broader array of contemporary life topics.

\subsubsection{Crowdsourcing}
Nine papers employed crowdsourcing techniques \cite{azunre_english-twi_2021, acheampong_language_2021, azunre_nlp_2021, afram_twieng_2022, agyei2024low, bannerman2024machine, raychawdhary2025empowering, li2025ssacometllmsoutperformlearned,muhammad-etal-2023-semeval} for their datasets. \citealp{azunre_english-twi_2021, azunre_nlp_2021} and \citealp{afram_twieng_2022} utilized Google Forms to collect translations, while \citealp{acheampong_language_2021} supplemented their Bible-derived dataset with 6,000 sentences gathered by volunteers from social media platforms, including WhatsApp, WeChat, and Signal. \citealp{li2025ssacometllmsoutperformlearned}  makes use of a global news platform that publishes articles in multiple languages through volunteer translators. \cite{muhammad-etal-2023-semeval} collected tweets using location- based and vocabulary-based heuristics. \citealp{zhang2025sentiment} subsequently utilised this tweet dataset to develop and compare translation-based and end-to-end sentiment analysis systems for Twitter.

\subsubsection{Web Scraping}
Web scraping techniques were applied by fourteen  papers \cite{ogueji_pidginunmt_2019, yvette_gelr_2021, afram_twieng_2022, agyei2024low, bannerman2024machine, gyimah2025asantetwisenti,adebara2023serengetimassivelymultilinguallanguage, adebara2024cheetahnaturallanguagegeneration, li2025ssacometllmsoutperformlearned, song2026smalllanguagemodelsilver, muhammad-etal-2023-semeval, buzaaba2025lughallamaadaptinglargelanguage, adelani-etal-2022-masakhaner, dione2023masakhapospartofspeechtaggingtypologically} to construct datasets from online sources such as news articles, literary texts, and government documents. Researchers employed various web scraping tools, including Beautiful Soup, SpiderLing, and the Chrome extension Web Scraper.

The scope of web-based data collection has expanded to include diverse sources. \citealp{bannerman2024machine} incorporated sentences from the Tatoeba.org website, a public repository of bilingual sentence pairs. Moving beyond translation, \citealp{gyimah2025asantetwisenti} utilized the Twitter API to collect a large-scale dataset of Asante Twi tweets, specifically curated for sentiment analysis tasks and highlighting the application of web scraping for monolingual NLP applications. 

\subsubsection{Commissioned Professional Translation}
A fourth approach commissions trained translators directly rather than deriving text from existing sources or relying on volunteer effort. \citealp{gyamfi2026ghananlp} document 41,513 English-aligned sentence pairs across Twi, Fante, Ewe, Ga, and Kusaal, produced by professional translators and annotators. \citealp{adelani2025irokobenchnewbenchmarkafrican} follow a related model, engaging language coordinators to recruit professional translators for benchmark construction.

\subsection{Direct Field Recording}
\label{sec:results--direct-field-recording}
A small but important complement to the above comes from direct field recording of speech. \citealp{WiafeUGSpeechDataPaper} introduces UGSpeechData, a spoken corpus spanning five Ghanaian languages. The repository \cite{WiafeUGSpeechDataRepo} lists 970,148 audio files totaling 5,384 hours. Unlike the text-first corpora that dominate this survey, UGSpeechData supports speech development.

\begin{table*}
  \centering
  \resizebox{\textwidth}{!}{%
  \begin{tabular}{p{0.35\textwidth}lp{0.2\textwidth}lll}
    \hline
    \textbf{Papers} & \textbf{Architecture} & \textbf{Language Pair} & \textbf{BLEU} & \textbf{CHRF} & \textbf{TER} \\
    \hline
    {\cite{agyei_akan-english_2021}}               & Transformer          & Akan-English               & 0.1296          & -          & - \\
                                                   & Transformer          & English-Akan               & 0.1757          & -          & - \\
    \hline
    {\cite{hacheme_english2gbe_2021}}              & Transformer   & English-Ewe                & 0.357           & 0.549      & 0.511 \\
                                                   & Transformer   & English-Fon                & 0.42            & 0.541      & 0.484 \\
                                                   & Transformer   & English-Gbe                & 0.412           & 0.566      & 0.465 \\
    \hline
    {\cite{acheampong_language_2021}}              & Transformer          & Akan-English               & 0.1129          & -          & - \\
                                                   & Transformer          & English-Akan               & 0.1666          & -          & - \\
    \hline
    {\cite{azunre_english-twi_2021}}               & Transformer          & English-Twi                & 0.720           & -          & - \\
    \hline
    {\cite{adjeisah_pseudotext_2021}}              & Bidirectional-LSTM   & English-Twi                & 0.1963          & -          & 0.5031 \\
                                                   & Bidirectional-LSTM   & Twi-English                & 0.1948          & -          & 0.5068 \\
    \cline{2-6}
                                                   & Transformer          & English-Twi                & 0.1836          & -          & 0.5451 \\
                                                   & Transformer          & Twi-English                & 0.1813          & -          & 0.5476 \\
    \hline
    {\cite{gyasi_twi_2023}}                        & Transformer          & Twi-French                 & 0.36            & -          & - \\
                                                   & Transformer          & French-Twi                 & 0.40            & -          & - \\
    \hline
    {\cite{bannerman2024machine}}                  & Transformer          & Twi-English                & 0.8111          & -          & - \\
    \hline
    \cite{adebara2024cheetahnaturallanguagegeneration} & Transformer      & Multilingual English-Twi   & 0.0464 ± 0.0013 & 0.2589 ± 0.002 & - \\
    \hline
    {\cite{moore2026nsankuevaluatingzeroshottranslation}} & Transformer & 43 Language Mean & 0.2460 & 0.2916 & - \\
    \hline
  \end{tabular}}
  \caption{Summary of the BLEU, CHRF, and TER scores of reviewed models. All scores are normalized to 0-1 scale.}
  \label{tab:summary-bleu-chrf-ter}
\end{table*}

\subsection{Model Architectures and Fine-tuning}
\label{sec:survery-results--model-architectures}
Among the papers reviewed for model development, the overwhelming majority employed Transformer-based architectures or their variants, with Bidirectional LSTMs appearing as a secondary approach, and one paper \cite{adjeisah_pseudotext_2021} evaluating both.

\subsubsection{Tokenization and Preprocessing}
Five of the reviewed modeling papers \cite{hacheme_english2gbe_2021, acheampong_language_2021, azunre_english-twi_2021, gyasi_twi_2023, bannerman2024machine} employed Byte-Pair Encoding (BPE) tokenization techniques, while three others \cite{agyei_akan-english_2021, adjeisah_pseudotext_2021, adebara2023serengetimassivelymultilinguallanguage} tokenized sentences into words, while \cite{agyei2024low} utilized Sketch Engine's universal tokenizer. Additionally, \cite{li2025ssacometllmsoutperformlearned} used the NLTK sentence tokenizer to segment larger articles prior to word-level processing.

Recent work has placed a stronger emphasis on comprehensive data cleaning and preprocessing pipelines to improve model quality. Several studies \cite{agyei2024low, afram_twieng_2022, bannerman2024machine} utilized tools like the Sketch Engine and jusText for deduplication and the removal of boilerplate content such as HTML tags, advertisements, and social media handles. Common text standardization steps included converting text to lowercase and removing non-ASCII characters \cite{agyei2025enhancing, agyei2025dynamic}. For transformer-based approaches, subword tokenization remains prevalent. The T5 tokenizer was specifically employed in multiple studies \cite{agyei2025enhancing,agyei2025dynamic} with a maximum sequence length of 256 tokens. Sentence length caps, such as 512 tokens, have also been used \cite{song2026smalllanguagemodelsilver}. In sentiment analysis tasks, preprocessing also involved stripping noise like emojis, URLs, and user mentions from social media text \cite{raychawdhary2025empowering}. A theme across recent research is the critical role of human experts for translation verification, correction, and alignment to ensure the creation of high-quality, reliable parallel corpus \cite{agyei2024low, bannerman2024machine, adebara2023serengetimassivelymultilinguallanguage}.

\subsubsection{Model Configuration and Training}
\label{sec:model-config}
Among the reviewed modeling papers, a majority employed Transformer-based architectures. Recent work continues this trend, with a strong focus on fine-tuning large pre-trained models. The modeling approaches reviewed fall into four broad paradigms. \textbf{(i) Train-from-scratch architectures}, where Transformers or BiLSTMs are trained directly on task-specific parallel data without pretrained initialization \citep{agyei_akan-english_2021, hacheme_english2gbe_2021, acheampong_language_2021, adjeisah_pseudotext_2021}. \textbf{(ii) Fine-tuning of pretrained MT models}, where existing checkpoints such as OPUS-MT or T5-Small are adapted to Ghanaian language pairs \citep{azunre_english-twi_2021, gyasi_twi_2023, bannerman2024machine, agyei2025enhancing, agyei2025dynamic}. \textbf{(iii) Africa-centric multilingual pretraining}, where models are pretrained or continually pretrained across hundreds of African languages before task-specific use \citep{adebara2023serengetimassivelymultilinguallanguage, adebara2024cheetahnaturallanguagegeneration, li2025ssacometllmsoutperformlearned, buzaaba2025lughallamaadaptinglargelanguage}. \textbf{(iv) Knowledge distillation and LLM-based approaches}, where general-purpose LLMs serve as teacher models, evaluation backbones, or direct interfaces \citep{song2026smalllanguagemodelsilver, glover2024exploring, zhang2025sentiment}. Table~\ref{tab:modeling-taxonomy} summarizes the base architecture and training paradigm used across the reviewed papers. Details of exact model configuration and parameters can be found in \ref{app:model-config}.

\subsection{Model Evaluation Metrics}
\label{sec:survery-results--model-evals-metrics}
Evaluating the performance of MT models is crucial for assessing their effectiveness and guiding future research. While human evaluation is the gold standard, it is time-consuming and expensive. To address these challenges, various automatic evaluation metrics have been developed. The reviewed papers primarily employed three metrics: BLEU \cite{papineni_bleu_2002}, TER \cite{snover_study_2006}, and CHRF \cite{popovic_chrf_2015}. Further details on model evaluation can be found in \ref{app:model-eval}.

Table \ref{tab:summary-bleu-chrf-ter} summarizes the models' BLEU, CHRF, and TER scores. The results highlight the variability in performance across different language pairs and models. Also, Table \ref{tab:summary-SpBLEU-rouge-wer-f1} displays the same data for SpBLEU, ROUGE, WER, and F1 scores.

\section{Conclusion}
\label{sec:conclusion}
This review surfaced a field advancing on a narrow front. Of the 36 core studies, only a handful target languages outside Akan and its dialects, twelve draw on religious corpora as a primary data source, and reported BLEU scores span an unusually wide 0.04--0.81, a range attributable as much to inconsistent reference-translation quality as to model capability. Meanwhile, 26 of the 36 studies appeared in peer-reviewed venues but few released their underlying datasets or checkpoints, so each new project largely repeats the data-construction work of the last. These patterns, rather than any single dataset or model, are the primary obstacle: the field has the methods it needs and lacks the shared infrastructure to apply them across all Ghanaian languages.

Addressing these gaps requires moving beyond single-language Twi models toward shared infrastructure: releasing open checkpoints, establishing standardized benchmark suites, and unifying orthographic practices across Niger-Congo sub-families. This is a problem of sustained commitment and coordination rather than methodological breakthroughs; the roadmap in Section~\ref{sec:discussion} sets out how to sequence that work, but the techniques it calls for already exist. What is missing is the collective will to apply them equitably across all 73 languages, not just the handful currently studied. Whether that will materializes will show up in a straightforward metric that this survey could not report, because it does not yet exist: the number of Ghana's languages with a public dataset, an evaluated model, and a released checkpoint.

Each of these languages carries centuries of accumulated linguistic and cultural heritage. Ensuring that NLP serves that heritage, rather than accelerating its erosion, is one of the clearest responsibilities the field now faces. We hope this survey helps orient the community toward meeting it.

\section*{Limitations}

This survey has several limitations that should inform how its findings are interpreted.

\textbf{Screening reliability}. All screening was performed manually by the authors, with borderline cases resolved by discussion (Section \ref{sec:selection-process}). We did not compute formal inter-annotator agreement statistics for inclusion. While our keyword-based retrievability checks (Section \ref{sec:reproducibility-of-search}) partially mitigate concerns about missed core papers, some subjectivity in inclusion decisions is unavoidable.

\textbf{Non-comparability of reported scores}. The scores in Tables 2 and A1 are drawn from studies that differ in test set, domain, reference-translation quality, tokenization, and metric implementation. Normalizing scores to a common 0–1 scale helps with readability but it does not make them comparable. Differences between rows reflect differences in evaluation setup at least as much as differences in model capability. The wide BLEU range we report (0.04–0.81) should be read as evidence of methodological heterogeneity in the field, not as a ranking of systems, and we caution readers against citing these tables for cross-system comparisons.

\textbf{No citation chasing or grey-literature search}. Our review draws exclusively on four databases queried with a fixed set of boolean, exact-phrase queries in English (Section \ref{sec:methodology-information-sources}). We performed no backward or forward citation chasing, snowballing, or author contact. Relevant work published in venues not indexed by these databases, such as non-English venues, local Ghanaian outlets, or under terminology our queries did not capture, may have been missed. 

\textbf{Publication and reporting bias.} Similar to all literature reviews, this survey can only synthesize what was written down. Negative results, failed data-collection efforts, and community projects that never produced a publication are systematically invisible, so the field's true activity, particularly informal or community-led work on smaller languages, is likely underrepresented here.

\textbf{Limited treatment of speech and audio resources.} Our review is oriented primarily toward text-based NLP and machine translation. Speech-related work, including ASR, TTS, and corpora such as BibleTTS and UGSpeechData, receives comparatively brief treatment (Section~\ref{sec:results--direct-field-recording}). A dedicated survey of the Ghanaian speech-technology landscape would likely surface findings and challenges not captured here.

\textbf{Temporal snapshot of a fast-moving field.} All searches reflect the state of each database in July 2026. The pace of multilingual LLM development means that specific model comparisons and performance figures, are likely to become outdated quickly. This is an inherent limitation of surveying any fast-moving subfield rather than one specific to Ghanaian NLP.

\clearpage
\bibliography{custom}

@inproceedings{oppong2025examining,
    title = "Examining the Cultural Encoding of Gender Bias in {LLM}s for Low-Resourced {A}frican Languages",
    author = "Oppong, Abigail  and
      Nigatu, Hellina Hailu  and
      Okolo, Chinasa T.",
    editor = "Fale{\'n}ska, Agnieszka  and
      Basta, Christine  and
      Costa-juss{\`a}, Marta  and
      Sta{\'n}czak, Karolina  and
      Nozza, Debora",
    booktitle = "Proceedings of the 6th Workshop on Gender Bias in Natural Language Processing (GeBNLP)",
    month = aug,
    year = "2025",
    address = "Vienna, Austria",
    publisher = "Association for Computational Linguistics",
    url = "https://aclanthology.org/2025.gebnlp-1.31/",
    doi = "10.18653/v1/2025.gebnlp-1.31",
    pages = "358--378",
    ISBN = "979-8-89176-277-0"
}

@inproceedings{alabi-etal-2025-charting,
  title     = {Charting the Landscape of {A}frican {NLP}:
               Mapping Progress and Shaping the Road Ahead},
  author    = {Alabi, Jesujoba Oluwadara and
               Hedderich, Michael A. and
               Adelani, David Ifeoluwa and
               Klakow, Dietrich},
  booktitle = {Proceedings of the 2025 Conference on
               Empirical Methods in Natural Language Processing},
  month     = nov,
  year      = {2025},
  address   = {Suzhou, China},
  publisher = {Association for Computational Linguistics},
  pages     = {27807--27841},
  doi       = {10.18653/v1/2025.emnlp-main.1414},
  url       = {https://aclanthology.org/2025.emnlp-main.1414/}
}

@inproceedings{hussen2026state,
  title     = {The State of Large Language Models for African Languages:
               Progress and Challenges},
  author    = {Hussen, Kedir and
               Sewunetie, Walelign and
               Ayele, Abinew and
               Imam, Sukairaj and
               Alemu, Eyob and
               Muhammad, Shamsuddeen and
               Yimam, Seid},
  booktitle = {{DLI} 2025 Research Track},
  series    = {Proceedings of Machine Learning Research},
  volume    = {302},
  pages     = {1--27},
  year      = {2026},
  publisher = {PMLR},
  url       = {https://proceedings.mlr.press/v302/hussen26a.html}
}

@misc{gyamfi2026ghananlp,
      title={Ghana{NLP} {P}arallel {C}orpora: {C}omprehensive {M}ultilingual {R}esources for {L}ow-{R}esource {G}hanaian {L}anguages}, 
      author={Lawrence Adu Gyamfi and Paul Azunre and Stephen Edward Moore and Joel Budu and Akwasi Asare and Mich-Seth Owusu and Jonathan Ofori Asiamah},
      year={2026},
      eprint={2603.13793},
      archivePrefix={arXiv},
      primaryClass={cs.CL},
      url={https://arxiv.org/abs/2603.13793}, 
}

@misc{openai2026gpt56,
  author       = {{OpenAI}},
  title        = {{GPT-5.6 System Card}},
  year         = {2026},
  month        = jul,
  howpublished = {\url{https://deploymentsafety.openai.com/gpt-5-6}},
  note         = {Accessed: 2026-07-22}
}

@inproceedings{kudugunta2023madlad400,
  title     = {{MADLAD}-400: A Multilingual and Document-Level Large Audited Dataset},
  author    = {Kudugunta, Sneha and
               Caswell, Isaac and
               Zhang, Biao and
               Garcia, Xavier and
              Choquette-Choo, Christopher A.  and
             Lee, Katherine and
               Xin, Derrick and
               Kusupati, Aditya and
               Stella, Romi and
               Bapna, Ankur and
               Firat, Orhan},
  booktitle = {Advances in Neural Information Processing Systems},
  volume    = {36},
  year      = {2023}
}

@article{dang2024ayaexpanse,
  title   = {Aya Expanse: Combining Research Breakthroughs for a New Multilingual Frontier},
  author  = {Dang, John and
             Singh, Shivalika and
             D'souza, Daniel and
             Ahmadian, Arash and
             Salamanca, Alejandro and
             Smith, Madeline and
             Peppin, Aidan and
             Hong, Sungjin and
             Govindassamy, Manoj and
             Zhao, Terrence and
             Kublik, Sandra and
             Amer, Meor and
             Aryabumi, Viraat and
             Campos, Jon Ander and
             Tan, Yi-Chern and
             Kocmi, Tom and
             Strub, Florian and
             Grinsztajn, Nathan and
             Flet-Berliac, Yannis and
             Locatelli, Acyr and
             Lin, Hangyu and
             Talupuru, Dwarak and
             Venkitesh, Bharat and
             Cairuz, David and
             Yang, Bowen and
             Chung, Tim and
             Ko, Wei-Yin and
             Shi, Sylvie Shang and
             Shukayev, Amir and
             Bae, Sammie and
             Piktus, Aleksandra and
             Castagn{\'e}, Roman and
             Cruz-Salinas, Felipe and
             Kim, Eddie and
             Crawhall-Stein, Lucas and
             Morisot, Adrien and
             Roy, Sudip and
             Blunsom, Phil and
             Zhang, Ivan and
             Gomez, Aidan and
             Frosst, Nick and
             Fadaee, Marzieh and
             Ermis, Beyza and
             {\"U}st{\"u}n, Ahmet and
             Hooker, Sara},
  journal = {arXiv preprint arXiv:2412.04261},
  year    = {2024},
  doi     = {10.48550/arXiv.2412.04261}
}

@article{tonja2024inkubalm,
  title   = {{InkubaLM}: A Small Language Model for Low-Resource African Languages},
  author  = {Tonja, Atnafu Lambebo and
             Dossou, Bonaventure F. P. and
             Ojo, Jessica and
             Rajab, Jenalea and
             Thior, Fadel and
             Wairagala, Eric Peter and
             Aremu, Anuoluwapo and
             Moiloa, Pelonomi and
             Abbott, Jade and
             Marivate, Vukosi and
             Rosman, Benjamin},
  journal = {arXiv preprint arXiv:2408.17024},
  year    = {2024},
  doi     = {10.48550/arXiv.2408.17024}
}

@inproceedings{ji-etal-2026-data,
  title     = {Data-Centric Continual Pre-training for 500+ Languages:
               A New Bilingual Translation Corpus and Multilingual Models},
  author    = {Ji, Shaoxiong and
               Li, Zihao and
               Paavola, Jaakko and
               Luo, Hengyu and
               Tiedemann, J{\"o}rg},
  booktitle = {Findings of the Association for Computational Linguistics:
               ACL 2026},
  month     = jul,
  year      = {2026},
  address   = {San Diego, California, United States},
  publisher = {Association for Computational Linguistics},
  pages     = {18776--18807},
  doi       = {10.18653/v1/2026.findings-acl.937},
  url       = {https://aclanthology.org/2026.findings-acl.937/}
}

@inproceedings{yu-etal-2026-afriquellm,
  title     = {{AfriqueLLM}: How Data Mixing and Model Architecture Impact
               Continued Pre-training for African Languages},
  author    = {Yu, Hao and
               Xu, Tianyi and
               Hedderich, Michael A. and
               Hamidouche, Wassim and
               Zamir, Syed Waqas and
               Adelani, David Ifeoluwa},
  booktitle = {Proceedings of the 64th Annual Meeting of the Association
               for Computational Linguistics (Volume 1: Long Papers)},
  month     = jul,
  year      = {2026},
  address   = {San Diego, California, United States},
  publisher = {Association for Computational Linguistics},
  pages     = {5909--5928},
  doi       = {10.18653/v1/2026.acl-long.267},
  url       = {https://aclanthology.org/2026.acl-long.267/}
}

@inproceedings{ojo-etal-2025-afrobench,
  title     = {{AfroBench}: How Good Are Large Language Models on
               African Languages?},
  author    = {Ojo, Jessica and
               Ogundepo, Odunayo and
               Oladipo, Akintunde and
               Ogueji, Kelechi and
               Lin, Jimmy and
               Stenetorp, Pontus and
               Adelani, David Ifeoluwa},
  booktitle = {Findings of the Association for Computational Linguistics:
               ACL 2025},
  month     = jul,
  year      = {2025},
  address   = {Vienna, Austria},
  publisher = {Association for Computational Linguistics},
  pages     = {19048--19095},
  doi       = {10.18653/v1/2025.findings-acl.976},
  url       = {https://aclanthology.org/2025.findings-acl.976/}
}

@inproceedings{dione2023masakhapospartofspeechtaggingtypologically,
    title = "{M}asakha{POS}: Part-of-Speech Tagging for Typologically Diverse {A}frican languages",
    author = "Dione, Cheikh M. Bamba  and
      Adelani, David Ifeoluwa  and
      Nabende, Peter  and
      Alabi, Jesujoba  and
      Sindane, Thapelo  and
      Buzaaba, Happy  and
      Muhammad, Shamsuddeen Hassan  and
      Emezue, Chris Chinenye  and
      Ogayo, Perez  and
      Aremu, Anuoluwapo  and
      Gitau, Catherine  and
      Mbaye, Derguene  and
      Mukiibi, Jonathan  and
      Sibanda, Blessing  and
      Dossou, Bonaventure F. P.  and
      Bukula, Andiswa  and
      Mabuya, Rooweither  and
      Tapo, Allahsera Auguste  and
      Munkoh-Buabeng, Edwin  and
      Memdjokam Koagne, Victoire  and
      Ouoba Kabore, Fatoumata  and
      Taylor, Amelia  and
      Kalipe, Godson  and
      Macucwa, Tebogo  and
      Marivate, Vukosi  and
      Gwadabe, Tajuddeen  and
      Elvis, Mboning Tchiaze  and
      Onyenwe, Ikechukwu  and
      Atindogbe, Gratien  and
      Adelani, Tolulope  and
      Akinade, Idris  and
      Samuel, Olanrewaju  and
      Nahimana, Marien  and
      Musabeyezu, Th{\'e}og{\`e}ne  and
      Niyomutabazi, Emile  and
      Chimhenga, Ester  and
      Gotosa, Kudzai  and
      Mizha, Patrick  and
      Agbolo, Apelete  and
      Traore, Seydou  and
      Uchechukwu, Chinedu  and
      Yusuf, Aliyu  and
      Abdullahi, Muhammad  and
      Klakow, Dietrich",
    editor = "Rogers, Anna  and
      Boyd-Graber, Jordan  and
      Okazaki, Naoaki",
    booktitle = "Proceedings of the 61st Annual Meeting of the Association for Computational Linguistics (Volume 1: Long Papers)",
    month = jul,
    year = "2023",
    address = "Toronto, Canada",
    publisher = "Association for Computational Linguistics",
    url = "https://aclanthology.org/2023.acl-long.609/",
    doi = "10.18653/v1/2023.acl-long.609",
    pages = "10883--10900"
}

@inproceedings{adelani-etal-2022-masakhaner,
    title = "{M}asakha{NER} 2.0: {A}frica-centric Transfer Learning for Named Entity Recognition",
    author = "Adelani, David Ifeoluwa  and
      Neubig, Graham  and
      Ruder, Sebastian  and
      Rijhwani, Shruti  and
      Beukman, Michael  and
      Palen-Michel, Chester  and
      Lignos, Constantine  and
      Alabi, Jesujoba O.  and
      Muhammad, Shamsuddeen H.  and
      Nabende, Peter  and
      Dione, Cheikh M. Bamba  and
      Bukula, Andiswa  and
      Mabuya, Rooweither  and
      Dossou, Bonaventure F. P.  and
      Sibanda, Blessing  and
      Buzaaba, Happy  and
      Mukiibi, Jonathan  and
      Kalipe, Godson  and
      Mbaye, Derguene  and
      Taylor, Amelia  and
      Kabore, Fatoumata  and
      Emezue, Chris Chinenye  and
      Aremu, Anuoluwapo  and
      Ogayo, Perez  and
      Gitau, Catherine  and
      Munkoh-Buabeng, Edwin  and
      Memdjokam Koagne, Victoire  and
      Tapo, Allahsera Auguste  and
      Macucwa, Tebogo  and
      Marivate, Vukosi  and
      Mboning, Elvis  and
      Gwadabe, Tajuddeen  and
      Adewumi, Tosin  and
      Ahia, Orevaoghene  and
      Nakatumba-Nabende, Joyce  and
      Mokono, Neo L.  and
      Ezeani, Ignatius  and
      Chukwuneke, Chiamaka  and
      Adeyemi, Mofetoluwa  and
      Hacheme, Gilles Q.  and
      Abdulmumin, Idris  and
      Ogundepo, Odunayo  and
      Yousuf, Oreen  and
      Moteu Ngoli, Tatiana  and
      Klakow, Dietrich",
    editor = "Goldberg, Yoav  and
      Kozareva, Zornitsa  and
      Zhang, Yue",
    booktitle = "Proceedings of the 2022 Conference on Empirical Methods in Natural Language Processing",
    month = dec,
    year = "2022",
    address = "Abu Dhabi, United Arab Emirates",
    publisher = "Association for Computational Linguistics",
    url = "https://aclanthology.org/2022.emnlp-main.298/",
    doi = "10.18653/v1/2022.emnlp-main.298",
    pages = "4488--4508"
}

@misc{buzaaba2025lughallamaadaptinglargelanguage,
      title={Lugha-Llama: Adapting Large Language Models for African Languages}, 
      author={Happy Buzaaba and Alexander Wettig and David Ifeoluwa Adelani and Christiane Fellbaum},
      year={2025},
      eprint={2504.06536},
      archivePrefix={arXiv},
      primaryClass={cs.CL},
      url={https://arxiv.org/abs/2504.06536}, 
}

@inproceedings{muhammad-etal-2023-semeval,
    title = "{S}em{E}val-2023 Task 12: Sentiment Analysis for {A}frican Languages ({A}fri{S}enti-{S}em{E}val)",
    author = "Muhammad, Shamsuddeen Hassan  and
      Abdulmumin, Idris  and
      Yimam, Seid Muhie  and
      Adelani, David Ifeoluwa  and
      Ahmad, Ibrahim Said  and
      Ousidhoum, Nedjma  and
      Ayele, Abinew Ali  and
      Mohammad, Saif  and
      Beloucif, Meriem  and
      Ruder, Sebastian",
    editor = {Ojha, Atul Kr.  and
      Do{\u{g}}ru{\"o}z, A. Seza  and
      Da San Martino, Giovanni  and
      Tayyar Madabushi, Harish  and
      Kumar, Ritesh  and
      Sartori, Elisa},
    booktitle = "Proceedings of the 17th International Workshop on Semantic Evaluation (SemEval-2023)",
    month = jul,
    year = "2023",
    address = "Toronto, Canada",
    publisher = "Association for Computational Linguistics",
    url = "https://aclanthology.org/2023.semeval-1.315/",
    doi = "10.18653/v1/2023.semeval-1.315",
    pages = "2319--2337"
}

@article{gyimah2025asantetwisenti,
    title = {AsanteTwiSenti: A Sentiment dataset of Ghanaian Asante Twi Tweets in a multilingual context},
    journal = {Data in Brief},
    volume = {60},
    pages = {111460},
    year = {2025},
    issn = {2352-3409},
    doi = {https://doi.org/10.1016/j.dib.2025.111460},
    url = {https://www.sciencedirect.com/science/article/pii/S2352340925001921},
    author = {Mavis Sarah Gyimah and James Benjamin Hayfron -Acquah and Rose-Mary Mensah Gyening and Michael Asante and Umar Farouk Ibn Abdulrahman and Evans Kotei}
    }

@inproceedings{glover2024exploring,
  author={Glover-Tay, Senam Afi and Ayorkor Korsah, G.},
  booktitle={2024 IEEE 9th International Conference on Adaptive Science and Technology (ICAST)}, 
  title={Exploring the Efficacy of a Local-Language Interfaced ChatGPT as A Self-Learning Support Tool for Junior High School Students in Rural Communities in Africa}, 
  year={2024},
  volume={9},
  number={},
  pages={1-8},
  doi={10.1109/ICAST61769.2024.10856463}
}

@article{zhang2025sentiment, 
    title={Sentiment Analysis for the African Language Twi: Translation-based vs. End-to-End Approaches}, 
    volume={38}, 
    url={https://journals.flvc.org/FLAIRS/article/view/138684}, 
    DOI={10.32473/flairs.38.1.138684}, 
    abstractNote={&amp;lt;p&amp;gt;This paper presents our approach to sentiment analysis for Twi, a low-resource African language. We developed two types of systems: a translation-based system and an end-to-end system. These systems leverage various popular large-scale language models (LLMs), such as BERT and its multilingual variants, and their performances were compared. Our evaluation focused on the accuracy and robustness of these systems in identifying sentiments within Twi text. We also explored the challenges associated with low-resource languages, including limited annotated datasets and the need for effective cross-lingual transfer. The results highlight the potential of end-to-end multilingual LLMs for low-resource languages while emphasizing the importance of translation quality in translation-based approaches to sentiment analysis tasks. Additionally, we provide insights into the practical implications of our findings for future research.&amp;lt;/p&amp;gt;&amp;lt;p&amp;gt;Accessibility Summary:&amp;lt;/p&amp;gt;&amp;lt;p&amp;gt;In accordance with Title II regulations this content meets all points of exemption as Archived web content and/or Preexisting conventional electronic documents.&amp;lt;/p&amp;gt;}, 
    number={1}, 
    journal={The International FLAIRS Conference Proceedings}, 
    author={Zhang, Linrui and Copus, Belinda}, 
    year={2025}, 
    month={May} 
}

@inproceedings{agyei2025enhancing,
  title={Enhancing Machine Translation for Low-Resource Languages: A Cross-Lingual Learning Approach for TWI},
  author={Agyei, Emmanuel and Xiaoling, Zhang and Quaye, Ama Bonuah and Adeyi, Odeh Victor and Arhin, Joseph Roger and others},
  booktitle={CS \& IT Conference Proceedings},
  volume={15},
  number={7},
  year={2025},
  organization={CS \& IT Conference Proceedings}
}

@article{agyei2025dynamic,
    author = {Agyei, Emmanuel and Zhang, Xiaoling and Quaye, Ama Bonuah and Odeh, Victor Adeyi and Arhin, Joseph Roger},
    title = {Dynamic Aggregation and Augmentation for Low-Resource Machine Translation Using Federated Fine-Tuning of Pretrained Transformer Models},
    journal = {Applied Sciences},
    volume = {15},
    year = {2025},
    number = {8},
    article-number = {4494},
    url = {https://www.mdpi.com/2076-3417/15/8/4494},
    issn = {2076-3417},
    doi = {10.3390/app15084494}
}

@article{agyei2024low,
  author    = {Agyei, Emmanuel and Zhang, Xiaoling and Bannerman, Stephen and Quaye, Ama Bonuah campaign and Yussi, Sophyani Banaamwini and Agbesi, Victor Kwaku},
  title     = {Low resource Twi-English parallel corpus for machine translation in multiple domains (Twi-2-ENG)},
  journal   = {Discover Computing},
  year      = {2024},
  volume    = {27},
  number    = {1},
  pages     = {17},
  month     = {jul},
  day       = {5},
  issn      = {2948-2992},
  doi       = {10.1007/s10791-024-09451-8},
  url       = {https://doi.org/10.1007/s10791-024-09451-8}
}

@article{raychawdhary2025empowering,
  author={Raychawdhary, Nilanjana and Bhattacharya, Sutanu and Seals, Cheryl and Dozier, Gerry V.},
  journal={IEEE Access}, 
  title={Empowering Sentiment Analysis in African Low-Resource Languages Through Transformer Models and Strategic Language Selection}, 
  year={2025},
  volume={13},
  number={},
  pages={147859-147873},
  doi={10.1109/ACCESS.2025.3599480}}

@techreport{bannerman2024machine,
  author      = {Bannerman, Stephen and Agyei, Emmanuel and Sarpong, Sarah and Quaye, Ama Bonuah and Yussif, Sophyani Banaamwini and Agbesi, Victor Kweku},
  title       = {Machine Translation from English-Twi in Parallel Corpus: Low Resource Ghanaian Language (Nlp)},
  institution = {Social Science Research Network (SSRN)},
  year        = {2024},
  type        = {Preprint},
  url         = {https://ssrn.com/abstract=4761197},
  doi         = {10.2139/ssrn.4761197}
}

@article{chowdhery_palm_2022,
    author = {Chowdhery, Aakanksha and Narang, Sharan and Devlin, Jacob and Bosma, Maarten and Mishra, Gaurav and Roberts, Adam and Barham, Paul and Chung, Hyung Won and Sutton, Charles and Gehrmann, Sebastian and Schuh, Parker and Shi, Kensen and Tsvyashchenko, Sashank and Maynez, Joshua and Rao, Abhishek and Barnes, Parker and Tay, Yi and Shazeer, Noam and Prabhakaran, Vinodkumar and Reif, Emily and Du, Nan and Hutchinson, Ben and Pope, Reiner and Bradbury, James and Austin, Jacob and Isard, Michael and Gur-Ari, Guy and Yin, Pengcheng and Duke, Toju and Levskaya, Anselm and Ghemawat, Sanjay and Dev, Sunipa and Michalewski, Henryk and Garcia, Xavier and Misra, Vedant and Robinson, Kevin and Fedus, Liam and Zhou, Denny and Ippolito, Daphne and Luan, David and Lim, Hyeontaek and Zoph, Barret and Spiridonov, Alexander and Sepassi, Ryan and Dohan, David and Agrawal, Shivani and Omernick, Mark and Dai, Andrew M. and Pillai, Thanumalayan Sankaranarayana and Pellat, Marie and Lewkowycz, Aitor and Moreira, Erica and Child, Rewon and Polozov, Oleksandr and Lee, Katherine and Zhou, Zongwei and Wang, Xuezhi and Saeta, Brennan and Diaz, Mark and Firat, Orhan and Catasta, Michele and Wei, Jason and Meier-Hellstern, Kathy and Eck, Douglas and Dean, Jeff and Petrov, Slav and Fiedel, Noah},
    title = {PaLM: scaling language modeling with pathways},
    year = {2023},
    issue_date = {January 2023},
    publisher = {JMLR.org},
    volume = {24},
    number = {1},
    issn = {1532-4435},
    journal = {J. Mach. Learn. Res.},
    month = jan,
    articleno = {240},
    numpages = {113}
}

@article{gyasi_twi_2023,
    author = {Gyasi, Frederick and Schlippe, Tim},
    title = {Twi Machine Translation},
    journal = {Big Data and Cognitive Computing},
    volume = {7},
    year = {2023},
    number = {2},
    article-number = {114},
    url = {https://www.mdpi.com/2504-2289/7/2/114},
    issn = {2504-2289},
    doi = {10.3390/bdcc7020114}
}

@misc{orife_masakhane_2020,
      title={Masakhane -- Machine Translation For Africa}, 
      author={Iroro Orife and Julia Kreutzer and Blessing Sibanda and Daniel Whitenack and Kathleen Siminyu and Laura Martinus and Jamiil Toure Ali and Jade Abbott and Vukosi Marivate and Salomon Kabongo and Musie Meressa and Espoir Murhabazi and Orevaoghene Ahia and Elan van Biljon and Arshath Ramkilowan and Adewale Akinfaderin and Alp Öktem and Wole Akin and Ghollah Kioko and Kevin Degila and Herman Kamper and Bonaventure Dossou and Chris Emezue and Kelechi Ogueji and Abdallah Bashir},
      year={2020},
      eprint={2003.11529},
      archivePrefix={arXiv},
      primaryClass={cs.CL},
      url={https://arxiv.org/abs/2003.11529}, 
}

@article{mabokela_multilingual_2023,
  author={Mabokela, Koena Ronny and Celik, Turgay and Raborife, Mpho},
  journal={IEEE Access}, 
  title={Multilingual Sentiment Analysis for Under-Resourced Languages: A Systematic Review of the Landscape}, 
  year={2023},
  volume={11},
  number={},
  pages={15996-16020},
  doi={10.1109/ACCESS.2022.3224136}
}

@article{ranathunga_neural_2023,
    author = {Ranathunga, Surangika and Lee, En-Shiun Annie and Prifti Skenduli, Marjana and Shekhar, Ravi and Alam, Mehreen and Kaur, Rishemjit},
    title = {Neural Machine Translation for Low-resource Languages: A Survey},
    year = {2023},
    issue_date = {November 2023},
    publisher = {Association for Computing Machinery},
    address = {New York, NY, USA},
    volume = {55},
    number = {11},
    issn = {0360-0300},
    url = {https://doi.org/10.1145/3567592},
    doi = {10.1145/3567592},
    journal = {ACM Comput. Surv.},
    month = feb,
    articleno = {229},
    numpages = {37}
}

@incollection{etim_state---art_2021,
	title = {A {State}-of-the-{Art} {Review} of {Nigerian} {Languages} {Natural} {Language} {Processing} {Research}:},
	isbn = {9781799834687 9781799834700},
	shorttitle = {A {State}-of-the-{Art} {Review} of {Nigerian} {Languages} {Natural} {Language} {Processing} {Research}},
	url = {http://services.igi-global.com/resolvedoi/resolve.aspx?doi=10.4018/978-1-7998-3468-7.ch008},
	urldate = {2024-03-18},
	booktitle = {Advances in {IT} {Standards} and {Standardization} {Research}},
	publisher = {IGI Global},
	author = {Asubiaro, Toluwase Victor and Igwe, Ebelechukwu Gloria},
	editor = {Etim, Alice S.},
	year = {2021},
	doi = {10.4018/978-1-7998-3468-7.ch008},
	pages = {147--167},
}

@article{haddow_survey_2022,
    title = "Survey of Low-Resource Machine Translation",
    author = "Haddow, Barry  and
      Bawden, Rachel  and
      Miceli Barone, Antonio Valerio  and
      Helcl, Jind{\v{r}}ich  and
      Birch, Alexandra",
    journal = "Computational Linguistics",
    volume = "48",
    number = "3",
    month = sep,
    year = "2022",
    address = "Cambridge, MA",
    publisher = "MIT Press",
    url = "https://aclanthology.org/2022.cl-3.6/",
    doi = "10.1162/coli_a_00446",
    pages = "673--732"
}

@misc{ghana_statistical_service_ghana_2023,
	title = {Ghana {Fact} {Sheet}},
	url = {https://statsghana.gov.gh/ghfactsheet.php},
	author = {{Ghana Statistical Service}},
	year = {2023},
}

@misc{eberhard_ethnologue_2024,
	title = {Ethnologue: {Languages} of the {World}},
	url = {https://www.ethnologue.com/country/GH/},
	language = {en},
	urldate = {2023-04-17},
	journal = {Ethnologue (Free All)},
	author = {Eberhard, David M. and Simons, Gary F. and Fennig, Charles D.},
	year = {2024},
}

@misc{radford_improving_2018,
	title = {Improving {Language} {Understanding} by {Generative} {Pre}-{Training}},
	url = {https://www.cs.ubc.ca/~amuham01/LING530/papers/radford2018improving.pdf},
	language = {en},
	author = {Radford, Alec and Narasimhan, Karthik and Salimans, Tim and Sutskever, Ilya},
	year = {2018},
}

@article{brown_statistical_1990,
	title = {A statistical approach to machine translation},
	volume = {16},
	issn = {0891-2017},
	number = {2},
	journal = {Computational Linguistics},
	author = {Brown, Peter F. and Cocke, John and Pietra, Stephen A. Della and Pietra, Vincent J. Della and Jelinek, Fredrick and Lafferty, John D. and Mercer, Robert L. and Roossin, Paul S.},
	month = jun,
	year = {1990},
	pages = {79--85},
}

@article{wang_progress_2022,
	title = {Progress in {Machine} {Translation}},
	volume = {18},
	issn = {2095-8099},
	url = {https://www.sciencedirect.com/science/article/pii/S2095809921002745},
	doi = {10.1016/j.eng.2021.03.023},
	language = {en},
	urldate = {2023-06-15},
	journal = {Engineering},
	author = {Wang, Haifeng and Wu, Hua and He, Zhongjun and Huang, Liang and Church, Kenneth Ward},
	month = nov,
	year = {2022},
	pages = {143--153},
}

@article{costa-jussa_latest_2015,
	series = {Hybrid {Machine} {Translation}: integration of linguistics and statistics},
	title = {Latest trends in hybrid machine translation and its applications},
	volume = {32},
	issn = {0885-2308},
	url = {https://www.sciencedirect.com/science/article/pii/S0885230814001077},
	doi = {10.1016/j.csl.2014.11.001},
	language = {en},
	number = {1},
	urldate = {2023-06-15},
	journal = {Computer Speech \& Language},
	author = {Costa-jussà, Marta R. and Fonollosa, José A. R.},
	month = jul,
	year = {2015},
	pages = {3--10},
}

@inproceedings{hunsicker_machine_2012,
	address = {USA},
	series = {{WMT} '12},
	title = {Machine learning for hybrid machine translation},
	urldate = {2023-06-15},
	booktitle = {Proceedings of the {Seventh} {Workshop} on {Statistical} {Machine} {Translation}},
	publisher = {Association for Computational Linguistics},
	author = {Hunsicker, Sabine and Yu, Chen and Federmann, Christian},
	month = jun,
	year = {2012},
	pages = {312--316},
}

@inproceedings{swathi_machine_2020,
	address = {Cham},
	title = {Machine {Translation} {Using} {Deep} {Learning}: {A} {Comparison}},
	isbn = {9783030240516},
	shorttitle = {Machine {Translation} {Using} {Deep} {Learning}},
	doi = {10.1007/978-3-030-24051-6_38},
	language = {en},
	booktitle = {Proceedings of {International} {Conference} on {Artificial} {Intelligence}, {Smart} {Grid} and {Smart} {City} {Applications}},
	publisher = {Springer International Publishing},
	author = {Swathi, S. and Jayashree, L. S.},
	editor = {Kumar, L. Ashok and Jayashree, L. S. and Manimegalai, R.},
	year = {2020},
	pages = {389--395},
}

@article{maazouzi_systematic_2017,
	title = {A {SYSTEMATIC} {READING} {IN} {STATISTICAL} {TRANSLATION}: {FROM} {THE} {STATISTICAL} {MACHINE} {TRANSLATION} {TO} {THE} {NEURAL} {TRANSLATION} {MODELS}.},
	volume = {16},
	copyright = {Copyright (c) 2020 Journal of Information and Communication Technology},
	issn = {2180-3862},
	shorttitle = {A {SYSTEMATIC} {READING} {IN} {STATISTICAL} {TRANSLATION}},
	url = {https://e-journal.uum.edu.my/index.php/jict/article/view/8239},
	language = {en},
	number = {2},
	urldate = {2023-06-15},
	journal = {Journal of Information and Communication Technology},
	author = {Maazouzi, Zakaria El and Mohajir, Badr Eddine El and Achhab, Mohammed Al},
	month = nov,
	year = {2017},
	pages = {408--441},
}

@article{mohamed_neural_2021,
	title = {Neural machine translation: past, present, and future},
	volume = {33},
	issn = {1433-3058},
	shorttitle = {Neural machine translation},
	url = {https://doi.org/10.1007/s00521-021-06268-0},
	doi = {10.1007/s00521-021-06268-0},
	language = {en},
	number = {23},
	urldate = {2023-06-15},
	journal = {Neural Computing and Applications},
	author = {Mohamed, Shereen A. and Elsayed, Ashraf A. and Hassan, Y. F. and Abdou, Mohamed A.},
	month = dec,
	year = {2021},
	pages = {15919--15931},
}

@misc{raffel_exploring_2020,
	title = {Exploring the {Limits} of {Transfer} {Learning} with a {Unified} {Text}-to-{Text} {Transformer}},
	url = {http://arxiv.org/abs/1910.10683},
	doi = {10.48550/arXiv.1910.10683},
	urldate = {2023-06-15},
	publisher = {arXiv},
	author = {Raffel, Colin and Shazeer, Noam and Roberts, Adam and Lee, Katherine and Narang, Sharan and Matena, Michael and Zhou, Yanqi and Li, Wei and Liu, Peter J.},
	month = jul,
	year = {2020},
	note = {arXiv:1910.10683 [cs, stat]},
}

@misc{devlin_bert_2019,
	title = {{BERT}: {Pre}-training of {Deep} {Bidirectional} {Transformers} for {Language} {Understanding}},
	shorttitle = {{BERT}},
	url = {http://arxiv.org/abs/1810.04805},
	doi = {10.48550/arXiv.1810.04805},
	urldate = {2023-06-15},
	publisher = {arXiv},
	author = {Devlin, Jacob and Chang, Ming-Wei and Lee, Kenton and Toutanova, Kristina},
	month = may,
	year = {2019},
	note = {arXiv:1810.04805 [cs]},
}

@article{wang_pre-trained_2022,
	title = {Pre-{Trained} {Language} {Models} and {Their} {Applications}},
	issn = {2095-8099},
	url = {https://www.sciencedirect.com/science/article/pii/S2095809922006324},
	doi = {10.1016/j.eng.2022.04.024},
	language = {en},
	urldate = {2023-06-15},
	journal = {Engineering},
	author = {Wang, Haifeng and Li, Jiwei and Wu, Hua and Hovy, Eduard and Sun, Yu},
	month = sep,
	year = {2022},
}

@misc{siminyu_ai4d_2021,
	title = {{AI4D} -- {African} {Language} {Program}},
	url = {http://arxiv.org/abs/2104.02516},
	doi = {10.48550/arXiv.2104.02516},
	urldate = {2023-06-15},
	publisher = {arXiv},
	author = {Siminyu, Kathleen and Kalipe, Godson and Orlic, Davor and Abbott, Jade and Marivate, Vukosi and Freshia, Sackey and Sibal, Prateek and Neupane, Bhanu and Adelani, David I. and Taylor, Amelia and ALI, Jamiil Toure and Degila, Kevin and Balogoun, Momboladji and DIOP, Thierno Ibrahima and David, Davis and Fourati, Chayma and Haddad, Hatem and Naski, Malek},
	month = apr,
	year = {2021},
	note = {arXiv:2104.02516 [cs]},
}

@misc{magueresse_low-resource_2020,
	title = {Low-resource {Languages}: {A} {Review} of {Past} {Work} and {Future} {Challenges}},
	shorttitle = {Low-resource {Languages}},
	url = {http://arxiv.org/abs/2006.07264},
	doi = {10.48550/arXiv.2006.07264},
	urldate = {2023-06-15},
	publisher = {arXiv},
	author = {Magueresse, Alexandre and Carles, Vincent and Heetderks, Evan},
	month = jun,
	year = {2020},
	note = {arXiv:2006.07264 [cs]},
}

@article{adika_english_2012,
	title = {English in {Ghana}: {Growth}, {Tensions}, and {Trends}},
	volume = {1},
	copyright = {Copyright (c)},
	issn = {2241-7214},
	shorttitle = {English in {Ghana}},
	url = {https://ejournals.epublishing.ekt.gr/index.php/latic/article/view/2723},
	doi = {10.12681/ijltic.17},
	language = {en},
	urldate = {2023-06-15},
	journal = {International Journal of Language, Translation and Intercultural Communication},
	author = {Adika, Gordon Senanu Kwame},
	month = jan,
	year = {2012},
	pages = {151--166},
}

@article{awedoba_attitudes_2009,
	title = {Attitudes towards instruction in the local language – a case study of the perspectives of the ‘small stakeholder’},
	copyright = {Copyright (c)},
	issn = {0855-4412},
	url = {https://www.ajol.info/index.php/rrias/article/view/45973},
	doi = {10.4314/rrias.v25i2.45973},
	language = {en},
	urldate = {2023-06-15},
	journal = {Research Review of the Institute of African Studies},
	author = {Awedoba, A.},
	year = {2009},
}

@article{davis_language_2012,
	title = {Language policy and instructional practice dichotomy: {The} case of primary schools in {Ghana}},
	volume = {53},
	issn = {0883-0355},
	shorttitle = {Language policy and instructional practice dichotomy},
	url = {https://www.sciencedirect.com/science/article/pii/S0883035512000390},
	doi = {10.1016/j.ijer.2012.04.007},
	language = {en},
	urldate = {2023-06-15},
	journal = {International Journal of Educational Research},
	author = {Davis, Ernest and Agbenyega, Joseph S.},
	month = jan,
	year = {2012},
	pages = {341--347},
}

@article{owu-ewie_language_2017,
	title = {Language, {Education} and {Linguistic} {Human} {Rights} in {Ghana}},
	volume = {28},
	copyright = {Copyright (c)},
	issn = {2458-746X},
	url = {https://www.ajol.info/index.php/ljh/article/view/163878},
	doi = {10.4314/ljh.v28i2},
	language = {en},
	number = {2},
	urldate = {2023-06-15},
	journal = {Legon Journal of the Humanities},
	author = {Owu-Ewie, Charles},
	month = oct,
	year = {2017},
	pages = {151--172},
}

@techreport{usaid_ghana_2020,
	title = {Ghana.  {Language} of {Instruction} {Country} {Profile}: {Ghana}},
	url = {https://pdf.usaid.gov/pdf_docs/PA00X9JT.pdf},
	author = {{USAID}},
        institution = {{USAID}},
	year = {2020},
}

@techreport{unesco_atlas_2010,
	title = {Atlas of the {World}’s {Languages} in {Danger}},
	url = {https://unesdoc.unesco.org/in/documentViewer.xhtml?v=2.1.196&id=p::usmarcdef_0000187026&file=/in/rest/annotationSVC/DownloadWatermarkedAttachment/attach_import_70c069f5-be69-478d-80ca-47a6ce68c154%3F_%3D187026eng.pdf&locale=en&multi=true&ark=/ark:/48223/pf0000187026/PDF/187026eng.pdf#%5B%7B%22num%22%3A206%2C%22gen%22%3A0%7D%2C%7B%22name%22%3A%22XYZ%22%7D%2Cnull%2Cnull%2C0%5D},
	author = {{UNESCO}},
	year = {2010},
        institution = {{UNESCO}},
}

@inproceedings{agic_jw300_2019,
	address = {Florence, Italy},
	title = {{JW300}: {A} {Wide}-{Coverage} {Parallel} {Corpus} for {Low}-{Resource} {Languages}},
	shorttitle = {{JW300}},
	url = {https://aclanthology.org/P19-1310},
	doi = {10.18653/v1/P19-1310},
	urldate = {2023-04-20},
	booktitle = {Proceedings of the 57th {Annual} {Meeting} of the {Association} for {Computational} {Linguistics}},
	publisher = {Association for Computational Linguistics},
	author = {Agić, Željko and Vulić, Ivan},
	month = jul,
	year = {2019},
	pages = {3204--3210},
}

@misc{ogueji_pidginunmt_2019,
	title = {{PidginUNMT}: {Unsupervised} {Neural} {Machine} {Translation} from {West} {African} {Pidgin} to {English}},
	shorttitle = {{PidginUNMT}},
	url = {http://arxiv.org/abs/1912.03444},
	doi = {10.48550/arXiv.1912.03444},
	urldate = {2023-04-17},
	publisher = {arXiv},
	author = {Ogueji, Kelechi and Ahia, Orevaoghene},
	month = dec,
	year = {2019},
	note = {arXiv:1912.03444 [cs]},
}

@inproceedings{adebara_afrolid_2022,
    title = "{A}fro{LID}: A Neural Language Identification Tool for {A}frican Languages",
    author = "Adebara, Ife  and
      Elmadany, AbdelRahim  and
      Abdul-Mageed, Muhammad  and
      Inciarte, Alcides",
    editor = "Goldberg, Yoav  and
      Kozareva, Zornitsa  and
      Zhang, Yue",
    booktitle = "Proceedings of the 2022 Conference on Empirical Methods in Natural Language Processing",
    month = dec,
    year = "2022",
    address = "Abu Dhabi, United Arab Emirates",
    publisher = "Association for Computational Linguistics",
    url = "https://aclanthology.org/2022.emnlp-main.128/",
    doi = "10.18653/v1/2022.emnlp-main.128",
    pages = "1958--1981"
}

@article{yvette_gelr_2021,
	title = {{GELR}: {A} {Bilingual} {Ewe}-{English} {Corpus} {Building} and {Evaluation}},
	volume = {10},
	issn = {2278-0181},
	shorttitle = {{GELR}},
	url = {https://www.ijert.org/research/gelr-a-bilingual-ewe-english-corpus-building-and-evaluation-IJERTV10IS080214.pdf, https://www.ijert.org/gelr-a-bilingual-ewe-english-corpus-building-and-evaluation},
	doi = {10.17577/IJERTV10IS080214},
	language = {en-US},
	number = {8},
	urldate = {2023-04-17},
	journal = {International Journal of Engineering Research \& Technology},
	author = {Yvette, Gbedevi Akouyo and Zhang, Dr Kevin and Jude, Tchaye-Kondi},
	month = sep,
	year = {2021},
	note = {Publisher: IJERT-International Journal of Engineering Research \& Technology},
}

@misc{azunre_nlp_2021,
	title = {{NLP} for {Ghanaian} {Languages}},
	url = {http://arxiv.org/abs/2103.15475},
	doi = {10.48550/arXiv.2103.15475},
	urldate = {2023-04-17},
	publisher = {arXiv},
	author = {Azunre, Paul and Osei, Salomey and Addo, Salomey and Adu-Gyamfi, Lawrence Asamoah and Moore, Stephen and Adabankah, Bernard and Opoku, Bernard and Asare-Nyarko, Clara and Nyarko, Samuel and Amoaba, Cynthia and Appiah, Esther Dansoa and Akwerh, Felix and Lawson, Richard Nii Lante and Budu, Joel and Debrah, Emmanuel and Boateng, Nana and Ofori, Wisdom and Buabeng-Munkoh, Edwin and Adjei, Franklin and Ampomah, Isaac Kojo Essel and Otoo, Joseph and Borkor, Reindorf and Mensah, Standylove Birago and Mensah, Lucien and Marcel, Mark Amoako and Amponsah, Anokye Acheampong and Hayfron-Acquah, James Ben},
	month = apr,
	year = {2021},
	note = {arXiv:2103.15475 [cs]},
}

@inproceedings{adjeisah_twi_2020,
	title = {Twi {Corpus}: {A} {Massively} {Twi}-to-{Handful} {Languages} {Parallel} {Bible} {Corpus}},
	shorttitle = {Twi {Corpus}},
	doi = {10.1109/ISPA-BDCloud-SocialCom-SustainCom51426.2020.00157},
	booktitle = {2020 {IEEE} {Intl} {Conf} on {Parallel} \& {Distributed} {Processing} with {Applications}, {Big} {Data} \& {Cloud} {Computing}, {Sustainable} {Computing} \& {Communications}, {Social} {Computing} \& {Networking} ({ISPA}/{BDCloud}/{SocialCom}/{SustainCom})},
	author = {Adjeisah, Michael and Liu, Guohua and Nortey, Richard Nuetey and Song, Jinling and Lamptey, Khalid Odartey and Frimpong, Felix Nana},
	month = dec,
	year = {2020},
	pages = {1043--1049},
}

@misc{azunre_contextual_2021,
	title = {Contextual {Text} {Embeddings} for {Twi}},
	url = {http://arxiv.org/abs/2103.15963},
	doi = {10.48550/arXiv.2103.15963},
	urldate = {2023-04-17},
	publisher = {arXiv},
	author = {Azunre, Paul and Osei, Salomey and Addo, Salomey and Adu-Gyamfi, Lawrence Asamoah and Moore, Stephen and Adabankah, Bernard and Opoku, Bernard and Asare-Nyarko, Clara and Nyarko, Samuel and Amoaba, Cynthia and Appiah, Esther Dansoa and Akwerh, Felix and Lawson, Richard Nii Lante and Budu, Joel and Debrah, Emmanuel and Boateng, Nana and Ofori, Wisdom and Buabeng-Munkoh, Edwin and Adjei, Franklin and Ampomah, Isaac Kojo Essel and Otoo, Joseph and Borkor, Reindorf and Mensah, Standylove Birago and Mensah, Lucien and Marcel, Mark Amoako and Amponsah, Anokye Acheampong and Hayfron-Acquah, James Ben},
	month = mar,
	year = {2021},
	note = {arXiv:2103.15963 [cs]},
}

@misc{afram_twieng_2022,
	title = {{TWIENG}: {A} {Multi}-{Domain} {Twi}-{English} {Parallel} {Corpus} for {Machine} {Translation} of {Twi}, a {Low}-{Resource} {African} {Language}},
	shorttitle = {{TWIENG}},
	url = {https://www.preprints.org/manuscript/202203.0303/v1},
	doi = {10.20944/preprints202203.0303.v1},
	language = {en},
	urldate = {2023-04-17},
	publisher = {Preprints},
	author = {Afram, Gabriel Kwadwo and Weyori, Benjamin Asubam and Adekoya, Felix Adebayo},
	month = mar,
	year = {2022},
}

@article{galla_indigenous_2016,
	title = {Indigenous {Language} {Revitalization}, {Promotion}, and {Education}: {Function} of {Digital} {Technology}},
	volume = {29},
	issn = {0958-8221},
	shorttitle = {Indigenous {Language} {Revitalization}, {Promotion}, and {Education}},
	doi = {10.1080/09588221.2016.1166137},
	language = {en},
	number = {7},
	urldate = {2023-04-17},
	journal = {Computer Assisted Language Learning},
	author = {Galla, Candace Kaleimamoowahinekapu},
	year = {2016},
	note = {ERIC Number: EJ1117501},
	pages = {1137--1151},
}

@article{akpanglo-nartey_endangered_2012,
	title = {Some {Endangered} {Languages} of {Ghana}},
	volume = {1},
	copyright = {http://creativecommons.org/licenses/by/2.0/},
	url = {http://article.sapub.org/The phenomenon of language endangerment and, ultimately, language loss is considered in regard to indigenous Ghanaian languages. It is established that two languages, namely, Ghanaian English (GhE) and Akan, especially the Twi dialect, and to a small degree, Ewe, are slowly killing off the smaller Ghanaian languages. For instance, in 1970 almost all Winneba natives spoke Efutu (Ewutu) as their first language. By 2010, 40 years later, only approximately 50% of children born to the Winneba natives speak Efutu as a first language. About 30% of these children speak no Efutu at all. Interestingly, medium-sized languages such as Ga, Dangme and Nzema are also slowly losing grounds to the three languages cited. Meanwhile there are some dozen Ghanaian languages that have less than 1000 estimated speakers each but which have held their own for a century. It is concluded that the closer a language community is to the major urban centers, the more likely it is to be endangered. It is further concluded that the language policy of the Ghana Government is contributing to the loss of Ghanaian languages._Url},
	language = {en},
	number = {2},
	urldate = {2023-04-17},
	journal = {American Journal of Linguistics},
	author = {Akpanglo-Nartey, Jonas N. and Akpanglo-Nartey, Rebecca A.},
	year = {2012},
	pages = {10--18},
}

@techreport{schachter_phonology_1968,
	title = {A {Phonology} of {Akan}: {Akuapem}, {Asante}, {Fante}},
	shorttitle = {A {Phonology} of {Akan}},
	url = {https://eric.ed.gov/?id=ED022189},
	language = {en},
	urldate = {2023-04-17},
	institution = {Textbook Department, Student Store, University of California, Los Angeles, California 90024 (\$3},
	author = {Schachter, Paul and Fromkin, Victoria},
	month = sep,
	year = {1968},
	note = {ERIC Number: ED022189},
}

@misc{osam_introduction_2003,
	title = {An introduction to the verbal and multi-verbal system of {Akan}},
	url = {https://www.academia.edu/6364317/AN_INTRODUCTION_TO_THE_VERBAL_AND_MULTI_VERBAL_SYSTEM_OF_AKAN},
	urldate = {2023-04-17},
	author = {Osam, Kweku},
	month = jan,
	year = {2003},
}

@incollection{childs_language_2020,
	title = {Language {Endangerment} in {Africa}},
	isbn = {9780199384655},
	url = {https://oxfordre.com/linguistics/view/10.1093/acrefore/9780199384655.001.0001/acrefore-9780199384655-e-102},
	language = {en},
	urldate = {2024-03-20},
	booktitle = {Oxford {Research} {Encyclopedia} of {Linguistics}},
	publisher = {Oxford University Press},
	author = {Childs, Tucker},
	collaborator = {Childs, Tucker},
	month = jan,
	year = {2020},
	doi = {10.1093/acrefore/9780199384655.013.102},
}

@book{batibo_language_2005,
	title = {Language decline and death in {Africa}: {Causes}, consequences and challenges},
	isbn = {9781853598098},
	language = {English},
	publisher = {Multilingual Matters},
	author = {Batibo, Herman},
	month = may,
	year = {2005},
}

@inproceedings{popovic_chrf_2015,
	address = {Lisbon, Portugal},
	title = {{chrF}: character n-gram {F}-score for automatic {MT} evaluation},
	shorttitle = {{chrF}},
	url = {https://aclanthology.org/W15-3049},
	doi = {10.18653/v1/W15-3049},
	urldate = {2023-04-15},
	booktitle = {Proceedings of the {Tenth} {Workshop} on {Statistical} {Machine} {Translation}},
	publisher = {Association for Computational Linguistics},
	author = {Popović, Maja},
	month = sep,
	year = {2015},
	pages = {392--395},
}

@inproceedings{lin-2004-rouge,
    title = "{ROUGE}: A Package for Automatic Evaluation of Summaries",
    author = "Lin, Chin-Yew",
    booktitle = "Text Summarization Branches Out",
    month = jul,
    year = "2004",
    address = "Barcelona, Spain",
    publisher = "Association for Computational Linguistics",
    url = "https://aclanthology.org/W04-1013/",
    pages = "74--81"
}

@inproceedings{snover_study_2006,
	address = {Cambridge, Massachusetts, USA},
	title = {A {Study} of {Translation} {Edit} {Rate} with {Targeted} {Human} {Annotation}},
	url = {https://aclanthology.org/2006.amta-papers.25},
	urldate = {2023-04-15},
	booktitle = {Proceedings of the 7th {Conference} of the {Association} for {Machine} {Translation} in the {Americas}: {Technical} {Papers}},
	publisher = {Association for Machine Translation in the Americas},
	author = {Snover, Matthew and Dorr, Bonnie and Schwartz, Rich and Micciulla, Linnea and Makhoul, John},
	month = aug,
	year = {2006},
	pages = {223--231},
}

@article{papineni_bleu_2002,
	title = {{BLEU}: a {Method} for {Automatic} {Evaluation} of {Machine} {Translation}},
	journal = {40th Annual Meeting of the Association for Computational Linguistics},
	author = {Papineni, Kishore and Roukos, Salim and Ward, Todd and Wei-Jing, Zhu},
	month = jul,
	year = {2002},
	pages = {311--318},
}

@article{adjeisah_pseudotext_2021,
	title = {Pseudotext {Injection} and {Advance} {Filtering} of {Low}-{Resource} {Corpus} for {Neural} {Machine} {Translation}},
	volume = {2021},
	issn = {1687-5273, 1687-5265},
	url = {https://www.hindawi.com/journals/cin/2021/6682385/},
	doi = {10.1155/2021/6682385},
	language = {en},
	urldate = {2023-04-15},
	journal = {Computational Intelligence and Neuroscience},
	author = {Adjeisah, Michael and Liu, Guohua and Nyabuga, Douglas Omwenga and Nortey, Richard Nuetey and Song, Jinling},
	editor = {Yuan, Qiangqiang},
	month = apr,
	year = {2021},
	pages = {1--10},
}

@inproceedings{acheampong_language_2021,
	address = {Cham},
	series = {Advances in {Intelligent} {Systems} and {Computing}},
	title = {Language {Revitalization}: {A} {Benchmark} for {Akan}-to-{English} {Machine} {Translation}},
	isbn = {9783030551872},
	shorttitle = {Language {Revitalization}},
	doi = {10.1007/978-3-030-55187-2_20},
	language = {en},
	booktitle = {Intelligent {Systems} and {Applications}},
	publisher = {Springer International Publishing},
	author = {Acheampong, Kingsley Nketia and Sackey, Nathaniel Nii Oku},
	editor = {Arai, Kohei and Kapoor, Supriya and Bhatia, Rahul},
	year = {2021},
	pages = {231--244},
}

@misc{hacheme_english2gbe_2021,
	title = {{English2Gbe}: {A} multilingual machine translation model for \{{Fon}/{Ewe}\}{Gbe}},
	shorttitle = {{English2Gbe}},
	url = {http://arxiv.org/abs/2112.11482},
	doi = {10.48550/arXiv.2112.11482},
	urldate = {2023-04-15},
	publisher = {arXiv},
	author = {Hacheme, Gilles},
	month = dec,
	year = {2021},
	note = {arXiv:2112.11482 [cs]},
}

@misc{azunre_english-twi_2021,
	title = {English-{Twi} {Parallel} {Corpus} for {Machine} {Translation}},
	url = {http://arxiv.org/abs/2103.15625},
	doi = {10.48550/arXiv.2103.15625},
	urldate = {2023-04-10},
	publisher = {arXiv},
	author = {Azunre, Paul and Osei, Salomey and Addo, Salomey and Adu-Gyamfi, Lawrence Asamoah and Moore, Stephen and Adabankah, Bernard and Opoku, Bernard and Asare-Nyarko, Clara and Nyarko, Samuel and Amoaba, Cynthia and Appiah, Esther Dansoa and Akwerh, Felix and Lawson, Richard Nii Lante and Budu, Joel and Debrah, Emmanuel and Boateng, Nana and Ofori, Wisdom and Buabeng-Munkoh, Edwin and Adjei, Franklin and Ampomah, Isaac Kojo Essel and Otoo, Joseph and Borkor, Reindorf and Mensah, Standylove Birago and Mensah, Lucien and Marcel, Mark Amoako and Amponsah, Anokye Acheampong and Hayfron-Acquah, James Ben},
	month = apr,
	year = {2021},
	note = {arXiv:2103.15625 [cs]},
}

@inproceedings{agyei_akan-english_2021,
	title = {Akan-{English}: {Transformer} for {Low} {Resource} {Translation}},
	shorttitle = {Akan-{English}},
	doi = {10.1109/ICCWAMTIP53232.2021.9674076},
	booktitle = {2021 18th {International} {Computer} {Conference} on {Wavelet} {Active} {Media} {Technology} and {Information} {Processing} ({ICCWAMTIP})},
	author = {Agyei, Emmanuel and Zhang, Xiaoling and Yussif, Sophyani Banaamwini and Agbley, Bless Lord Y.},
	month = dec,
	year = {2021},
	note = {ISSN: 2576-8964},
	pages = {256--259},
}

@inproceedings{adelani2025irokobenchnewbenchmarkafrican,
    title = "{I}roko{B}ench: A New Benchmark for {A}frican Languages in the Age of Large Language Models",
    author = "Adelani, David Ifeoluwa  and
      Ojo, Jessica  and
      Azime, Israel Abebe  and
      Zhuang, Jian Yun  and
      Alabi, Jesujoba Oluwadara  and
      He, Xuanli  and
      Ochieng, Millicent  and
      Hooker, Sara  and
      Bukula, Andiswa  and
      Lee, En-Shiun Annie  and
      Chukwuneke, Chiamaka Ijeoma  and
      Buzaaba, Happy  and
      Sibanda, Blessing Kudzaishe  and
      Kalipe, Godson Koffi  and
      Mukiibi, Jonathan  and
      Kabongo Kabenamualu, Salomon  and
      Yuehgoh, Foutse  and
      Setaka, Mmasibidi  and
      Ndolela, Lolwethu  and
      Odu, Nkiruka  and
      Mabuya, Rooweither  and
      Osei, Salomey  and
      Muhammad, Shamsuddeen Hassan  and
      Samb, Sokhar  and
      Guge, Tadesse Kebede  and
      Sherman, Tombekai Vangoni  and
      Stenetorp, Pontus",
    editor = "Chiruzzo, Luis  and
      Ritter, Alan  and
      Wang, Lu",
    booktitle = "Proceedings of the 2025 Conference of the Nations of the Americas Chapter of the Association for Computational Linguistics: Human Language Technologies (Volume 1: Long Papers)",
    month = apr,
    year = "2025",
    address = "Albuquerque, New Mexico",
    publisher = "Association for Computational Linguistics",
    url = "https://aclanthology.org/2025.naacl-long.139/",
    doi = "10.18653/v1/2025.naacl-long.139",
    pages = "2732--2757",
    ISBN = "979-8-89176-189-6"
}

@inproceedings{meyer2022biblettslargehighfidelitymultilingual,
    title = "BibleTTS: a large, high-fidelity, multilingual, and uniquely African speech corpus",
    author = "Josh Meyer and Adelani, \{David Ifeoluwa\} and Edresson Casanova and Alp {\"O}ktem and Daniel Whitenack and Julian Weber and Salomon Kabongo and Elizabeth Salesky and Iroro Orife and Colin Leong and Perez Ogayo and Chris Emezue and Jonathan Mukiibi and Salomey Osei and Apelete Agbolo and Victor Akinode and Bernard Opoku and Samuel Olanrewaju and Jesujoba Alabi and Shamsuddeen Muhammad",
    note = "Publisher Copyright: Copyright {\textcopyright} 2022 ISCA.; 23rd Annual Conference of the International Speech Communication Association, INTERSPEECH 2022 ; Conference date: 18-09-2022 Through 22-09-2022",
    year = "2022",
    month = sep,
    day = "18",
    doi = "10.21437/interspeech.2022-10850",
    language = "English",
    series = "Proceedings of the Annual Conference of the International Speech Communication Association, INTERSPEECH",
    publisher = "International Speech Communication Association",
    pages = "2383--2387",
    booktitle = "Interspeech 2022",
    }

@inproceedings{adebara2023serengetimassivelymultilinguallanguage,
    title = "{SERENGETI}: Massively Multilingual Language Models for {A}frica",
    author = "Adebara, Ife  and
      Elmadany, AbdelRahim  and
      Abdul-Mageed, Muhammad  and
      Alcoba Inciarte, Alcides",
    editor = "Rogers, Anna  and
      Boyd-Graber, Jordan  and
      Okazaki, Naoaki",
    booktitle = "Findings of the Association for Computational Linguistics: ACL 2023",
    month = jul,
    year = "2023",
    address = "Toronto, Canada",
    publisher = "Association for Computational Linguistics",
    url = "https://aclanthology.org/2023.findings-acl.97/",
    doi = "10.18653/v1/2023.findings-acl.97",
    pages = "1498--1537"
}

@inproceedings{adebara2024cheetahnaturallanguagegeneration,
    title = "Cheetah: Natural Language Generation for 517 {A}frican Languages",
    author = "Adebara, Ife  and
      Elmadany, AbdelRahim  and
      Abdul-Mageed, Muhammad",
    editor = "Ku, Lun-Wei  and
      Martins, Andre  and
      Srikumar, Vivek",
    booktitle = "Proceedings of the 62nd Annual Meeting of the Association for Computational Linguistics (Volume 1: Long Papers)",
    month = aug,
    year = "2024",
    address = "Bangkok, Thailand",
    publisher = "Association for Computational Linguistics",
    url = "https://aclanthology.org/2024.acl-long.691/",
    doi = "10.18653/v1/2024.acl-long.691",
    pages = "12798--12823"
}

@inproceedings{adebara2025weevaluatingllmperformance,
    title = "Where Are We? Evaluating {LLM} Performance on {A}frican Languages",
    author = "Adebara, Ife  and
      Toyin, Hawau Olamide  and
      Ghebremichael, Nahom Tesfu  and
      Elmadany, AbdelRahim A.  and
      Abdul-Mageed, Muhammad",
    editor = "Che, Wanxiang  and
      Nabende, Joyce  and
      Shutova, Ekaterina  and
      Pilehvar, Mohammad Taher",
    booktitle = "Proceedings of the 63rd Annual Meeting of the Association for Computational Linguistics (Volume 1: Long Papers)",
    month = jul,
    year = "2025",
    address = "Vienna, Austria",
    publisher = "Association for Computational Linguistics",
    url = "https://aclanthology.org/2025.acl-long.1572/",
    doi = "10.18653/v1/2025.acl-long.1572",
    pages = "32704--32731",
    ISBN = "979-8-89176-251-0"
}

@inproceedings{li2025ssacometllmsoutperformlearned,
    title = "{SSA}-{COMET}: Do {LLM}s Outperform Learned Metrics in Evaluating {MT} for Under-Resourced {A}frican Languages?",
    author = "Li, Senyu  and
      Wang, Jiayi  and
      Ali, Felermino D. M. A.  and
      Cherry, Colin  and
      Deutsch, Daniel  and
      Briakou, Eleftheria  and
      Sousa-Silva, Rui  and
      Lopes Cardoso, Henrique  and
      Stenetorp, Pontus  and
      Adelani, David Ifeoluwa",
    editor = "Christodoulopoulos, Christos  and
      Chakraborty, Tanmoy  and
      Rose, Carolyn  and
      Peng, Violet",
    booktitle = "Proceedings of the 2025 Conference on Empirical Methods in Natural Language Processing",
    month = nov,
    year = "2025",
    address = "Suzhou, China",
    publisher = "Association for Computational Linguistics",
    url = "https://aclanthology.org/2025.emnlp-main.656/",
    doi = "10.18653/v1/2025.emnlp-main.656",
    pages = "12979--12998",
    ISBN = "979-8-89176-332-6"
}

@inproceedings{song2026smalllanguagemodelsilver,
    title = "Are Small Language Models the Silver Bullet to Low-Resource Languages Machine Translation?",
    author = "Song, Yewei  and
      Li, Lujun  and
      Lothritz, Cedric  and
      Ezzini, Saad  and
      Sleem, Lama  and
      Gentile, Niccolo'  and
      State, Radu  and
      Bissyand{\'e}, Tegawend{\'e} F.  and
      Klein, Jacques",
    editor = "Ojha, Atul Kr.  and
      Liu, Chao-hong  and
      Vylomova, Ekaterina  and
      Pirinen, Flammie  and
      Washington, Jonathan  and
      Oco, Nathaniel  and
      Zhao, Xiaobing",
    booktitle = "Proceedings for the Ninth Workshop on Technologies for Machine Translation of Low Resource Languages ({L}o{R}es{MT} 2026)",
    month = mar,
    year = "2026",
    address = "Rabat, Morocco",
    publisher = "Association for Computational Linguistics",
    url = "https://aclanthology.org/2026.loresmt-1.1/",
    doi = "10.18653/v1/2026.loresmt-1.1",
    pages = "1--26",
    ISBN = "979-8-89176-366-1"
}

@misc{issaka2026proxy,
      title={Translation as a Scalable Proxy for Multilingual Evaluation}, 
      author={Sheriff Issaka and Erick Rosas Gonzalez and Lieqi Liu and Evans Kofi Agyei and Lucas Bandarkar and Nanyun Peng and David Ifeoluwa Adelani and Francisco Guzmán and Saadia Gabriel},
      year={2026},
      eprint={2601.11778},
      archivePrefix={arXiv},
      primaryClass={cs.CL},
      url={https://arxiv.org/abs/2601.11778}, 
}

@article{qin2025survey,
    title = {A survey of multilingual large language models},
    journal = {Patterns},
    volume = {6},
    number = {1},
    pages = {101118},
    year = {2025},
    issn = {2666-3899},
    doi = {https://doi.org/10.1016/j.patter.2024.101118},
    url = {https://www.sciencedirect.com/science/article/pii/S2666389924002903},
    author = {Libo Qin and Qiguang Chen and Yuhang Zhou and Zhi Chen and Yinghui Li and Lizi Liao and Min Li and Wanxiang Che and Philip S. Yu}
}

@inproceedings{adebara2022afrocentricnlpafricanlanguages,
    title = "Towards Afrocentric {NLP} for {A}frican Languages: Where We Are and Where We Can Go",
    author = "Adebara, Ife  and
      Abdul-Mageed, Muhammad",
    editor = "Muresan, Smaranda  and
      Nakov, Preslav  and
      Villavicencio, Aline",
    booktitle = "Proceedings of the 60th Annual Meeting of the Association for Computational Linguistics (Volume 1: Long Papers)",
    month = may,
    year = "2022",
    address = "Dublin, Ireland",
    publisher = "Association for Computational Linguistics",
    url = "https://aclanthology.org/2022.acl-long.265/",
    doi = "10.18653/v1/2022.acl-long.265",
    pages = "3814--3841"
}

@inproceedings{alabi2025chartinglandscapeafricannlp,
    title = "Charting the Landscape of {A}frican {NLP}: Mapping Progress and Shaping the Road Ahead",
    author = "Alabi, Jesujoba Oluwadara  and
      Hedderich, Michael A.  and
      Adelani, David Ifeoluwa  and
      Klakow, Dietrich",
    editor = "Christodoulopoulos, Christos  and
      Chakraborty, Tanmoy  and
      Rose, Carolyn  and
      Peng, Violet",
    booktitle = "Proceedings of the 2025 Conference on Empirical Methods in Natural Language Processing",
    month = nov,
    year = "2025",
    address = "Suzhou, China",
    publisher = "Association for Computational Linguistics",
    url = "https://aclanthology.org/2025.emnlp-main.1414/",
    doi = "10.18653/v1/2025.emnlp-main.1414",
    pages = "27807--27841",
    ISBN = "979-8-89176-332-6"
}

@article{matsinhe01012006,
    author = {Sozinho Matsinhe},
    title = {Language decline and death in Africa: Causes, consequences and challenges},
    journal = {Language Matters},
    volume = {37},
    number = {1},
    pages = {118--121},
    year = {2006},
    publisher = {Routledge},
    doi = {10.1080/10228190608566255},
    URL = {https://doi.org/10.1080/10228190608566255},
    eprint = {https://doi.org/10.1080/10228190608566255}
}

@article{lupke2009margin,
  author   = {L\"{u}pke, Friederike},
  title    = {At the Margin - African Endangered Languages in the Context of Global Endangerment Discourses},
  journal  = {African Research \& Documentation},
  year     = {2009},
  volume   = {109},
  pages    = {15--41},
  doi      = {10.1017/S0305862X00016472}
}

@incollection{lupke2019language,
  title={Language endangerment and language documentation in Africa},
  author={L{\"u}pke, Friederike},
  booktitle={The Cambridge handbook of African linguistics},
  pages={468--490},
  year={2019},
  publisher={Cambridge University Press}
}

@incollection{essegbey2020language,
    author = {Essegbey, James},
    editor = {Vossen, Rainer and Dimmendaal, Gerrit J.},
    isbn = {9780199609895},
    title = {Language Endangerment, Documentation, and Revitalization},
    booktitle = {The Oxford Handbook of African Languages},
    publisher = {Oxford University Press},
    year = {2020},
    month = {03},
    doi = {10.1093/oxfordhb/9780199609895.013.11},
    url = {https://doi.org/10.1093/oxfordhb/9780199609895.013.11},
    eprint = {https://academic.oup.com/book/0/chapter/334730300/chapter-ag-pdf/47589212/book_38608_section_334730300.ag.pdf},
}

@misc{nllbteam2022languageleftbehindscaling,
      title={No Language Left Behind: Scaling Human-Centered Machine Translation}, 
      author={NLLB Team and Marta R. Costa-jussà and James Cross and Onur Çelebi and Maha Elbayad and Kenneth Heafield and Kevin Heffernan and Elahe Kalbassi and Janice Lam and Daniel Licht and Jean Maillard and Anna Sun and Skyler Wang and Guillaume Wenzek and Al Youngblood and Bapi Akula and Loic Barrault and Gabriel Mejia Gonzalez and Prangthip Hansanti and John Hoffman and Semarley Jarrett and Kaushik Ram Sadagopan and Dirk Rowe and Shannon Spruit and Chau Tran and Pierre Andrews and Necip Fazil Ayan and Shruti Bhosale and Sergey Edunov and Angela Fan and Cynthia Gao and Vedanuj Goswami and Francisco Guzmán and Philipp Koehn and Alexandre Mourachko and Christophe Ropers and Safiyyah Saleem and Holger Schwenk and Jeff Wang},
      year={2022},
      eprint={2207.04672},
      archivePrefix={arXiv},
      primaryClass={cs.CL},
      url={https://arxiv.org/abs/2207.04672}, 
}

@inproceedings{singh2025globalmmluunderstandingaddressing,
    title = "Global {MMLU}: Understanding and Addressing Cultural and Linguistic Biases in Multilingual Evaluation",
    author = "Singh, Shivalika  and
      Romanou, Angelika  and
      Fourrier, Cl{\'e}mentine  and
      Adelani, David Ifeoluwa  and
      Ngui, Jian Gang  and
      Vila-Suero, Daniel  and
      Limkonchotiwat, Peerat  and
      Marchisio, Kelly  and
      Leong, Wei Qi  and
      Susanto, Yosephine  and
      Ng, Raymond  and
      Longpre, Shayne  and
      Ruder, Sebastian  and
      Ko, Wei-Yin  and
      Bosselut, Antoine  and
      Oh, Alice  and
      Martins, Andre  and
      Choshen, Leshem  and
      Ippolito, Daphne  and
      Ferrante, Enzo  and
      Fadaee, Marzieh  and
      Ermis, Beyza  and
      Hooker, Sara",
    editor = "Che, Wanxiang  and
      Nabende, Joyce  and
      Shutova, Ekaterina  and
      Pilehvar, Mohammad Taher",
    booktitle = "Proceedings of the 63rd Annual Meeting of the Association for Computational Linguistics (Volume 1: Long Papers)",
    month = jul,
    year = "2025",
    address = "Vienna, Austria",
    publisher = "Association for Computational Linguistics",
    url = "https://aclanthology.org/2025.acl-long.919/",
    doi = "10.18653/v1/2025.acl-long.919",
    pages = "18761--18799",
    ISBN = "979-8-89176-251-0"
}

@inproceedings{Bandarkar_2024,
    title = "The Belebele Benchmark: a Parallel Reading Comprehension Dataset in 122 Language Variants",
    author = "Bandarkar, Lucas  and
      Liang, Davis  and
      Muller, Benjamin  and
      Artetxe, Mikel  and
      Shukla, Satya Narayan  and
      Husa, Donald  and
      Goyal, Naman  and
      Krishnan, Abhinandan  and
      Zettlemoyer, Luke  and
      Khabsa, Madian",
    editor = "Ku, Lun-Wei  and
      Martins, Andre  and
      Srikumar, Vivek",
    booktitle = "Proceedings of the 62nd Annual Meeting of the Association for Computational Linguistics (Volume 1: Long Papers)",
    month = aug,
    year = "2024",
    address = "Bangkok, Thailand",
    publisher = "Association for Computational Linguistics",
    url = "https://aclanthology.org/2024.acl-long.44/",
    doi = "10.18653/v1/2024.acl-long.44",
    pages = "749--775"
}

@inproceedings{issaka-etal-2026-african,
    title = "The {A}frican Languages Lab: A Collaborative Approach to Advancing Low-Resource {A}frican {NLP}",
    author = "Issaka, Sheriff  and
      Wang, Keyi  and
      Ajibola, Yinka  and
      Samuel-Ipaye, Oluwatumininu  and
      Zhang, Zhaoyi  and
      Jimenez, Nicte Aguillon  and
      Agyei, Evans Kofi  and
      Lin, Abraham  and
      Ramachandran, Rohan  and
      Mumin, Sadick Abdul  and
      Nchifor, Faith  and
      Issah, Mohammed Shuraim  and
      Gonzalez, Erick Rosas  and
      Liu, Lieqi  and
      Kpei, Sylvester  and
      Osei, Jemimah Kusi  and
      Ajeneza, Carlene  and
      Boateng, Persis  and
      Yeboah, Prisca Adwoa Dufie  and
      Gabriel, Saadia",
    editor = "Liakata, Maria  and
      Moreira, Viviane P.  and
      Zhang, Jiajun  and
      Jurgens, David",
    booktitle = "Proceedings of the 64th Annual Meeting of the {A}ssociation for {C}omputational {L}inguistics (Volume 1: Long Papers)",
    month = jul,
    year = "2026",
    address = "San Diego, California, United States",
    publisher = "Association for Computational Linguistics",
    url = "https://aclanthology.org/2026.acl-long.1965/",
    doi = "10.18653/v1/2026.acl-long.1965",
    pages = "42460--42477",
    ISBN = "979-8-89176-390-6"
}

@misc{hurst2024gpt,
      title={GPT-4o System Card}, 
      author={OpenAI and : and Aaron Hurst and Adam Lerer and Adam P. Goucher and Adam Perelman and Aditya Ramesh and Aidan Clark and AJ Ostrow and Akila Welihinda and Alan Hayes and Alec Radford and Aleksander Mądry and Alex Baker-Whitcomb and Alex Beutel and Alex Borzunov and Alex Carney and Alex Chow and Alex Kirillov and Alex Nichol and Alex Paino and Alex Renzin and Alex Tachard Passos and Alexander Kirillov and Alexi Christakis and Alexis Conneau and Ali Kamali and Allan Jabri and Allison Moyer and Allison Tam and Amadou Crookes and Amin Tootoochian and Amin Tootoonchian and Ananya Kumar and Andrea Vallone and Andrej Karpathy and Andrew Braunstein and Andrew Cann and Andrew Codispoti and Andrew Galu and Andrew Kondrich and Andrew Tulloch and Andrey Mishchenko and Angela Baek and Angela Jiang and Antoine Pelisse and Antonia Woodford and Anuj Gosalia and Arka Dhar and Ashley Pantuliano and Avi Nayak and Avital Oliver and Barret Zoph and Behrooz Ghorbani and Ben Leimberger and Ben Rossen and Ben Sokolowsky and Ben Wang and Benjamin Zweig and Beth Hoover and Blake Samic and Bob McGrew and Bobby Spero and Bogo Giertler and Bowen Cheng and Brad Lightcap and Brandon Walkin and Brendan Quinn and Brian Guarraci and Brian Hsu and Bright Kellogg and Brydon Eastman and Camillo Lugaresi and Carroll Wainwright and Cary Bassin and Cary Hudson and Casey Chu and Chad Nelson and Chak Li and Chan Jun Shern and Channing Conger and Charlotte Barette and Chelsea Voss and Chen Ding and Cheng Lu and Chong Zhang and Chris Beaumont and Chris Hallacy and Chris Koch and Christian Gibson and Christina Kim and Christine Choi and Christine McLeavey and Christopher Hesse and Claudia Fischer and Clemens Winter and Coley Czarnecki and Colin Jarvis and Colin Wei and Constantin Koumouzelis and Dane Sherburn and Daniel Kappler and Daniel Levin and Daniel Levy and David Carr and David Farhi and David Mely and David Robinson and David Sasaki and Denny Jin and Dev Valladares and Dimitris Tsipras and Doug Li and Duc Phong Nguyen and Duncan Findlay and Edede Oiwoh and Edmund Wong and Ehsan Asdar and Elizabeth Proehl and Elizabeth Yang and Eric Antonow and Eric Kramer and Eric Peterson and Eric Sigler and Eric Wallace and Eugene Brevdo and Evan Mays and Farzad Khorasani and Felipe Petroski Such and Filippo Raso and Francis Zhang and Fred von Lohmann and Freddie Sulit and Gabriel Goh and Gene Oden and Geoff Salmon and Giulio Starace and Greg Brockman and Hadi Salman and Haiming Bao and Haitang Hu and Hannah Wong and Haoyu Wang and Heather Schmidt and Heather Whitney and Heewoo Jun and Hendrik Kirchner and Henrique Ponde de Oliveira Pinto and Hongyu Ren and Huiwen Chang and Hyung Won Chung and Ian Kivlichan and Ian O'Connell and Ian O'Connell and Ian Osband and Ian Silber and Ian Sohl and Ibrahim Okuyucu and Ikai Lan and Ilya Kostrikov and Ilya Sutskever and Ingmar Kanitscheider and Ishaan Gulrajani and Jacob Coxon and Jacob Menick and Jakub Pachocki and James Aung and James Betker and James Crooks and James Lennon and Jamie Kiros and Jan Leike and Jane Park and Jason Kwon and Jason Phang and Jason Teplitz and Jason Wei and Jason Wolfe and Jay Chen and Jeff Harris and Jenia Varavva and Jessica Gan Lee and Jessica Shieh and Ji Lin and Jiahui Yu and Jiayi Weng and Jie Tang and Jieqi Yu and Joanne Jang and Joaquin Quinonero Candela and Joe Beutler and Joe Landers and Joel Parish and Johannes Heidecke and John Schulman and Jonathan Lachman and Jonathan McKay and Jonathan Uesato and Jonathan Ward and Jong Wook Kim and Joost Huizinga and Jordan Sitkin and Jos Kraaijeveld and Josh Gross and Josh Kaplan and Josh Snyder and Joshua Achiam and Joy Jiao and Joyce Lee and Juntang Zhuang and Justyn Harriman and Kai Fricke and Kai Hayashi and Karan Singhal and Katy Shi and Kavin Karthik and Kayla Wood and Kendra Rimbach and Kenny Hsu and Kenny Nguyen and Keren Gu-Lemberg and Kevin Button and Kevin Liu and Kiel Howe and Krithika Muthukumar and Kyle Luther and Lama Ahmad and Larry Kai and Lauren Itow and Lauren Workman and Leher Pathak and Leo Chen and Li Jing and Lia Guy and Liam Fedus and Liang Zhou and Lien Mamitsuka and Lilian Weng and Lindsay McCallum and Lindsey Held and Long Ouyang and Louis Feuvrier and Lu Zhang and Lukas Kondraciuk and Lukasz Kaiser and Luke Hewitt and Luke Metz and Lyric Doshi and Mada Aflak and Maddie Simens and Madelaine Boyd and Madeleine Thompson and Marat Dukhan and Mark Chen and Mark Gray and Mark Hudnall and Marvin Zhang and Marwan Aljubeh and Mateusz Litwin and Matthew Zeng and Max Johnson and Maya Shetty and Mayank Gupta and Meghan Shah and Mehmet Yatbaz and Meng Jia Yang and Mengchao Zhong and Mia Glaese and Mianna Chen and Michael Janner and Michael Lampe and Michael Petrov and Michael Wu and Michele Wang and Michelle Fradin and Michelle Pokrass and Miguel Castro and Miguel Oom Temudo de Castro and Mikhail Pavlov and Miles Brundage and Miles Wang and Minal Khan and Mira Murati and Mo Bavarian and Molly Lin and Murat Yesildal and Nacho Soto and Natalia Gimelshein and Natalie Cone and Natalie Staudacher and Natalie Summers and Natan LaFontaine and Neil Chowdhury and Nick Ryder and Nick Stathas and Nick Turley and Nik Tezak and Niko Felix and Nithanth Kudige and Nitish Keskar and Noah Deutsch and Noel Bundick and Nora Puckett and Ofir Nachum and Ola Okelola and Oleg Boiko and Oleg Murk and Oliver Jaffe and Olivia Watkins and Olivier Godement and Owen Campbell-Moore and Patrick Chao and Paul McMillan and Pavel Belov and Peng Su and Peter Bak and Peter Bakkum and Peter Deng and Peter Dolan and Peter Hoeschele and Peter Welinder and Phil Tillet and Philip Pronin and Philippe Tillet and Prafulla Dhariwal and Qiming Yuan and Rachel Dias and Rachel Lim and Rahul Arora and Rajan Troll and Randall Lin and Rapha Gontijo Lopes and Raul Puri and Reah Miyara and Reimar Leike and Renaud Gaubert and Reza Zamani and Ricky Wang and Rob Donnelly and Rob Honsby and Rocky Smith and Rohan Sahai and Rohit Ramchandani and Romain Huet and Rory Carmichael and Rowan Zellers and Roy Chen and Ruby Chen and Ruslan Nigmatullin and Ryan Cheu and Saachi Jain and Sam Altman and Sam Schoenholz and Sam Toizer and Samuel Miserendino and Sandhini Agarwal and Sara Culver and Scott Ethersmith and Scott Gray and Sean Grove and Sean Metzger and Shamez Hermani and Shantanu Jain and Shengjia Zhao and Sherwin Wu and Shino Jomoto and Shirong Wu and Shuaiqi and Xia and Sonia Phene and Spencer Papay and Srinivas Narayanan and Steve Coffey and Steve Lee and Stewart Hall and Suchir Balaji and Tal Broda and Tal Stramer and Tao Xu and Tarun Gogineni and Taya Christianson and Ted Sanders and Tejal Patwardhan and Thomas Cunninghman and Thomas Degry and Thomas Dimson and Thomas Raoux and Thomas Shadwell and Tianhao Zheng and Todd Underwood and Todor Markov and Toki Sherbakov and Tom Rubin and Tom Stasi and Tomer Kaftan and Tristan Heywood and Troy Peterson and Tyce Walters and Tyna Eloundou and Valerie Qi and Veit Moeller and Vinnie Monaco and Vishal Kuo and Vlad Fomenko and Wayne Chang and Weiyi Zheng and Wenda Zhou and Wesam Manassra and Will Sheu and Wojciech Zaremba and Yash Patil and Yilei Qian and Yongjik Kim and Youlong Cheng and Yu Zhang and Yuchen He and Yuchen Zhang and Yujia Jin and Yunxing Dai and Yury Malkov},
      year={2024},
      eprint={2410.21276},
      archivePrefix={arXiv},
      primaryClass={cs.CL},
      url={https://arxiv.org/abs/2410.21276}, 
}

@inproceedings{alabi2025charting,
    title = "Charting the Landscape of {A}frican {NLP}: Mapping Progress and Shaping the Road Ahead",
    author = "Alabi, Jesujoba Oluwadara  and
      Hedderich, Michael A.  and
      Adelani, David Ifeoluwa  and
      Klakow, Dietrich",
    editor = "Christodoulopoulos, Christos  and
      Chakraborty, Tanmoy  and
      Rose, Carolyn  and
      Peng, Violet",
    booktitle = "Proceedings of the 2025 Conference on Empirical Methods in Natural Language Processing",
    month = nov,
    year = "2025",
    address = "Suzhou, China",
    publisher = "Association for Computational Linguistics",
    url = "https://aclanthology.org/2025.emnlp-main.1414/",
    doi = "10.18653/v1/2025.emnlp-main.1414",
    pages = "27807--27841",
    ISBN = "979-8-89176-332-6"
}

@inproceedings{chen2026linguistically,
    title = "Linguistically Informed Evaluation of Multilingual {ASR} for {A}frican Languages",
    author = "Chen, Fei-Yueh  and
      Adeleke, Lateef  and
      Downey, C. M.",
    editor = "Chimoto, Everlyn Asiko  and
      Lignos, Constantine  and
      Muhammad, Shamsuddeen  and
      Abdulmumin, Idris  and
      Siro, Clemencia  and
      Adelani, David Ifeoluwa",
    booktitle = "Proceedings of the 7th Workshop on {A}frican Natural Language Processing ({A}frica{NLP} 2026)",
    month = mar,
    year = "2026",
    address = "Rabat, Morocco",
    publisher = "Association for Computational Linguistics",
    url = "https://aclanthology.org/2026.africanlp-main.14/",
    doi = "10.18653/v1/2026.africanlp-main.14",
    pages = "149--162",
    ISBN = "979-8-89176-364-7"
}

@misc{mbaye2026opportunities,
      title={Opportunities and Challenges of Natural Language Processing for Low-Resource Senegalese Languages in Social Science Research}, 
      author={Derguene Mbaye and Tatiana D. P. Mbengue and Madoune R. Seye and Moussa Diallo and Mamadou L. Ndiaye and Dimitri S. Adjanohoun and Cheikh S. Wade and Djiby Sow and Jean-Claude B. Munyaka and Jerome Chenal},
      year={2025},
      eprint={2601.09716},
      archivePrefix={arXiv},
      primaryClass={cs.CL},
      url={https://arxiv.org/abs/2601.09716}, 
}

@inproceedings{uemura2026afrimteb,
    title = "{A}fri{MTEB} and {A}fri{E}5: Benchmarking and Adapting Text Embedding Models for {A}frican Languages",
    author = "Uemura, Kosei  and
      Zhang, Miaoran  and
      Adelani, David Ifeoluwa",
    editor = "Demberg, Vera  and
      Inui, Kentaro  and
      Marquez, Llu{\'i}s",
    booktitle = "Proceedings of the 19th Conference of the {E}uropean Chapter of the {A}ssociation for {C}omputational {L}inguistics (Volume 1: Long Papers)",
    month = mar,
    year = "2026",
    address = "Rabat, Morocco",
    publisher = "Association for Computational Linguistics",
    url = "https://aclanthology.org/2026.eacl-long.171/",
    doi = "10.18653/v1/2026.eacl-long.171",
    pages = "3697--3717",
    ISBN = "979-8-89176-380-7"
}

@article{waliya2026metadata,
	doi = {10.20944/preprints202604.0360.v1},
	url = {https://doi.org/10.20944/preprints202604.0360.v1},
	year = 2026,
	month = {April},
	publisher = {Preprints},
	author = {Yohanna Joseph Waliya and Margaret Mary Okon},
	title = {Metadata Analysis of the Generative AI Usefulness for African Languages},
	journal = {Preprints}
}

@inproceedings{blasi-etal-2022-systematic,
    title = "Systematic Inequalities in Language Technology Performance across the World{'}s Languages",
    author = "Blasi, Damian  and
      Anastasopoulos, Antonios  and
      Neubig, Graham",
    editor = "Muresan, Smaranda  and
      Nakov, Preslav  and
      Villavicencio, Aline",
    booktitle = "Proceedings of the 60th Annual Meeting of the Association for Computational Linguistics (Volume 1: Long Papers)",
    month = may,
    year = "2022",
    address = "Dublin, Ireland",
    publisher = "Association for Computational Linguistics",
    url = "https://aclanthology.org/2022.acl-long.376/",
    doi = "10.18653/v1/2022.acl-long.376",
    pages = "5486--5505"
}

@misc{WiafeUGSpeechDataRepo,
  author       = {Wiafe, Isaac and Abdulai, Jamal-Deen and Ekpezu, Akon Obu and Helegah, Raynard Dodzi and Atsakpo, Elikem Doe and Nutrokpor, Charles and Winful, Fiifi Baffoe Payin and Solaga, Kafui Kwashia},
  title        = {{UGSpeechData: A Multilingual Speech Dataset of Ghanaian Languages}},
  year         = 2026,
  month        = feb,
  publisher    = {Science Data Bank},
  version      = {V7},
  doi          = {10.57760/sciencedb.22298},
  url          = https://doi.org/10.57760/sciencedb.22298
}

@article{WiafeUGSpeechDataPaper,
    title = {Advancing automatic speech recognition for low-resource ghanaian languages: Audio datasets for Akan, Ewe, Dagbani, Dagaare, and Ikposo},
    journal = {Data in Brief},
    volume = {61},
    pages = {111880},
    year = {2025},
    issn = {2352-3409},
    doi = {https://doi.org/10.1016/j.dib.2025.111880},
    url = {https://www.sciencedirect.com/science/article/pii/S2352340925006043},
    author = {Isaac Wiafe and Jamal-Deen Abdulai and Akon Obu Ekpezu and Raynard Dodzi Helegah and Elikem Doe Atsakpo and Charles Nutrokpor and Fiifi Baffoe {Payin Winful} and Kafui Kwashie Solaga}
}

@misc{moore2026nsankuevaluatingzeroshottranslation,
      title={Nsanku: Evaluating Zero-Shot Translation Performance of LLMs for Ghanaian Languages}, 
      author={Stephen E. Moore and Mich-Seth Owusu and Akwasi Asare and Lawrence Adu Gyamfi and Paul Azunre and Joel Budu and Jonathan Asiamah and Elias Dzobo and Kelvin Newman and Edmund O. Benefo and Gerhardt Datsomor and Onesimus Addo Appiah and Ama Branoa Banful and Lucas Woedem Kpatah and Saani Mustapha Deishini and John Ayernor},
      year={2026},
      eprint={2605.04208},
      archivePrefix={arXiv},
      primaryClass={cs.CL},
      url={https://arxiv.org/abs/2605.04208}, 
}

@misc{salamanca2026tinyayabridgingscale,
      title={Tiny Aya: Bridging Scale and Multilingual Depth}, 
      author={Alejandro R. Salamanca and Diana Abagyan and Daniel D'souza and Ammar Khairi and David Mora and Saurabh Dash and Viraat Aryabumi and Sara Rajaee and Mehrnaz Mofakhami and Ananya Sahu and Thomas Euyang and Brittawnya Prince and Madeline Smith and Hangyu Lin and Acyr Locatelli and Sara Hooker and Tom Kocmi and Aidan Gomez and Ivan Zhang and Phil Blunsom and Nick Frosst and Joelle Pineau and Beyza Ermis and Ahmet Üstün and Julia Kreutzer and Marzieh Fadaee},
      year={2026},
      eprint={2603.11510},
      archivePrefix={arXiv},
      primaryClass={cs.CL},
      url={https://arxiv.org/abs/2603.11510}, 
}

@misc{awagptv1_2025,
  title={N-ATLaS-LLM: A Multilingual African Language Model},
  author={{Awarri Technologies and National Information Technology and Development Agency}},
  year={2025},
  publisher={Hugging Face},
  note={Fine-tuned Llama-3 8B model for African languages developed in collaboration with the Federal Government of Nigeria}
}

@software{anthropic2026opus5,
  author = {{Anthropic}},
  title = {Claude Opus 5},
  year = {2026},
  url = {https://claude.ai},
  version = {Opus 5},
  type = {Large language model}
}

@software{google2026gemini31pro,
  author = {{Google}},
  title = {Gemini 3.1 Pro},
  year = {2026},
  url = {https://gemini.google.com},
  version = {3.1 Pro},
  type = {Large language model}
}

@article{min2023recent,
    author = {Min, Bonan and Ross, Hayley and Sulem, Elior and Veyseh, Amir Pouran Ben and Nguyen, Thien Huu and Sainz, Oscar and Agirre, Eneko and Heintz, Ilana and Roth, Dan},
    title = {Recent Advances in Natural Language Processing via Large Pre-trained Language Models: A Survey},
    year = {2023},
    issue_date = {February 2024},
    publisher = {Association for Computing Machinery},
    address = {New York, NY, USA},
    volume = {56},
    number = {2},
    issn = {0360-0300},
    url = {https://doi.org/10.1145/3605943},
    doi = {10.1145/3605943},
    journal = {ACM Comput. Surv.},
    month = sep,
    articleno = {30},
    numpages = {40}
}

@inproceedings{nguefack-etal-2025-pretraining,
    title = "Pretraining Strategies using Monolingual and Parallel Data for Low-Resource Machine Translation",
    author = "Nguefack, Idriss Nguepi  and
      Finkelstein, Mara  and
      Sakayo, Toadoum Sari",
    editor = "Lignos, Constantine  and
      Abdulmumin, Idris  and
      Adelani, David",
    booktitle = "Proceedings of the Sixth Workshop on African Natural Language Processing (AfricaNLP 2025)",
    month = jul,
    year = "2025",
    address = "Vienna, Austria",
    publisher = "Association for Computational Linguistics",
    url = "https://aclanthology.org/2025.africanlp-1.6/",
    doi = "10.18653/v1/2025.africanlp-1.6",
    pages = "31--38",
    ISBN = "979-8-89176-257-2"
}

@inproceedings{kargaran-etal-2023-glotlid,
    title = "{G}lot{LID}: Language Identification for Low-Resource Languages",
    author = "Kargaran, Amir Hossein  and
      Imani, Ayyoob  and
      Yvon, Fran{\c{c}}ois  and
      Schuetze, Hinrich",
    editor = "Bouamor, Houda  and
      Pino, Juan  and
      Bali, Kalika",
    booktitle = "Findings of the Association for Computational Linguistics: EMNLP 2023",
    month = dec,
    year = "2023",
    address = "Singapore",
    publisher = "Association for Computational Linguistics",
    url = "https://aclanthology.org/2023.findings-emnlp.410/",
    doi = "10.18653/v1/2023.findings-emnlp.410",
    pages = "6155--6218"
}

@inproceedings{bremang-etal-2026-creating,
  title = {Creating Task-Specific Speech Recognition Datasets from Scratch for Low-Resource Languages: Assessing the Impact of Token Sequence Overlap},
  author = {Bremang, Adwoa Asantewaa and Owusu, Dennis Asamoah and Quagraine, Victor and Annor-Adjaye, Leanne M.M.},
  booktitle = {Proceedings of the Fifteenth Language Resources and Evaluation Conference (LREC 2026)},
  month = {May},
  year = {2026},
  pages = {3075--3082},
  address = {Palma, Mallorca, Spain},
  publisher = {European Language Resources Association (ELRA)},
  editor = {Piperidis, Stelios and Bel, Núria and van den Heuvel, Henk and Ide, Nancy and Krek, Simon and Toral, Antonio},
  doi = {10.63317/3myb33sgskfb}
}

@inproceedings{tonja-etal-2026-afri,
    title = "Afri-{MCQA}: Multimodal Cultural Question Answering for {A}frican Languages",
    author = "Tonja, Atnafu Lambebo  and
      Anand, Srija  and
      Villa-Cueva, Emilio  and
      Azime, Israel Abebe  and
      Alabi, Jesujoba Oluwadara  and
      Mohamed, Muhidin A.  and
      Yadeta, Debela Desalegn  and
      Abadi, Negasi Haile  and
      Oppong, Abigail  and
      Obiefuna, Nnaemeka Casmir  and
      Abdulmumin, Idris  and
      Etori, Naome A  and
      Wairagala, Eric Peter  and
      Tshinu, Kanda Patrick  and
      Emmanuel, Imanigirimbabazi  and
      Malema, Gabofetswe  and
      Aji, Alham Fikri  and
      Adelani, David Ifeoluwa  and
      Solorio, Thamar",
    editor = "Liakata, Maria  and
      Moreira, Viviane P.  and
      Zhang, Jiajun  and
      Jurgens, David",
    booktitle = "Proceedings of the 64th Annual Meeting of the {A}ssociation for {C}omputational {L}inguistics (Volume 1: Long Papers)",
    month = jul,
    year = "2026",
    address = "San Diego, California, United States",
    publisher = "Association for Computational Linguistics",
    url = "https://aclanthology.org/2026.acl-long.1869/",
    doi = "10.18653/v1/2026.acl-long.1869",
    pages = "40249--40282",
    ISBN = "979-8-89176-390-6"
}

@misc{glottolog5_3,
    author       = {Hammarstr{\"o}m, Harald and Forkel, Robert and Haspelmath, Martin and Bank, Sebastian},
    title        = {Glottolog 5.3},
    year         = {2026},
    address      = {Leipzig},
    publisher    = {Max Planck Institute for Evolutionary Anthropology},
    url          = {https://glottolog.org},
    doi          = {10.5281/zenodo.18840935},
    note         = {Available online at \url{http://glottolog.org}}
}

\clearpage
\appendix

\appendix
\renewcommand{\thetable}{A\arabic{table}}
\renewcommand{\thefigure}{A\arabic{figure}}
\setcounter{table}{0}
\setcounter{figure}{0}

\section{Appendix}
\label{sec:appendix}
This appendix expands three parts of the main text that space required us to compress there: the model configurations behind Table \ref{tab:summary-SpBLEU-rouge-wer-f1} (\ref{app:model-config}), the evaluation metrics reported in Tables \ref{tab:summary-bleu-chrf-ter} and \ref{tab:summary-SpBLEU-rouge-wer-f1} (\ref{app:model-eval}), and the four MT paradigms referenced in Section \ref{sec:nlp-and-mt-in-lrls-revitalization} (\ref{app:mt-paradigms}).

\subsection{Detailed Model Configuration and Training}
\label{app:model-config}
Section \ref{sec:survery-results--model-architectures} grouped the reviewed modeling papers into four paradigms. We record the configuration and training details for each below, in that same order, for readers who want to reproduce or extend a particular system.

\textbf{Train-from-scratch Architectures.} The earliest modeling work in our review trained models directly on task-specific parallel data. \citealp{agyei_akan-english_2021}preprocessed input with an embedding layer and positional encoding, utilizing a 6-layer encoder with 2-head Multi-Head Attention (MHA) and a feed-forward layer of 300 units. The modified Transformer encoder was trained on 26,000 sentences (85\% train, 10\% validation, 5\% test) for 100 epochs using the Adam optimizer with a learning rate of 0.0002 and a batch size of 16.

Working at a comparable scale, \citealp{acheampong_language_2021} trained a BPE tokenizer with a vocabulary size of 10,000, using 300-dimensional source and target embeddings and a hidden layer size of 512. The encoder-decoder was fine-tuned using the Adam optimizer with a learning rate of 0.0002 and regularization techniques to reduce overfitting.

\citealp{hacheme_english2gbe_2021} extended this setting to a multilingual configuration, training BPE tokenizers with vocabulary sizes of 4,000 and 10,000 for Fon and Ewe, respectively, and 10,000 and 6,000 for English and Gbe in the multilingual model. The base Transformer had a 6-layer encoder/decoder with an embedding size of 512, a hidden size of 512, a feed-forward size of 2,048, and an 8-head MHA. While the bilingual Fon model was trained on 29,899 sentence pairs, the Ewe and multilingual models utilized a subset of 100,000 randomly selected sentence pairs. The models were trained for 30 epochs with a maximum sequence length of 150 and batch sizes of 300 (bilingual) and 400 (multilingual).

\citealp{adjeisah_pseudotext_2021} is the one paper in this group to compare architectures directly, employing a Neural Transformer with 4 encoder-decoder layers, a batch size of 64, a dropout rate of 0.5, and a weight constraint of 0.5. The OpenNMT BiLSTM model was trained with a batch size of 32, the Adadelta optimizer, and regularization techniques to prevent overfitting.

\textbf{Fine-tuning of Pretrained MT Models.} A second group starts from existing checkpoints rather than random initialization, which lowers the data requirement considerably. \citealp{azunre_english-twi_2021} and \citealp{gyasi_twi_2023}, 2023 both loaded the OPUS-MT model using the Hugging Face Transformers package. \citealp{azunre_english-twi_2021} trained the model on 25,421 sentence pairs and evaluated it on 697 crowd-sourced pairs, fine-tuning it using the Adam optimizer for 20 epochs. \citealp{gyasi_twi_2023} trained the model on 8,566 sentence pairs and validated it on 1,071 pairs, fine-tuning it using the Adam optimizer with a batch size of 8 and a learning rate of 0.00002.

\citealp{bannerman2024machine} also adopted a standard Transformer architecture, fine-tuning the OPUS-MT English-to-Twi model. Their setup consisted of a 6-layer encoder and decoder with 8 attention heads per layer, utilizing the Adam optimizer. They employed techniques like beam search with 4 beams and a temperature of 1.0 to enhance translation quality during inference.

\citealp{agyei2025enhancing,agyei2025dynamic} move to a different base model, utilizing T5-Small as their baseline, which features a 6-layer encoder and decoder, a hidden dimension of 512, and 8 attention heads, totaling approximately 60 million parameters. They employed the AdamW optimizer with a weight decay of 0.01 and applied gradient clipping, training on Google Colab Pro with NVIDIA Tesla A100 GPUs and the HuggingFace Transformers library for 100 epochs with dynamic batch weighting. A key innovation in their work is the Cross-Lingual Optimization Framework (CLOF), a federated learning approach that assigns higher aggregation weights to updates from low-resource (Twi) clients to prevent their linguistic signals from being diluted by high-resource languages; the authors report that dynamic gradient weighting improved SpBLEU by more than 10x compared to equal weighting approaches.

\textbf{Africa-centric Multilingual Pretraining.} The third group shifts the starting point itself, pretraining across African languages before any task-specific adaptation.\citealp{adebara2023serengetimassivelymultilinguallanguage} experiments with an XLM-R based architecture trained with a vocabulary size of 250k for 20 epochs, a batch size of 8 and a sentence length, and featuring 517 African languages. \citealp{adebara2024cheetahnaturallanguagegeneration} pretrained with encoder and decoder components similar in size and configuration to T5, with 12 layers each with 12 attention heads, and 768 hidden units for the base model, resulting in a model with 580 million parameters.

Two more recent systems apply this strategy at larger scale and to non-MT objectives. \citealp{li2025ssacometllmsoutperformlearned} employs a multilingual encoder AfroXLMR-76L with the open-source COMET codebase. STL and QE models training is conducted on an NVIDIA L40S GPU, and the MTL model is trained on an NVIDIA A100-SXM4-80GB GPU. A batch size of 16 with gradient accumulation over 2 steps is used. \citealp{buzaaba2025lughallamaadaptinglargelanguage} initializes the models with Llama-3.1-8B, training with a batch size of 512 sequences of 8192 tokens each. Training is conducted for 2400 steps, amounting to approximately 10 billion tokens. A cosine learning rate schedule is employed, beginning with a linear warmup of 240 steps and decaying from 1e-5 to 1e-6. Attention across document boundaries is disabled within sequences, following the pretraining strategy of Llama-3.1-8B.

\textbf{Knowledge distillation and LLM-based approaches.} The final group uses general-purpose LLMs as teachers or as the system itself. \cite{song2026smalllanguagemodelsilver} show that knowledge distillation has proved useful to train models for LRLs. They distilled four fake targets using facebook/nllb200-3.3B, meta-llama/Llama-3.3-70B-Instruct, GPT-4o-mini, and dictionary checking, each containing 621,033 data samples for training. The LRL side-texts are the same and the corresponding fake targets are generated by different teacher models. Input sentence length was capped at 512 tokens and models were trained for only one epoch.

Outside translation, \citealp{zhang2025sentiment} explored a range of models for sentiment analysis including mBERT, AfriBERTa, AfroXLMR, and a monolingual TwiBERT. Their English-based models, such as RoBERTa (12 layers, 125M parameters) and AfriBERTa (111M parameters), were used in a translation-based approach. \citealp{glover2024exploring} leveraged the proprietary GPT-4 model, building a custom speech-based interface to allow interaction in Twi, demonstrating the application of very large language models (LLMs) for educational tools.

\textbf{Trends across paradigms.} Read in this order, a chronological and methodological trend emerges. Early work (2021–2022) trained Transformers and BiLSTMs from scratch on small parallel corpora, typically under 30,000 sentence pairs. Recent work shifts toward fine-tuning large pretrained multilingual models, adopting Africa-centric pretraining as a stronger starting point than generic multilingual checkpoints, and introducing optimization strategies to compensate for the scarcity of labeled Ghanaian-language data. This progression suggests the field is converging on pretraining and adaptation as the dominant strategy for Ghanaian NLP. The evaluation practices accompanying these systems are detailed next.

\begin{table*}
  \centering
  \resizebox{\textwidth}{!}{%
  \begin{tabular}{p{0.25\textwidth}p{0.15\textwidth}p{0.15\textwidth}llll}
    \hline
    \textbf{Papers} & \textbf{Architecture} & \textbf{Language Pair} & \textbf{SpBLEU} & \textbf{ROUGE} & \textbf{WER} & \textbf{F1} \\
    \hline

    \cite{agyei2025dynamic} & Transformer & English-Twi  & 0.7130 & 0.6524 & 0.6832 & - \\
    \hline

    \cite{agyei2025enhancing} & Transformer & English-Twi & 0.7130 & 0.6524 & 0.6832 & - \\
    \hline

    \cite{zhang2025sentiment} & Transformer & English-Twi  & - & - & - & 0.577 \\
    \cline{2-7}
                              & Transformer & Multilingual & - & - & - & 0.644 \\
    \hline

    \cite{adebara2023serengetimassivelymultilinguallanguage} & Transformer & Multilingual & - & - & - & 0.8227 \\
    \hline


    \cite{adebara_afrolid_2022} & Transformer & Twi & - & - & - & 1.000 \\
                                & Transformer & Fon & - & - & - & 0.970 \\
                                & Transformer & Akan & - & - & - & 0.970 \\
    \hline


  \end{tabular}}
  \caption{Summary of the SpBLEU, ROUGE, WER, and F1 scores of reviewed models.}
  \label{tab:summary-SpBLEU-rouge-wer-f1}
\end{table*}

\subsection{Detailed Model Evaluation Metrics}
\label{app:model-eval}

Section \ref{sec:survery-results--model-evals-metrics} introduced the three metrics that dominate the reviewed literature. This section records the additional metrics and evaluation frameworks that appear in more recent work, which broaden the picture considerably.

Recent work has expanded beyond BLEU, TER, and chrF to provide a more comprehensive assessment, particularly for LRLs. \citealp{agyei2025enhancing,agyei2025dynamic} employed a multi-faceted suite including SpBLEU (SentencePiece BLEU), which operates on subword units and is better suited to morphologically rich languages; Recall-Oriented Understudy for Gisting Evaluation (ROUGE) \cite{lin-2004-rouge}, which focuses on the recall of meaningful content and n-grams; and Word Error Rate (WER), which measures the alignment between predicted and reference translations based on edit operations.

Task type also shapes metric choice. For sentiment analysis tasks, \citealp{zhang2025sentiment} and \citealp{raychawdhary2025empowering} used macro-averaged and weighted F1 scores to evaluate model performance, as these metrics account for class imbalance.

Beyond individual metrics, several evaluation frameworks and benchmarks are now available for African languages. \citealp{adelani2025irokobenchnewbenchmarkafrican} developed a comprehensive, human-curated benchmark evaluating LLMs on 17 African languages, with tasks covering natural language inference (AfriXNLI), mathematical reasoning (AfriMGSM), and knowledge-based multiple-choice question answering (AfriMMLU). AfroNLU and AfroNLG, developed by \citealp{adebara2023serengetimassivelymultilinguallanguage, adebara2024cheetahnaturallanguagegeneration}, support 517 African languages. For MT evaluation specifically, the learned metrics SSA-COMET and SSA-COMET-QE achieve the highest average correlation with human judgment among available approaches for Sub-Saharan African languages, offering a more efficient alternative to LLM prompting  \cite{li2025ssacometllmsoutperformlearned}.

What this inventory does not settle is whether metrics developed largely for high-resource, non-tonal languages capture the errors that matter most in Ghanaian languages. We take up that question, and its implications for how the field measures progress, in Section \ref{sec:eval-toward-culturally-grounded-benchmark}.

\subsection{Machine Translation Paradigms}
\label{app:mt-paradigms}
Section \ref{sec:nlp-and-mt-in-lrls-revitalization} noted that MT has developed through four paradigms; we summarize each here, since the reviewed systems draw on assumptions inherited from all four.

RBMT relies on grammatical rules and extensive linguistic knowledge, resulting in poor translation quality and requiring significant manual post-editing \cite{wang_pre-trained_2022,wang_progress_2022}. SMT, which uses statistical models to generate translations based on bilingual text analysis \cite{brown_statistical_1990}, faces challenges such as lack of context consideration and reordering issues for distant language pairs \cite{wang_progress_2022}. HMT combines SMT and RBMT, leveraging translation memory to improve translation quality \cite{hunsicker_machine_2012, costa-jussa_latest_2015}. 

NMT, the current state-of-the-art approach, employs Neural Networks to enhance translation accuracy and speed, enabling training and scaling to new languages \cite{maazouzi_systematic_2017, swathi_machine_2020, mohamed_neural_2021, ranathunga_neural_2023}. Recent advancements in Large Language Models (LLMs) have further revolutionized the field, with early models like GPT-4o \cite{hurst2024gpt} and PaLM \cite{chowdhery_palm_2022}  demonstrating strong performance across a wide range of NLP tasks, including MT, and more recent frontier models such as GPT-5.6 \cite{openai2026gpt56}, Claude Opus 5 \cite{anthropic2026opus5}, and Gemini 3.1 Pro \cite{google2026gemini31pro} extending these gains to reasoning-intensive and low-resource multilingual settings, including translation, summarization, and question answering.

\section{Discussion and Roadmap}
\label{sec:discussion}

Our systematic review of 36 papers reveals both meaningful progress and sobering structural gaps in NLP for Ghanaian languages. The following sections synthesize the methodological patterns identified in Section~\ref{sec:survery-results}, critically assess their limitations, and translate these observations into a prioritized research roadmap. We organize the discussion around four interconnected pillars: data, modeling, evaluation, and community infrastructure. These are sorted with purpose: data constraints shape which models are viable, modeling choices determine what evaluation must be sensitive to, and all three depend on infrastructure that does not currently exist. Throughout, we offer these as directions we see as promising rather than as requirements on the field; each rests on evidence from a literature that is itself small, and reasonable alternatives exist at every point.

\subsection{The Data Bottleneck: Diversity, Quality, and Bias}

The single most persistent obstacle to advancing Ghanaian NLP is the scarcity of high-quality, diverse, and linguistically representative data. Our review reveals that this problem is not merely quantitative, it is also structural. Despite decades of interest in documenting Ghanaian languages, available corpora remain narrow in domain, skewed toward a single language family, and heavily dependent on a small number of recurring source types. These three constraints, domain, language coverage and quality, are separable and we take them in turn.

\subsubsection{Overreliance on Religious Corpora}
\label{sec:discussion--religious-corpora}

Religious texts, primarily the Bible and Jehovah's Witness publications, appear as data sources in at least twelve of the reviewed papers (Section \ref{sec:survery-results}). While this approach is understandable given the near-universal availability of such texts in many African languages, it introduces systematic limitations that the field has not yet adequately addressed. Religious corpora reflect archaic register and vocabulary, are culturally specific to a faith tradition, and are structurally atypical relative to everyday speech, legal, medical, or journalistic language. Models trained predominantly on such data are unlikely to perform well in the domains where language technology is most consequential: healthcare, education, civic participation, and commerce.

We therefore recommend that future work treat religious corpora as a \emph{baseline} or \emph{supplement}, rather than a primary resource, and that researchers explicitly report the domain distribution of their training data to enable meaningful cross-study comparison. Parallel corpora derived from news, government publications, social media, and oral tradition transcriptions should be actively pursued.

Domain narrowness compounds a second and more consequential imbalance in which languages are represented at all.

\subsubsection{The Twi-Centricity Problem}
A striking pattern across the reviewed literature is the near-exclusive concentration on Twi, and specifically Asante Twi, among the 73 living indigenous Ghanaian languages. Ewe and Fante receive marginal attention, while the remaining 70+ languages are largely absent from the NLP literature. This asymmetry reflects and reinforces existing sociolinguistic inequalities: the languages most at risk of extinction receive the least research attention, creating a feedback loop that accelerates their marginalization in the digital sphere.

Addressing this requires deliberate prioritization. We advocate for a \emph{language-first} research agenda in which the selection of target languages is driven by endangerment status and community need, rather than by the convenience of available data. Concretely, this means expanding data collection efforts to include Dagbane, Ga, Nzema, Gonja, and Kasem, all of which are government-sponsored languages under Ghana's current language policy, as an immediate next step, before targeting the broader set of undocumented languages. We note that this ordering trades breadth for tractability, and a research group with existing community relationships in a non-sponsored language would have good reason to sequence differently.

A recent release from the GhanaNLP initiative illustrates what this alternative sequencing can look like in practice. \citealp{gyamfi2026ghananlp} document 41,513 English-aligned sentence pairs across Twi, Fante, Ewe, Ga, and Kusaal, produced by professional translators and annotators, released with structural metadata. Kusaal is the notable case: a Gur-family language absent from the rest of the reviewed literature and outside the eleven sponsored languages, included alongside Ga, which our own priority list names. 

\subsubsection{Data Quality and Standardization}
\label{sec:discussion--data-quality}

Across the reviewed papers, data cleaning and preprocessing practices are inconsistently reported and highly variable. Some studies apply comprehensive pipelines using tools such as Sketch Engine and jusText for deduplication and boilerplate removal, while others provide limited documentation of their preprocessing steps. This inconsistency makes it difficult to attribute performance differences to modeling decisions versus data quality, and impedes reproducibility.

Additionally, many Ghanaian languages, particularly within the Akan family, exhibit significant orthographic variation and lack fully standardized written forms. Akuapem Twi has historically served as the literary standard, while Asante Twi dominates contemporary usage; most NLP datasets do not clearly specify which orthographic convention they follow. We recommend that the community adopt and enforce explicit orthographic documentation standards for all future dataset releases, and invest in building shared normalization tools for Akan dialectal variation.

These data constraints set the terms for the modeling work reviewed in Section \ref{sec:survery-results--model-architectures}, to which we now turn.

\subsection{Modeling: Strengths, Limitations, and Emerging Paradigms}
{Section \ref{sec:survery-results--model-architectures} described what the reviewed papers build; here we ask how well those choices fit the setting. We take the near-consensus on Transformer architectures first, then the shift toward African-language pretraining that partly answers its limitations.

\subsubsection{The Transformer Consensus and Its Limits}
Transformer-based architectures dominate the reviewed literature, appearing in the overwhelming majority of modeling papers, with Bidirectional LSTMs serving as a secondary approach. This preference is well-founded: Transformers achieve state-of-the-art performance across a wide range of NLP tasks, support pre-training and fine-tuning paradigms that are especially valuable in low-resource settings, and benefit from the rich ecosystem of open-source tools such as the HuggingFace Transformers library.

However, the dominance of Transformers should not obscure their particular challenges in low-resource contexts. Pre-training on high-resource languages and fine-tuning on small Ghanaian language datasets risks surface-level adaptation without genuine linguistic transfer. Several papers in our review fine-tuned OPUS-MT or T5-Small models on datasets of fewer than 30,000 sentence pairs, raising questions about the depth of adaptation achievable at these scales. This limitation is about the starting point rather than the architecture, which is precisely what the next line of work changes.

\subsubsection{Multilingual Models and African Language Pretraining}
\label{sec:multilingual-models-and-al-pretraining}
The development of Africa-focused multilingual models, including Serengeti and Cheetah, which jointly cover 517 African languages, represents a qualitative shift in the field. These models provide a more linguistically appropriate pretraining base than generic multilingual models, whose African language coverage is sparse and whose dominant training languages may introduce inductive biases.

We recommend that future modeling work adopt African-centric pretrained models as the default baseline, rather than general multilingual models, and systematically report comparisons against both to quantify the benefit of Africa-specific pretraining for Ghanaian languages.

\subsection{Evaluation: Toward Culturally Grounded Benchmarks}
\label{sec:eval-toward-culturally-grounded-benchmark}
Evaluation is where the constraints described above are hardest to see, because a metric insensitive to them still returns a number. We examine what the reviewed papers measure, what those measurements miss in tonal and morphologically rich languages, and which alternatives have recently become available.

\subsubsection{Limitations of Lexical Metrics and Afrocentric Benchmarking}
\label{sec:limits-of-lexical-metrics-and-afrocentric}
BLEU is a widely used metric that calculates the similarity between candidate and reference translations by computing the precision of n-grams. It is computationally inexpensive, language-independent, and has shown a high correlation with human evaluation. TER measures the minimum number of edits required to transform the candidate translation into the reference, normalized by the average reference length, involving a search over possible shifts to minimize the edit count. chrF calculates an F-1 score over character n-grams, providing a more nuanced translation quality assessment, and has demonstrated a superior correlation with human evaluations compared to BLEU and TER \citep{popovic_chrf_2015}.

Despite their widespread use in the literature, automatic lexical metrics are fundamentally ill-suited for morphologically rich, tonal languages common across Ghana. Lexical overlap criteria fail to account for tonal shifts and diacritic variations that fundamentally alter meaning, meaning that a high BLEU score can obscure catastrophic semantic errors. This concern is not specific to Ghanaian languages. Comparable findings in other tonal, low-resource African languages illustrate the same failure mode. For instance, work on multilingual ASR evaluation for Yoruba and Uneme has argued that standard lexical metrics obscure phonologically and tonally significant errors, motivating metrics that are more sensitive to these dimensions \cite{chen2026linguistically}.

Furthermore, evaluating Ghanaian models in isolation fails to capture their downstream utility. To establish rigorous baselines, the local research roadmap must pivot toward standardized, multi-task Afrocentric evaluation suites. Frameworks like \textbf{AfriMTEB} provide a massive text embedding benchmark spanning tasks like semantic textual similarity, bitext mining, and retrieval across underrepresented African languages \cite{uemura2026afrimteb}. Utilizing these macro-level benchmarks is vital for neutralizing the English-dominance bias inherent in current generative validation pipelines. Metadata-level analyses of generative AI usefulness for other under-resourced African language communities (e.g., Ibibio and Margi) similarly caution that evaluation criteria imported wholesale from high-resource settings risk missing culturally and linguistically relevant distinctions \cite{waliya2026metadata}, a caution we believe applies with equal force to Ghana.

\subsubsection{Zero-Shot LLM Evaluation Across Broader Language Coverage}
\label{sec:zero-shot-llm}

These concerns about metric sensitivity bear directly on how a recent broad-coverage result should be read. A notable recent departure from the fine-tuning paradigm that dominates this survey is \citealp{moore2026nsankuevaluatingzeroshottranslation}, who introduce Nsanku, a zero-shot evaluation of LLM translation performance spanning 43 Ghanaian languages. Rather than fine-tuning a model on a curated parallel corpus, this work directly prompts general-purpose LLMs and reports aggregate performance across the full language set, with the best-performing model, gemini-2.5-flash, achieving a 43-language average BLEU of 0.2460 and chrF of 0.2916 (reported as 24.60 and 29.16 on the conventional 0 -- 100 scale in the original paper).
This result is encouraging but should be interpreted with caution. The reported scores are averaged across 43 languages of widely varying resource availability, and an aggregate BLEU/chrF figure at this scale can obscure substantial per-language variance; a handful of higher-resource languages (e.g., Twi, Ewe) may be propping up an average that does not reflect performance on the majority of included languages. Unlike the fine-tuned, single-language-pair results reported elsewhere in Table \ref{tab:summary-bleu-chrf-ter}, this figure is not directly comparable to prior Twi-specific benchmarks. We include it here as evidence that broad-coverage zero-shot evaluation is now feasible for Ghanaian languages, while flagging that per-language breakdowns, rather than aggregate scores, will be necessary before this approach can be said to have closed the coverage gap documented throughout this survey.

\subsubsection{What the Reviewed Papers Measure, and What Is Emerging}
The problems identified in \ref{sec:limits-of-lexical-metrics-and-afrocentric} are not hypothetical for this literature: the reviewed papers rely predominantly on BLEU~\cite{papineni_bleu_2002}, with TER~\cite{snover_study_2006} and CHRF~\cite{popovic_chrf_2015} appearing less frequently. BLEU's known weaknesses are amplified in the low-resource setting, where reference translations are few and often crowd-sourced without expert validation. The wide variance in BLEU scores across the reviewed papers (ranging from 0.04 to 0.81) is in part attributable to methodological heterogeneity in reference translation quality rather than model capability alone, which is the practical cost of the metric problem, since it leaves the field without a reliable way to tell whether a new system is better than an old one.

Recent papers in our review reflect a welcome shift toward more comprehensive evaluation suites. SpBLEU, which operates on subword units and is better suited to morphologically rich languages, and ROUGE, which focuses on content recall, offer a more nuanced picture of translation quality. More broadly, we recommend that the field move toward evaluation approaches that address the specific challenges highlighted above: sensitivity to morphological and tonal variation, robustness to variable reference quality, and correlation with human judgment on African languages rather than on English. The recently proposed SSA-COMET and SSA-COMET-QE metrics are one example of what this can look like in practice; they achieve higher correlation with human judgments than both BLEU-based metrics and LLM prompting-based evaluation for Sub-Saharan African languages \cite{li2025ssacometllmsoutperformlearned}, and are worth considering alongside, or in place of, lexical metrics for MT work in this setting. Learned metrics carry their own dependencies, on the languages and domains represented in their training data, so we would expect reporting both a learned and a lexical metric to remain informative for some time.

\subsubsection{Task-Specific and Holistic Benchmarks}
Beyond translation, the IrokoBench benchmark~\cite{adelani2025irokobenchnewbenchmarkafrican}, covering natural language inference, mathematical reasoning, and multi-choice question answering across 17 African languages, illustrates the value of holistic, multi-task evaluation for assessing the true capabilities of LLMs in African language contexts. Ghanaian languages are partially represented in this benchmark, but dedicated Ghanaian language benchmarks covering the full task spectrum (named entity recognition, part-of-speech tagging, sentiment analysis, question answering, and summarization) do not yet exist. Constructing such benchmarks, with the involvement of native speakers and professional linguists, is a critical prerequisite for tracking progress meaningfully across the field. Similarly, Afri-MCQA \cite{tonja-etal-2026-afri}, a multimodal cultural question-answering benchmark spanning 15 African languages, demonstrates the value of coupling question-answering tasks with native-speaker oversight and speech-modality support, highlighting pathways toward more holistic and culturally grounded evaluation.

\subsubsection{Bias and Fairness Evaluation}
\label{sec:bias-and-fairness-evaluation}
Perhaps the most underexplored dimension of model evaluation in our review is the assessment of social biases. The work of \citealp{oppong2025examining} is a notable exception: their examination of gender bias in GPT-4o for English-Twi translation, finding systematic bias against female gender in pronoun allocation, underscores that evaluation cannot be reduced to accuracy metrics alone. This finding is particularly significant given the cultural and linguistic complexity of gender expression in Ghanaian languages, which differs substantially from Indo-European conventions. Comprehensive model evaluation for Ghanaian NLP must include fairness audits alongside performance metrics. We recommend developing a standardized gender and social bias evaluation suite adapted to the cultural norms of Ghanaian language communities.

Data, modeling, and evaluation all point toward a common underlying constraint, which is the subject of the final pillar.

\subsection{Infrastructure, Coordination, and Community Engagement}
\label{sec:infrastructure-coordination-and-community}
Each of the preceding three pillars assumes resources that someone has already built and released. This section turns to that substrate: the shared datasets, checkpoints, standards, and community relationships that Ghanaian NLP currently lacks.

\subsubsection{Fragmentation and the Absence of Shared Standards}
A recurring challenge across the reviewed literature is the absence of shared dataset releases, standardized evaluation protocols, and coordinated research infrastructure. Many datasets constructed for individual papers are not publicly released, forcing subsequent researchers to repeat data collection from scratch. Model checkpoints are rarely shared in a manner that permits direct comparison or fine-tuning by other groups. This fragmentation inflates the cost of entry for new researchers and prevents the field from building cumulatively on prior work.

A centralized Ghanaian NLP resource hub aggregating datasets, model checkpoints, evaluation benchmarks, and documentation standards across all Ghanaian languages would reduce this redundancy substantially. Whether such a hub is best hosted by a university, a government body, or an existing pan-African research network is an open question, and one that likely matters as much to its longevity as any technical decision.

\subsubsection{Community-Centered Data Collection}
The long-term sustainability of NLP for Ghanaian languages depends on building data collection capacity within Ghana itself. Crowdsourcing approaches used in the reviewed literature demonstrate that community involvement is both feasible and valuable. However, these efforts remain ad hoc and under-resourced. Structured community annotation programs, in partnership with Ghanaian universities, language boards, and the Bureau of Ghana Languages, would put this work on firmer footing, particularly where they provide fair compensation to annotators, support quality control through expert adjudication and generate data in formats compatible with shared infrastructure. The reviewed literature offers little guidance on what compensation or adjudication practices work best at scale, which is itself a gap worth closing. Where such programs are established, we see strong reasons for community members to participate as co-designers rather than as data suppliers alone.

\subsubsection{Ethical Considerations and Data Sovereignty}
That distinction between participation and extraction raises questions the NLP community has generally left implicit. Data collection for endangered and LRLs raises ethical questions around consent, attribution, and community ownership of language data that are particularly salient for indigenous language communities that have historically experienced extractive relationships with outside researchers. Community data agreements specifying data use, sharing restrictions, and attribution rights offer one mechanism for addressing these concerns, and ethical review documentation for language data collection involving indigenous communities would make the relevant commitments visible at publication time. These are norms the field would have to arrive at collectively rather than measures any single group can adopt unilaterally, and existing frameworks from Indigenous data sovereignty work elsewhere are a plausible starting point.

\subsubsection{The Digital Divide and Algorithmic Language Extinction}
The rapid proliferation of LLMs has shifted this boundary toward \textit{algorithmic language extinction} while structural barriers like fragmented digital infrastructure and limited literacy historically defined the digital divide in Sub-Saharan Africa. The exclusion of Ghanaian languages from state-of-the-art commercial architectures like Llama 3 and Claude 3.5 acts as an institutional bottleneck for local human-computer interaction, effectively creating an AI-driven tier of digital marginalization. As noted by \cite{mbaye2026opportunities}, a new paradigm of digital inequity has emerged, tied explicitly to the depth and accuracy with which indigenous languages are captured by generative models. This systemic oversight risks the technological erasure of major Ghanaian linguistic communities without active intervention, despite the broader continent seeing a quadrupling of published African NLP literature over the last five years \cite{alabi2025charting}. This pattern is consistent with \citet{blasi-etal-2022-systematic}'s finding that NLP investment tracks the economic and political power of speaker communities rather than population size, highlighting that Ghana's experience, however locally specific in its mechanisms, is an instance of a worldwide structural phenomenon.

\subsection{Prioritized Research Roadmap}

Drawing on the analysis above, we propose the following prioritized agenda for the next phase of Ghanaian NLP research. The roadmap is organized into three time horizons.

\paragraph{Near-Term (1--2 years).}
\begin{enumerate}
    \item \textbf{Expand language coverage beyond Twi.} Initiate parallel corpus construction for at minimum the 11 government-sponsored languages, with priority given to those most at risk: Nzema, Gonja, Kasem, and Dagaare. Even small, high-quality datasets of 10,000-50,000 sentence pairs represent transformative resources for languages that currently have none.
    \item \textbf{Release existing datasets and model checkpoints.} Researchers who have constructed unpublished datasets in the course of prior work should be encouraged, and, where possible, supported to release these resources under open licenses, with appropriate documentation of sources, preprocessing, and known limitations.
    \item \textbf{Adopt Africa-centric pretrained models and SSA-COMET as the community standard.} Serengeti, Cheetah, and AfroXLMR are natural default pretraining baselines for this setting, and reporting them alongside generic multilingual models would establish how much Africa-specific pretraining contributes. On the evaluation side, we recommend metrics that address the challenges set out in \ref{sec:eval-toward-culturally-grounded-benchmark} tonal and morphological sensitivity, robustness to variable reference quality, and demonstrated correlation with human judgment on African languages. SSA-COMET is one current example; reporting it alongside BLEU, rather than in place of it, would let the field assess the transition rather than assume it.
    \item \textbf{Establish orthographic standards.} In collaboration with the Bureau of Ghana Languages and academic linguists, develop and disseminate orthographic guidelines for the major Ghanaian languages, enabling consistent dataset construction and cross-paper comparison.
\end{enumerate}

\paragraph{Medium-Term (2--5 years).}
\begin{enumerate}
    \item \textbf{Build a comprehensive Ghanaian NLP benchmark.} Construct a multi-task evaluation suite covering translation, NER, POS tagging, sentiment analysis, and question answering across at least 10 Ghanaian languages, with human-validated reference outputs and culturally grounded evaluation criteria.
    \item \textbf{Develop specialized architectures for tonal and morphologically rich languages.} Investigate tokenization strategies, positional encoding schemes, and pretraining objectives specifically adapted to the tonal and morphological properties of Akan, Ewe, and Gur-family languages, rather than importing architectures designed for morphologically simpler high-resource languages.
    \item \textbf{Institutionalize bias evaluation.} Establish robust evaluation frameworks that extend across a wider range of models, languages, and bias dimensions, including gender, ethnicity, and religious affiliation, and standardize bias reporting as a mandatory component of model evaluation within the research community.
    \item \textbf{Establish a Ghanaian NLP resource hub.} Formalize coordination infrastructure that aggregates datasets, benchmarks, checkpoints, and documentation, and provides mechanisms for community feedback and ongoing dataset curation.
    \item \textbf{Develop culturally and phonologically grounded evaluation methods.} No equivalent phonologically-aware evaluation study yet exists for a Ghanaian language; future work should extend methodologies demonstrated for other tonal, low-resource African languages such as Yoruba and Uneme \citep{chen2026linguistically} to capture tonal and phonological errors that lexical metrics like BLEU obscure. Similarly, no Ghana-specific study has yet examined whether evaluation criteria imported from high-resource settings adequately capture Ghanaian linguistic and cultural distinctions, an open question raised for other underrepresented communities such as Ibibio and Margi \citep{waliya2026metadata}.
\end{enumerate}

\paragraph{Long-Term (5+ years).}
\begin{enumerate}
    \item \textbf{Extend NLP capabilities to all 73 living Ghanaian languages.} The long-term goal is equitable technological representation for every living Ghanaian language. Achieving this at scale will require advances in cross-lingual transfer from related languages, unsupervised and semi-supervised learning from unlabeled data, and possibly dedicated speech corpus construction to support text-to-speech and speech recognition for oral-dominant language communities.
    \item \textbf{Deploy community-facing tools and measure real-world impact.} The ultimate measure of success for Ghanaian NLP is not benchmark performance but community benefit: literate access to government services, educational materials in indigenous languages, healthcare communication tools, and digital content creation by speakers themselves. Research programs should build in participatory design and longitudinal impact evaluation from the outset.
    \item \textbf{Integrate Ghanaian language perspectives into global AI governance.} Ghana's linguistic diversity represents an important test case for the claims of universal AI. Ensuring that Ghanaian linguists, educators, and community members have a voice in the international standards and governance frameworks that will shape the future of multilingual AI is a non-technical but essential component of the long-term roadmap.
\end{enumerate}

In summary, the path forward for Ghanaian NLP requires simultaneous progress on data, modeling, evaluation, and infrastructure, and must be grounded in meaningful partnership with the communities whose languages are at stake. The findings of this survey suggest that the technical barriers, while real, are tractable. The more fundamental challenge is one of sustained commitment, coordination, and equitable resource allocation.

\section{Extended Synthesis and Reproducibility Protocols}
This section provides granular details regarding the systematic review process, specifically focusing on the data synthesis framework and the measures taken to ensure reproducibility.

\subsection{Synthesis Methods}
Included studies were synthesized narratively and grouped by data collection approach (Section \ref{sec:survery-results-data-collections}), model architecture and training paradigm (Section \ref{sec:survery-results--model-architectures}), and evaluation metric (Section \ref{sec:survery-results--model-evals-metrics}). Because the included studies differ in target language pair, dataset size and domain, and choice of evaluation metric, a quantitative meta-analysis was not appropriate; we instead tabulate reported scores alongside architecture and language pair, normalize all scores to the 0-1 scale, and flag results that are not directly comparable to others in the same table.

\subsection{Reproducibility of the Search Protocol}
\label{sec:reproducibility-of-search}
To make the search protocol reproducible, we verified the retrievability of each core paper under every query via title-constrained membership checks, appending a distinctive quoted fragment of each paper's title to the original query. The final column of Table \ref{tab:search-queries} reports these verified counts.

\section{Wikipedia Representation of Ghanaian Languages}
\label{app:wikipedia}

Table~\ref{tab:web-representation} illustrates the resource disparity discussed in Section~\ref{sec:survery-results} at the level of a single, widely used digital resource: Wikipedia. Coverage varies sharply across languages and does not track NLP research attention. Ga currently has no Wikipedia edition at all, while Twi and Fante remain in the low thousands of articles despite being the most studied Ghanaian languages in the NLP literature. Dagbani's comparatively high count is driven substantially by the Content Translation tool, enabled by default on Dagbani Wikipedia since 2024, which supports semi-automated article creation; this pattern, also observed on other small Wikipedia editions, means raw article counts can overstate genuine digitization depth rather than reflect organic community documentation. We report these counts separately from the corpus-level analysis in Section~\ref{sec:survery-results} because encyclopedia article counts and corpus sentence counts measure different things, and because the Dagbani case illustrates how readily a single tooling decision can distort a naive count.

\begin{table}[t]
    \centering
    \begin{tabular}{lr}
    \hline
    Language & Wikipedia Articles  \\
    \hline
    Twi    & 4,736  \\
    Ewe     & 1,351  \\
    Ga     & None  \\
    Fante   & 1,801	  \\
    Dagbani & 14,505  \\
    \hline
    \end{tabular}
    \caption{Digital representation of major Ghanaian languages on Wikipedia. Akan Wikipedia (ak) was closed in April 2023 and its content merged into Twi Wikipedia, reflecting the orthographic standardization challenges discussed in Appendix~\ref{sec:discussion--data-quality}.}
    \label{tab:web-representation}
\end{table}

\section{Distribution of Ghanaian Language Speakers}
This section presents the distribution of speakers across major indigenous Ghanaian languages (Table \ref{tab:percentage-top-speakers}). The values indicate the percentage share of speakers associated with each language category in the dataset.

\begin{table}[h]
  \centering
  \begin{tabular}{ll}
    \hline
    \textbf{Language} & \textbf{\% of Speakers} \\
    \hline
    Asante            & 16\%   \\
    Ewe               & 14\%   \\
    Fante             & 11.6\% \\
    Boron (Brong)     & 4.9\%  \\
    Dagomba           & 4.4\%  \\
    Dangme            & 4.2\%  \\
    Dagarte (Dagaba)  & 3.9\%  \\
    Kokomba           & 3.5\%  \\
    Akyem             & 3.2\%  \\
    Ga                & 3.1\%  \\
    Other             & 31.2\% \\
    \hline
  \end{tabular}
  \caption{Percentage of speakers for top indigenous Ghanaian languages.}
  \label{tab:percentage-top-speakers}
\end{table}

\begin{figure*}
    \centering
    \includegraphics[width=\textwidth]{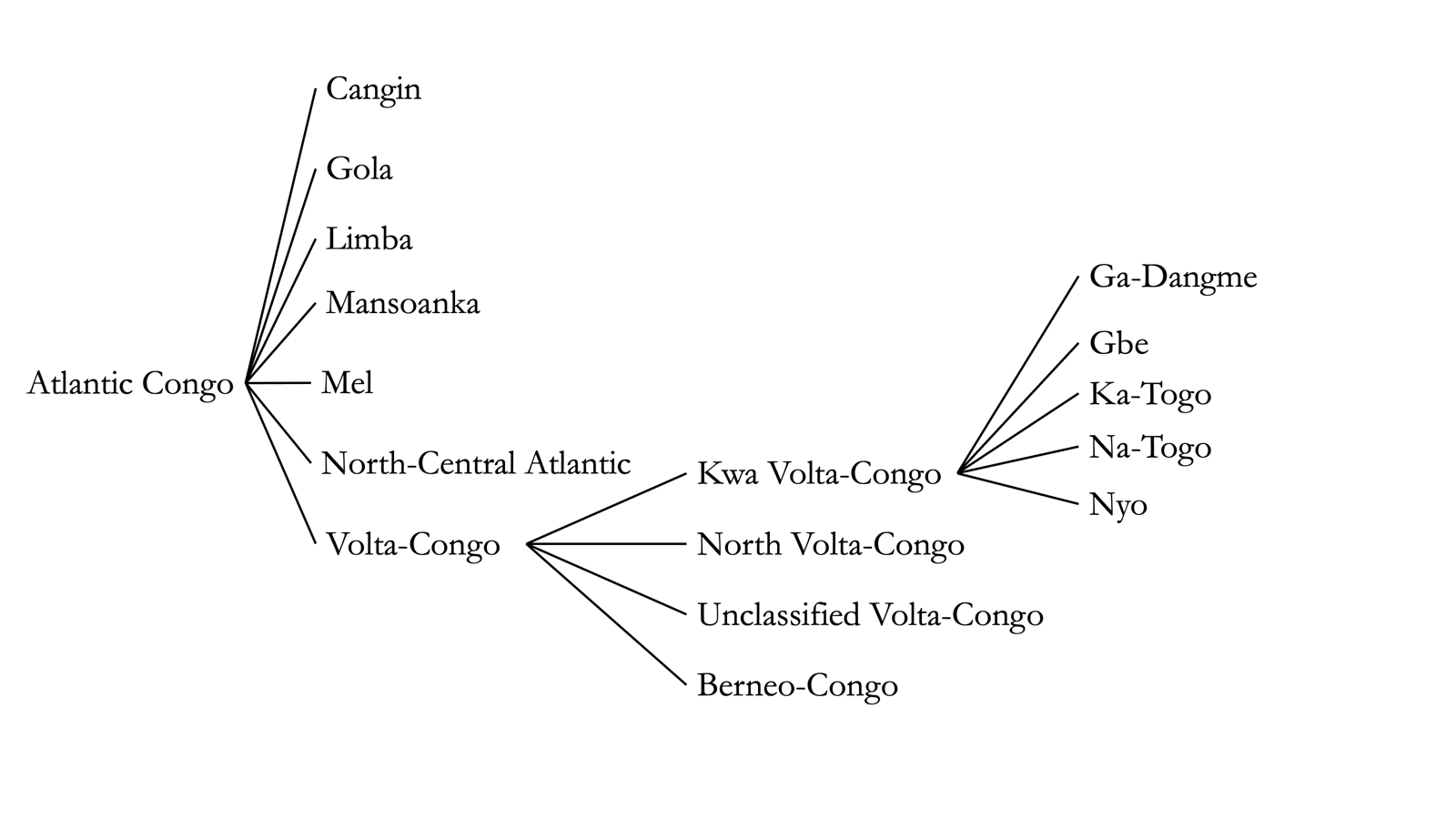}
    \caption{A representation of the Atlantic-Congo language family tree: Languages spoken in Ghana are members of this family. Those in southern Ghana and neighboring countries belong to Kwa Volta-Congo, a group of about 81 languages, including Ga-Dangme (2 languages), Gbe (22), Ka-Togo (8) and Na-Togo (8) as well as Nyo. The Nyo comprises 41 languages, including Akan, which itself entails 8 languages of the Tano subgroup of Potou-Tano \cite{glottolog5_3}}.
    \label{fig:Atlantic-Congo Family tree}
\end{figure*}

\section{Benchmark Coverage Across Ghanaian Languages}
This section compares major multilingual benchmarks in terms of their coverage of Ghanaian languages (Table \ref{tab:matrix_coverage_flipped}). The matrix highlights significant disparities in language inclusion, revealing that most benchmarks prioritize a small subset of languages, while only a few datasets provide broad coverage across the full linguistic spectrum.

\begin{table*}[ht]
\centering 
\caption{Multilingual and Regional Benchmark Coverage Matrix}
\label{tab:matrix_coverage_flipped}
\small
    \begin{tabular}{lcccccccccc}
    \hline
    \textbf{Benchmark} & \rotatebox{90}{\textbf{Asante}} & \rotatebox{90}{\textbf{Ewe}} & \rotatebox{90}{\textbf{Fante}} & \rotatebox{90}{\textbf{Boron}} & \rotatebox{90}{\textbf{Dagomba}} & \rotatebox{90}{\textbf{Dangme}} & \rotatebox{90}{\textbf{Dagarte}} & \rotatebox{90}{\textbf{Kokomba}} & \rotatebox{90}{\textbf{Akyem}} & \rotatebox{90}{\textbf{Ga}} \\
    \hline
    IrokoBench \cite{adelani2025irokobenchnewbenchmarkafrican} & $\checkmark$ & $\checkmark$ & ---          & ---          & ---          & ---          & ---          & ---          & ---          & ---          \\
    
    Belebele \cite{Bandarkar_2024} & $\checkmark$ & ---          & ---          & ---          & ---          & ---          & ---          & ---          & ---          & ---          \\
    
    Global MMLU \cite{singh2025globalmmluunderstandingaddressing} & $\checkmark$ & ---          & ---          & ---          & ---          & ---          & ---          & ---          & ---          & ---          \\
    
    FLORES-200 \cite{nllbteam2022languageleftbehindscaling} & $\checkmark$ & $\checkmark$ & ---          & ---          & --- & ---          & ---          & ---          & ---          & --- \\
    
    SAHARA \cite{adebara2025weevaluatingllmperformance} & $\checkmark$ & $\checkmark$ & $\checkmark$ & $\checkmark$ & $\checkmark$ & $\checkmark$ & $\checkmark$ & $\checkmark$ & $\checkmark$ & $\checkmark$ \\
    Afri-MCQA \cite{tonja-etal-2026-afri} & $\checkmark$ & ---          & ---          & ---          & ---          & ---          & ---          & ---          & ---          & ---          \\
    \hline
    \end{tabular}
\end{table*}

\section{Overview of Core Papers Included in This Survey}
This table provides a consolidated overview of the core studies included in this survey. The listed works span a range of contributions, including machine translation systems, dataset construction efforts, benchmark development, sentiment analysis, multilingual pretraining, and evaluation frameworks relevant to Ghanaian languages (Table \ref{tab:corepapers}). Together, these studies form the empirical and methodological foundation of this review. They reflect the evolution of NLP research in the region, from early corpus creation to recent advances in transformer-based architectures and large-scale multilingual models.

\begin{table*}[t]
\centering
\resizebox{\textwidth}{!}{%
    \begin{tabular}{p{0.42\textwidth}p{0.25\textwidth}p{0.52\textwidth}}
        \hline
        \textbf{Paper} & \textbf{Paradigm} & \textbf{Base Model} \\
        \hline
        \citealp{agyei_akan-english_2021} & Scratch & Transformer \\
        \citealp{adjeisah_pseudotext_2021} & Scratch & Transformer / BiLSTM \\
        \citealp{hacheme_english2gbe_2021} & Scratch & Transformer \\
        \citealp{acheampong_language_2021} & Scratch & Transformer \\
        \hline
        \citealp{azunre_english-twi_2021} & Fine-tune & OPUS-MT \\
        \citealp{gyasi_twi_2023} & Fine-tune & OPUS-MT \\
        \citealp{bannerman2024machine} & Fine-tune & OPUS-MT \\
        \citealp{agyei2025enhancing,agyei2025dynamic} & Fine-tune & T5-Small \\
        \hline
        \citealp{adebara2023serengetimassivelymultilinguallanguage} & Africa-centric PT & XLM-R \\
        \citealp{adebara2024cheetahnaturallanguagegeneration} & Africa-centric PT & T5-like \\
        \citealp{li2025ssacometllmsoutperformlearned} & Africa-centric PT & AfroXLMR-76L \\
        \citealp{buzaaba2025lughallamaadaptinglargelanguage} & Africa-centric PT & Llama-3.1-8B \\
        \hline
        \citealp{song2026smalllanguagemodelsilver} & Distillation & NLLB / Llama / GPT-4o-mini (teachers) \\
        \citealp{glover2024exploring} & LLM-based & GPT-4 \\
        \citealp{zhang2025sentiment} & LLM-based / comparative & mBERT, AfriBERTa, AfroXLMR, TwiBERT \\
        \hline
    \end{tabular}}
    \caption{Taxonomy of modeling paradigms across reviewed papers, by base architecture.}
    \label{tab:modeling-taxonomy}
\end{table*}

\begin{table*}[htbp]
\small
\centering
\resizebox{\textwidth}{!}{%
    \begin{tabular}{p{0.62\textwidth}ll}
        \hline
        \textbf{Paper Title} & \textbf{Citation} & \textbf{Peer-Reviewed} \\
        \hline
        PidginUNMT: Unsupervised Neural Machine Translation from West African Pidgin to English & \cite{ogueji_pidginunmt_2019} & No \\
        Twi Corpus: A Massively Twi-to-Handful Languages Parallel Bible Corpus & \cite{adjeisah_twi_2020} & Yes \\
        Language Revitalization: A Benchmark for Akan-to-English Machine Translation & \cite{acheampong_language_2021} & Yes \\
        Akan-English: Transformer for Low Resource Translation & \cite{agyei_akan-english_2021} & Yes \\
        Pseudotext Injection and Advance Filtering of Low-Resource Corpus for Neural Machine Translation & \cite{adjeisah_pseudotext_2021} & Yes \\
        GELR: A Bilingual Ewe-English Corpus Building and Evaluation & \cite{yvette_gelr_2021} & Yes \\
        NLP for Ghanaian Languages & \cite{azunre_nlp_2021} & No \\
        English-Twi Parallel Corpus for Machine Translation & \cite{azunre_english-twi_2021} & No \\
        Contextual Text Embeddings for Twi & \cite{azunre_contextual_2021} & No \\
        English2Gbe: A Multilingual Machine Translation Model for Fon/Ewe-Gbe & \cite{hacheme_english2gbe_2021} & No \\
        TWIENG: A Multi-Domain Twi-English Parallel Corpus for Machine Translation of Twi, a Low-Resource African Language & \cite{afram_twieng_2022} & No \\
        AfroLID: A Neural Language Identification Tool for African Languages & \cite{adebara_afrolid_2022} & Yes \\
        MasakhaNER 2.0: Africa-Centric Transfer Learning for Named Entity Recognition & \cite{adelani-etal-2022-masakhaner} & Yes \\
        BibleTTS: A Large, High-Fidelity, Multilingual, and Uniquely African Speech Corpus & \cite{meyer2022biblettslargehighfidelitymultilingual} & Yes \\
        SERENGETI: Massively Multilingual Language Models for Africa & \cite{adebara2023serengetimassivelymultilinguallanguage} & Yes \\
        MasakhaPOS: Part-of-Speech Tagging for Typologically Diverse African Languages & \cite{dione2023masakhapospartofspeechtaggingtypologically} & Yes \\
        SemEval-2023 Task 12: Sentiment Analysis for African Languages (AfriSenti-SemEval) & \cite{muhammad-etal-2023-semeval} & Yes \\
        Twi Machine Translation & \cite{gyasi_twi_2023} & Yes \\
        Machine Translation from English-Twi in Parallel Corpus: Low Resource Ghanaian Language (NLP) & \cite{bannerman2024machine} & No \\
        Cheetah: Natural Language Generation for 517 African Languages & \cite{adebara2024cheetahnaturallanguagegeneration} & Yes \\
        Exploring the Efficacy of a Local-Language Interfaced ChatGPT as a Self-Learning Support Tool for Junior High School Students in Rural Communities in Africa & \cite{glover2024exploring} & Yes \\
        Low Resource Twi-English Parallel Corpus for Machine Translation in Multiple Domains (Twi-2-ENG) & \cite{agyei2024low} & Yes \\
        Where Are We? Evaluating LLM Performance on African Languages (SAHARA) & \cite{adebara2025weevaluatingllmperformance} & Yes \\
        IrokoBench: A New Benchmark for African Languages in the Age of Large Language Models & \cite{adelani2025irokobenchnewbenchmarkafrican} & Yes \\
        Enhancing Machine Translation for Low-Resource Languages: A Cross-Lingual Learning Approach for Twi & \cite{agyei2025enhancing} & Yes \\
        Dynamic Aggregation and Augmentation for Low-Resource Machine Translation Using Federated Fine-Tuning of Pretrained Transformer Models & \cite{agyei2025dynamic} & Yes \\
        SSA-COMET: Do LLMs Outperform Learned Metrics in Evaluating MT for Under-Resourced African Languages? & \cite{li2025ssacometllmsoutperformlearned} & Yes \\
        Empowering Sentiment Analysis in African Low-Resource Languages Through Transformer Models and Strategic Language Selection & \cite{raychawdhary2025empowering} & Yes \\
        AsanteTwiSenti: A Sentiment Dataset of Ghanaian Asante Twi Tweets in a Multilingual Context & \cite{gyimah2025asantetwisenti} & Yes \\
        Sentiment Analysis for the African Language Twi: Translation-Based vs.\ End-to-End Approaches & \cite{zhang2025sentiment} & Yes \\
        Lugha-Llama: Adapting Large Language Models for African Languages & \cite{buzaaba2025lughallamaadaptinglargelanguage} & No \\
        Examining the Cultural Encoding of Gender Bias in LLMs for Low-Resourced African Languages & \cite{oppong2025examining} & Yes \\
        Advancing Automatic Speech Recognition for Low-Resource Ghanaian Languages: Audio Datasets for Akan, Ewe, Dagbani, Dagaare, and Ikposo (UGSpeechData) & \cite{WiafeUGSpeechDataPaper, WiafeUGSpeechDataRepo} & Yes\textsuperscript{\dag} \\
        \\ 
        GhanaNLP Parallel Corpora: Comprehensive Multilingual Resources for Low-Resource Ghanaian Languages & \cite{gyamfi2026ghananlp} & No \\
        Are Small Language Models the Silver Bullet to Low-Resource Languages Machine Translation? & \cite{song2026smalllanguagemodelsilver} & Yes \\
        Nsanku: Evaluating Zero-Shot Translation Performance of LLMs for Ghanaian Languages & \cite{moore2026nsankuevaluatingzeroshottranslation} & No \\
        \hline
    \end{tabular}}
    \caption{Core papers analyzed in this review ($n=36$), ordered chronologically, with peer-review status. Status reflects the version cited in this survey: 26 of the 36 core works appeared in peer-reviewed venues, while 10 are preprints or otherwise unrefereed. UGSpeechData is counted as a single core work described across two records. \textsuperscript{\dag}The accompanying dataset article is peer-reviewed; the linked repository record is not.}
    \label{tab:corepapers}
\end{table*}

\section{Abstract Translations}
\label{app:translations}

To promote accessibility, we provide full machine translations of the abstract into 15 languages below, generated using Mansa.

\begin{figure}[htbp]
    \centering
    \includegraphics[width=\linewidth]{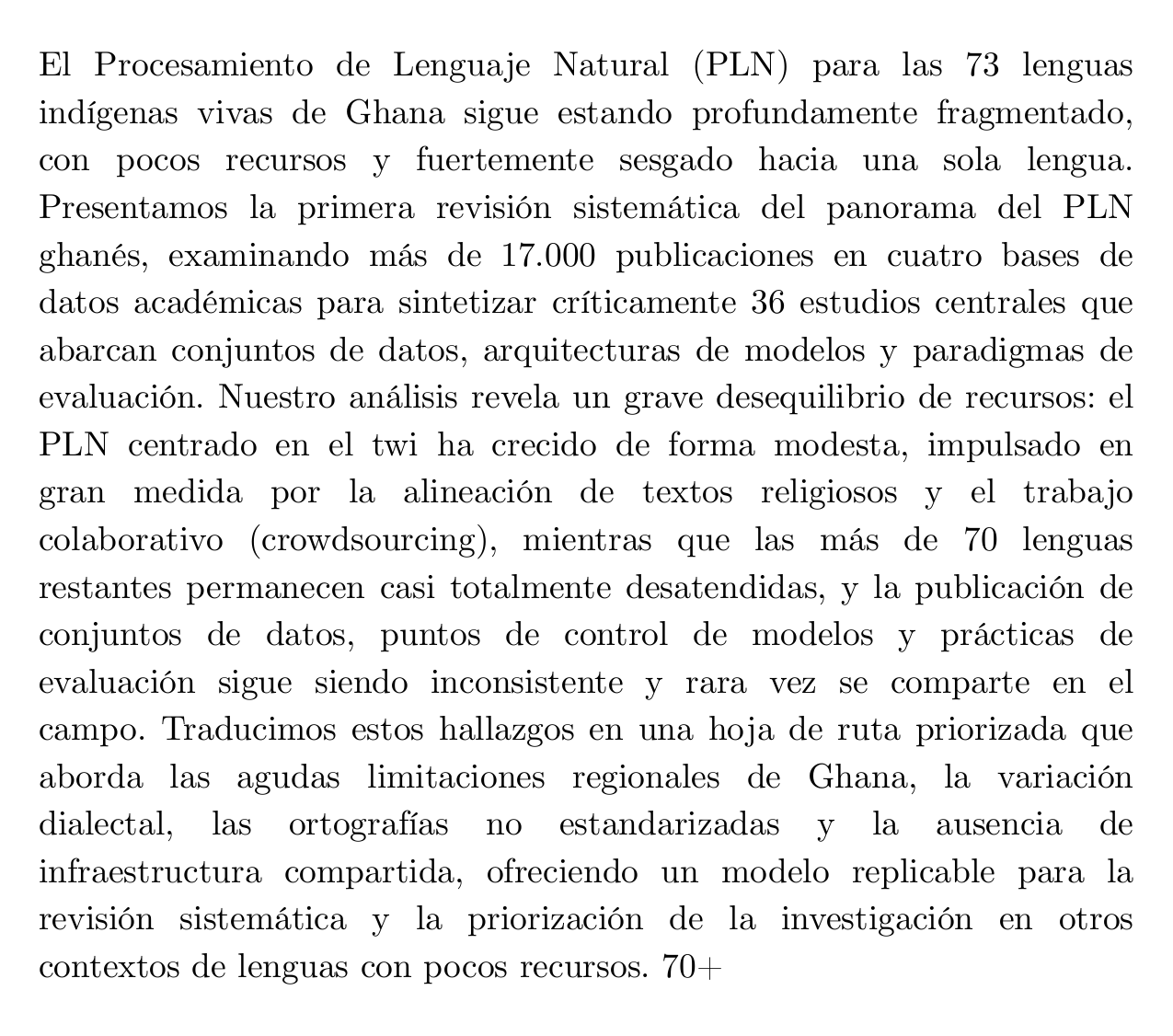}
    \caption{Spanish Translation}
\end{figure}

\begin{figure}[htbp]
    \centering
    \includegraphics[width=\linewidth]{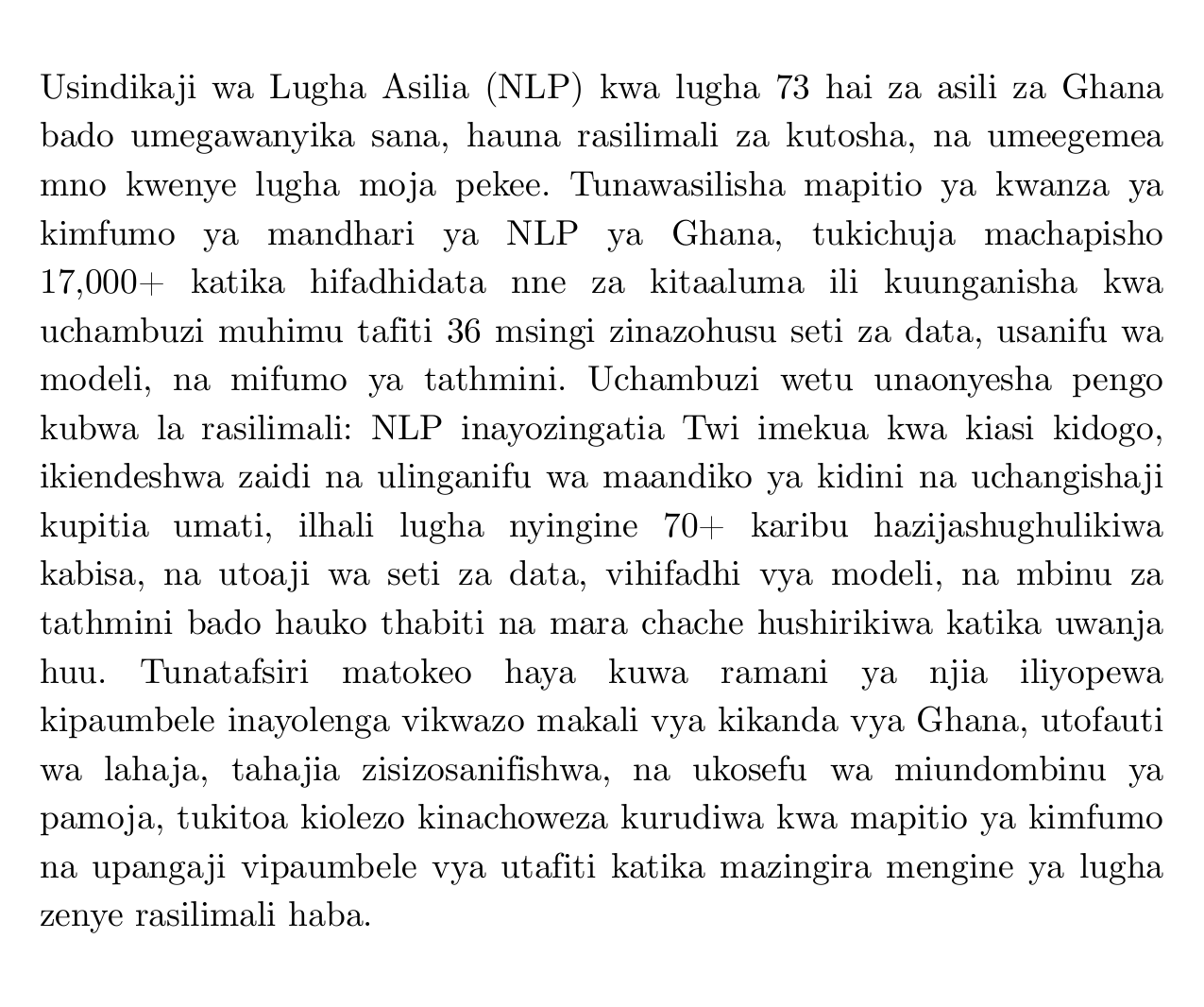}
    \caption{Swahili Translation}
\end{figure}

\begin{figure}[htbp]
    \centering
    \includegraphics[width=\linewidth]{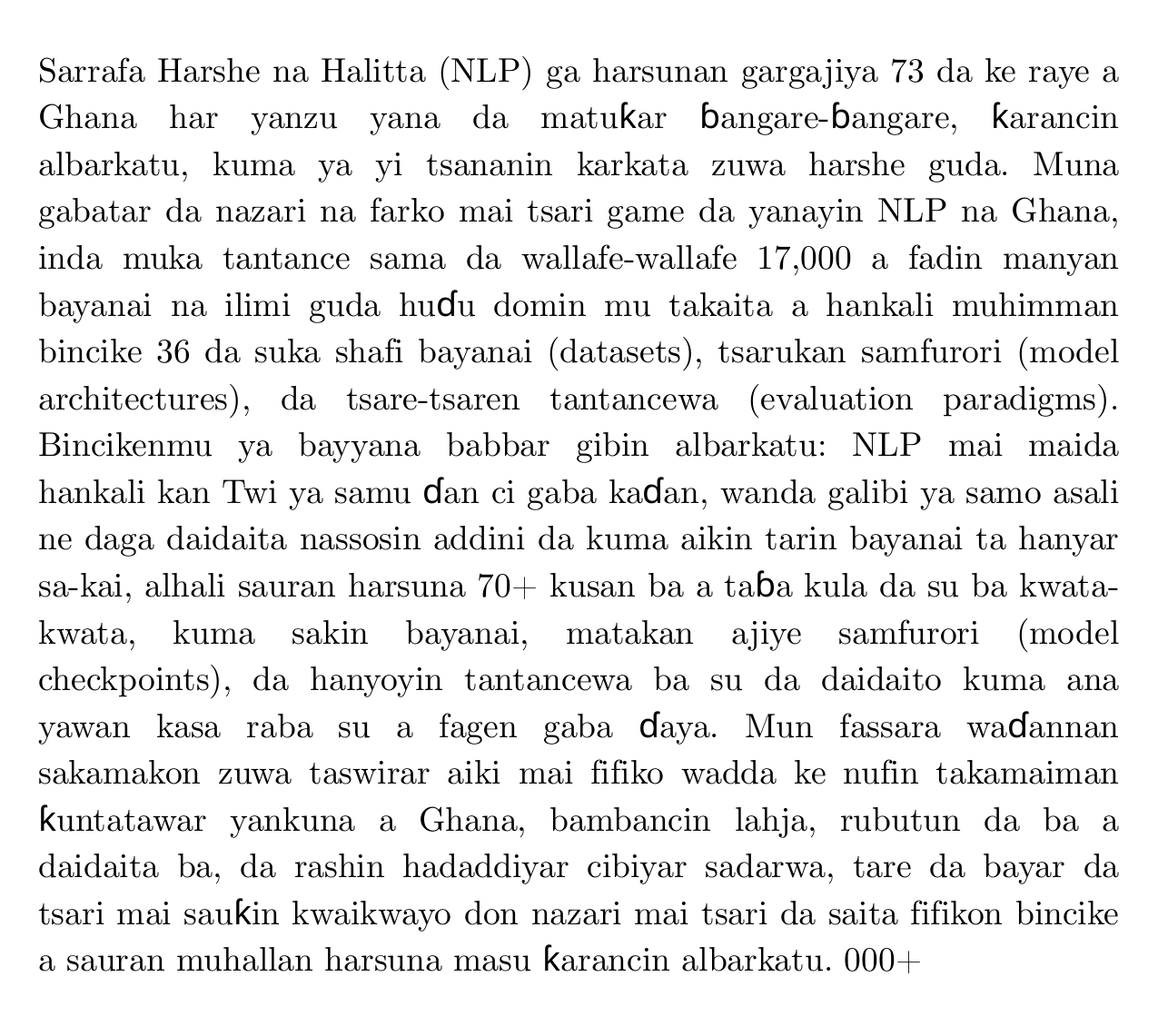}
    \caption{Hausa Translation}
\end{figure}

\begin{figure}[htbp]
    \centering
    \includegraphics[width=\linewidth]{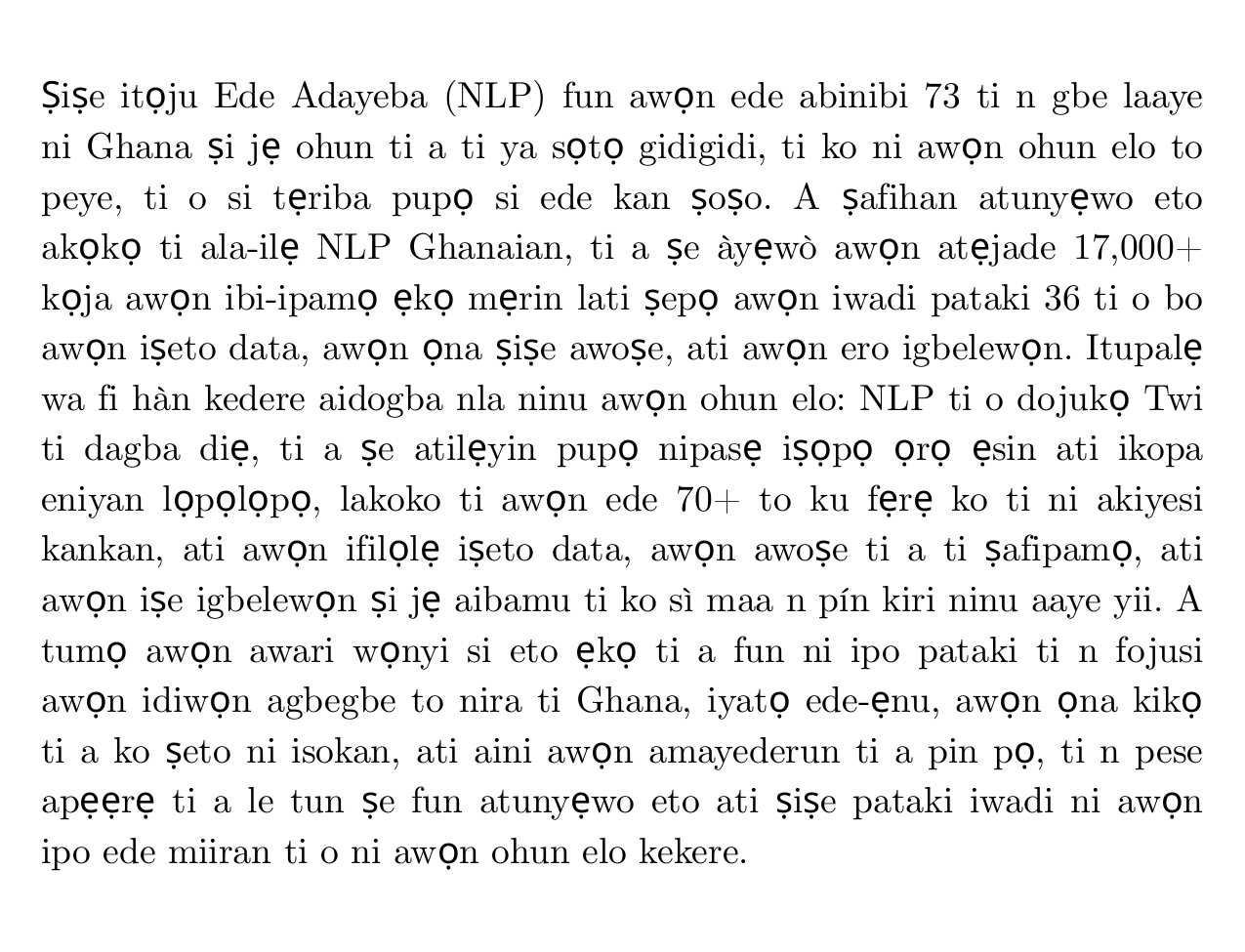}
    \caption{Yoruba Translation}
\end{figure}

\begin{figure}[htbp]
    \centering
    \includegraphics[width=\linewidth]{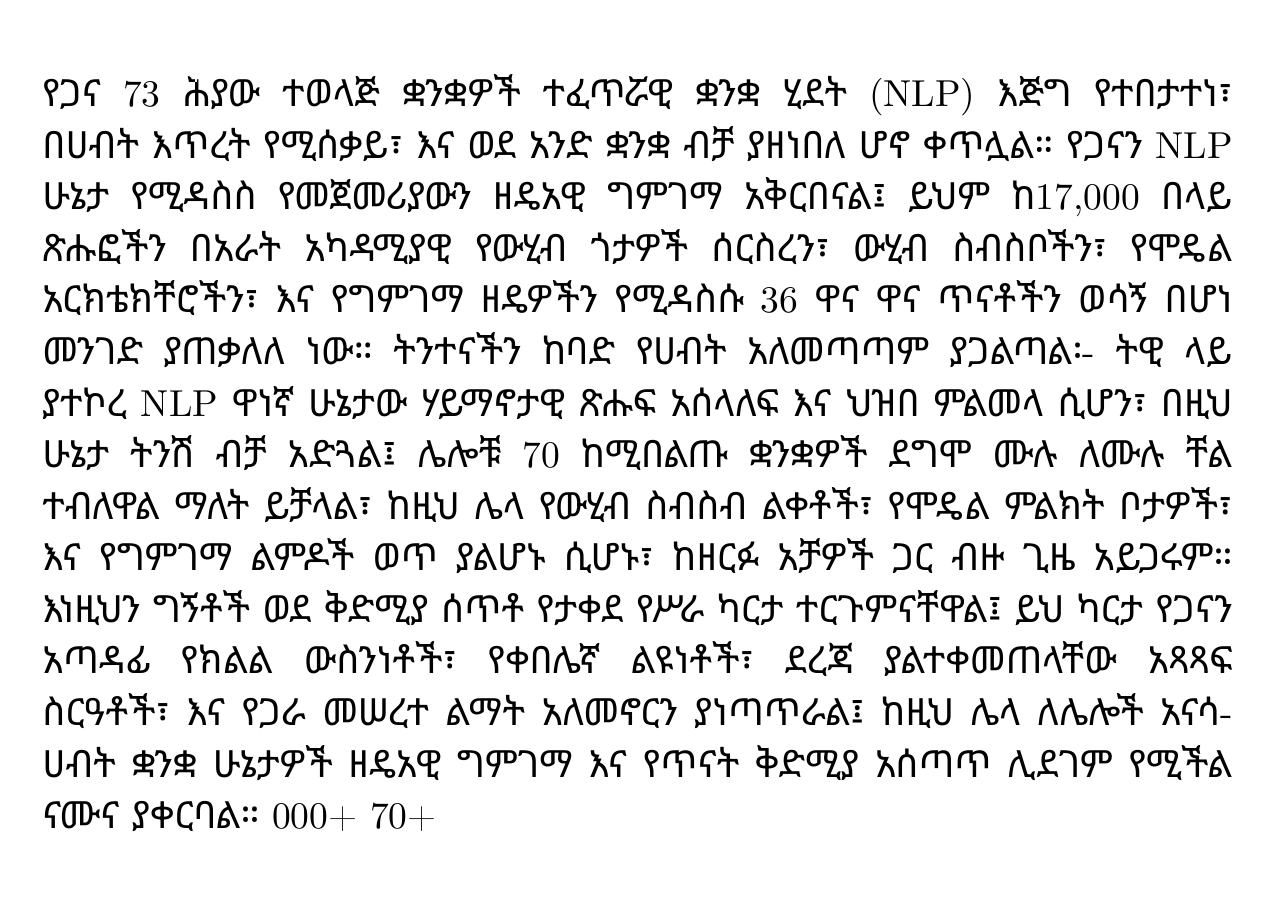}
    \caption{Amharic Translation}
\end{figure}

\begin{figure}[htbp]
    \centering
    \includegraphics[width=\linewidth]{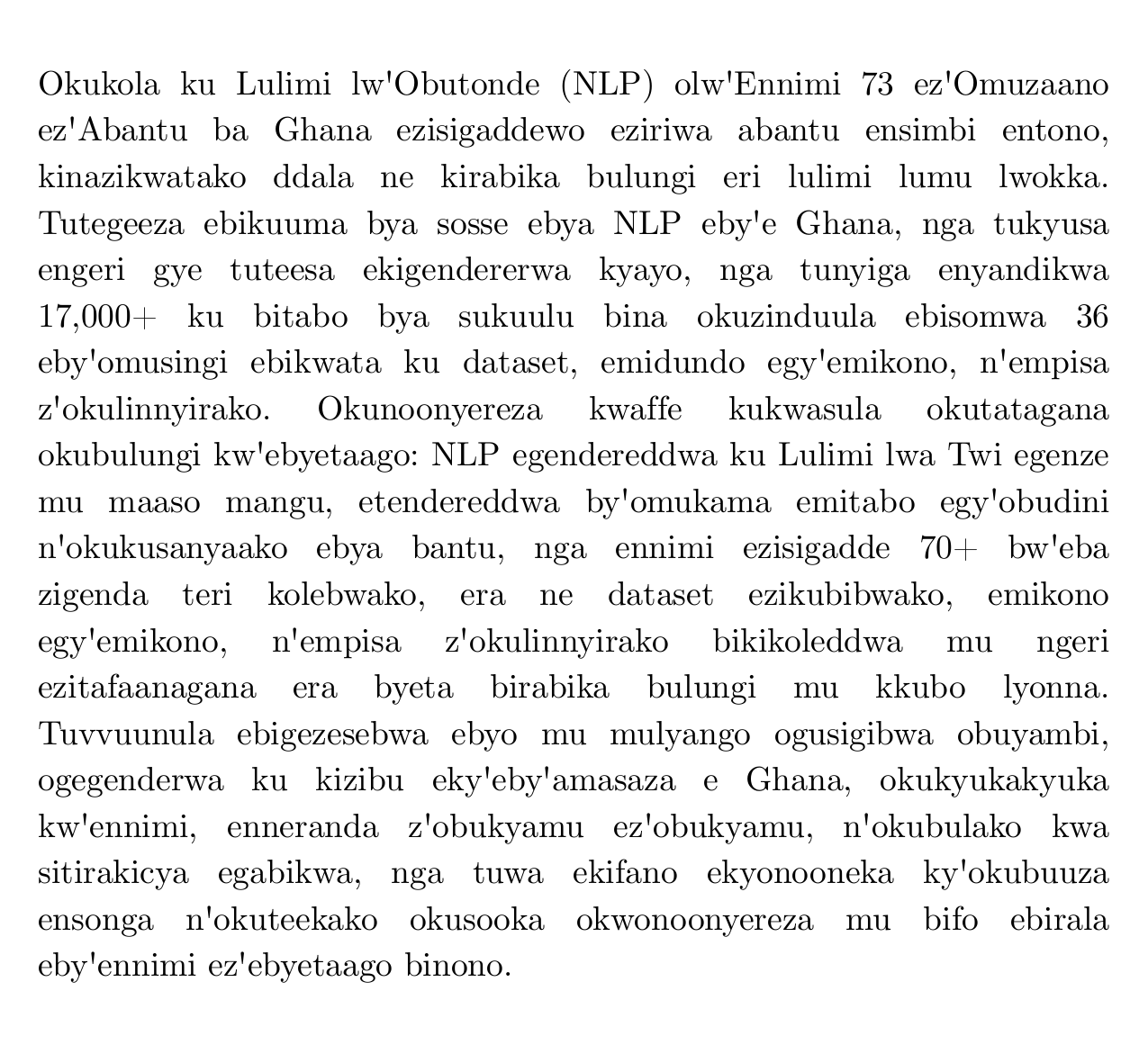}
    \caption{Luganda Translation}
\end{figure}

\begin{figure}[htbp]
    \centering
    \includegraphics[width=\linewidth]{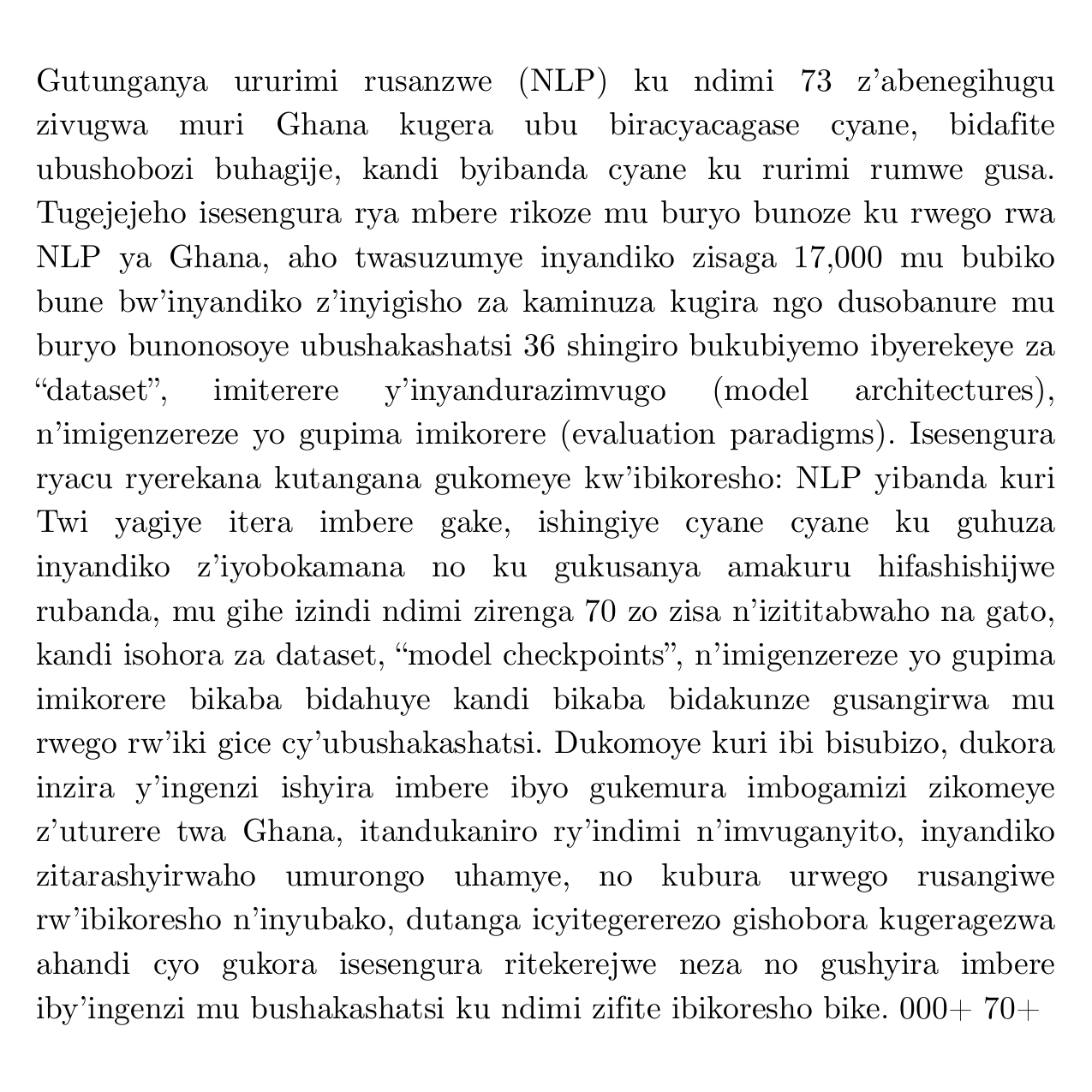}
    \caption{Kinyarwanda Translation}
\end{figure}

\begin{figure}[htbp]
    \centering
    \includegraphics[width=\linewidth]{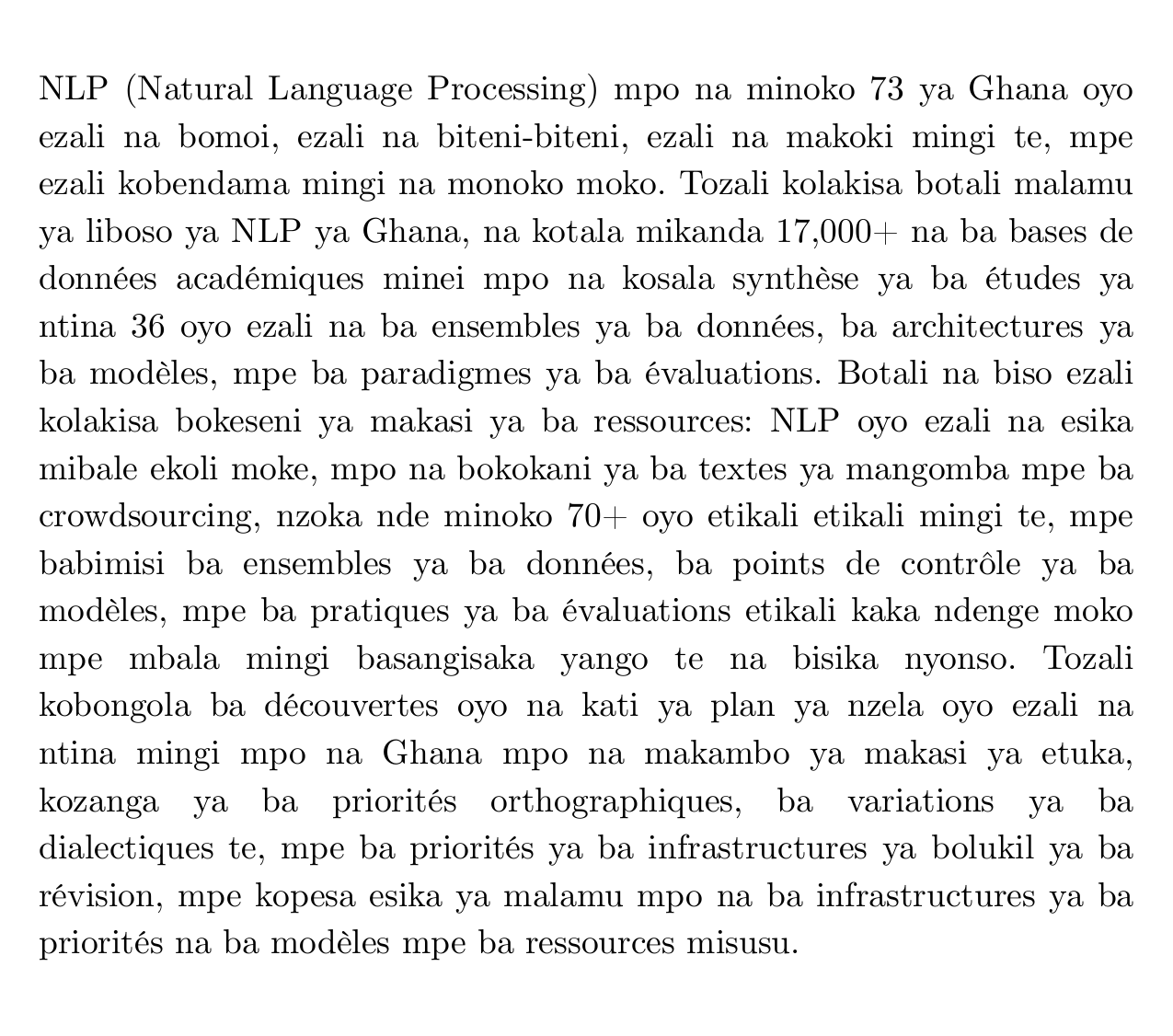}
    \caption{Lingala Translation}
\end{figure}

\begin{figure}[htbp]
    \centering
    \includegraphics[width=\linewidth]{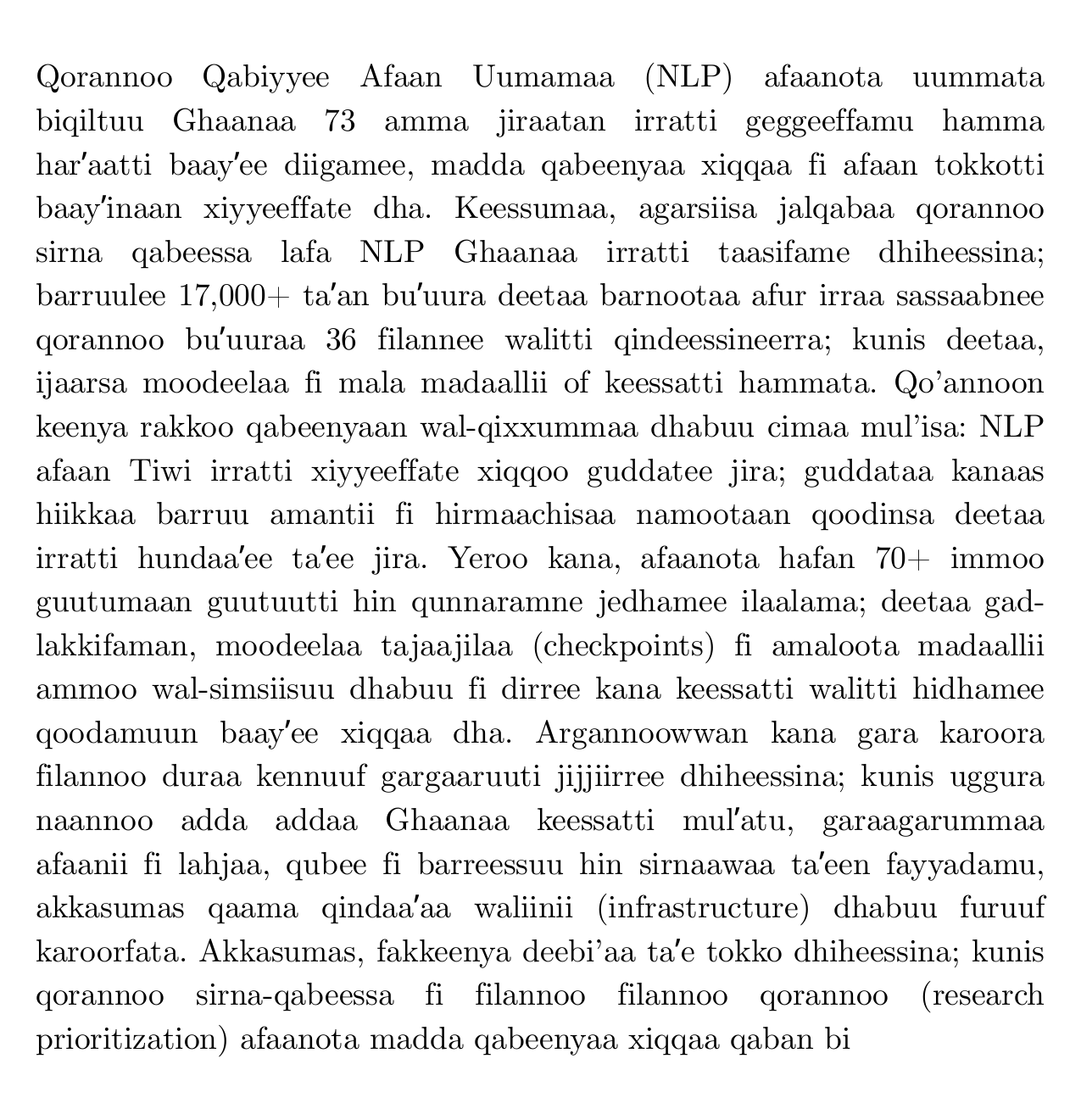}
    \caption{Oromo Translation}
\end{figure}

\begin{figure}[htbp]
    \centering
    \includegraphics[width=\linewidth]{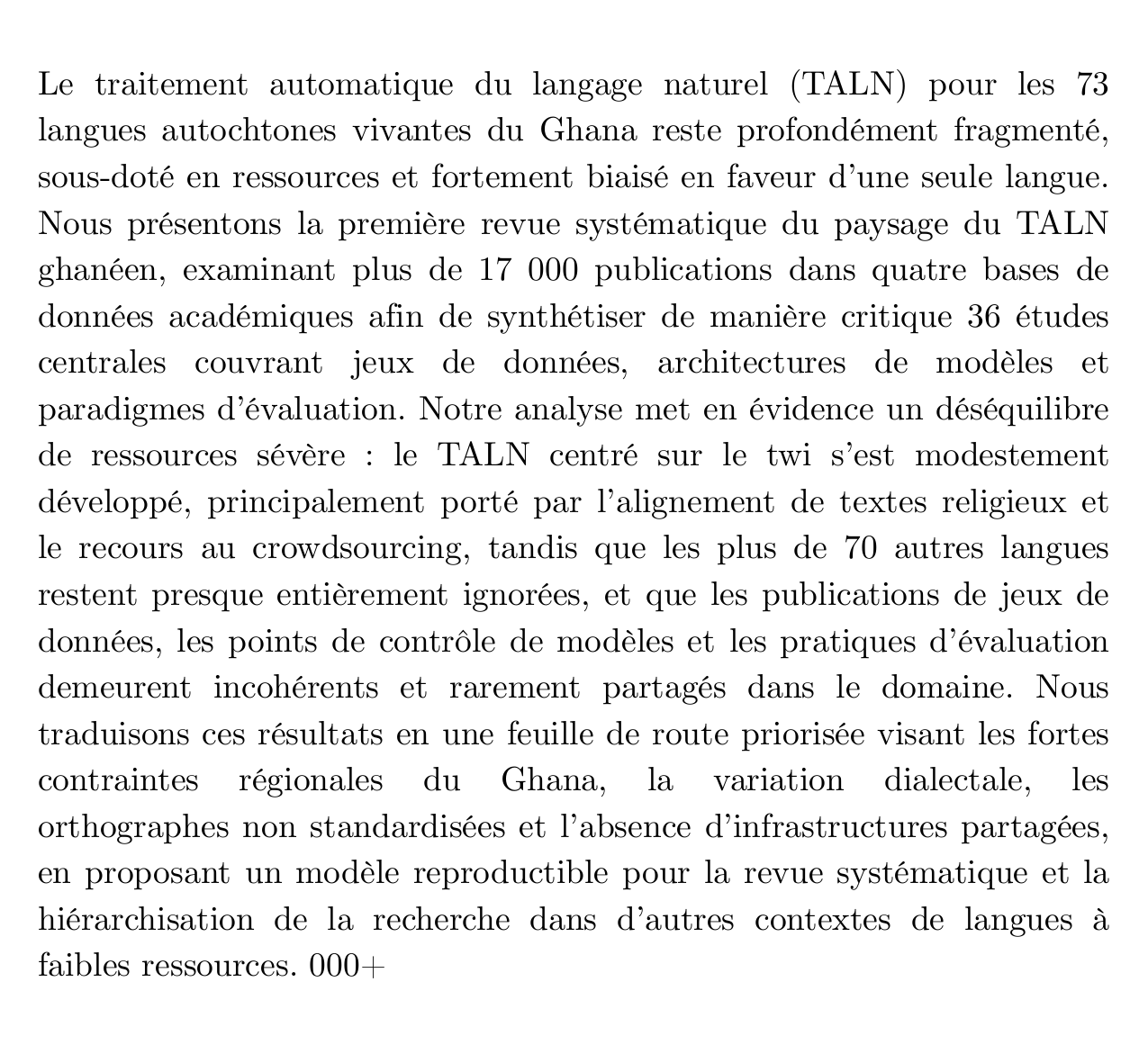}
    \caption{French Translation}
\end{figure}

\begin{figure}[htbp]
    \centering
    \includegraphics[width=\linewidth]{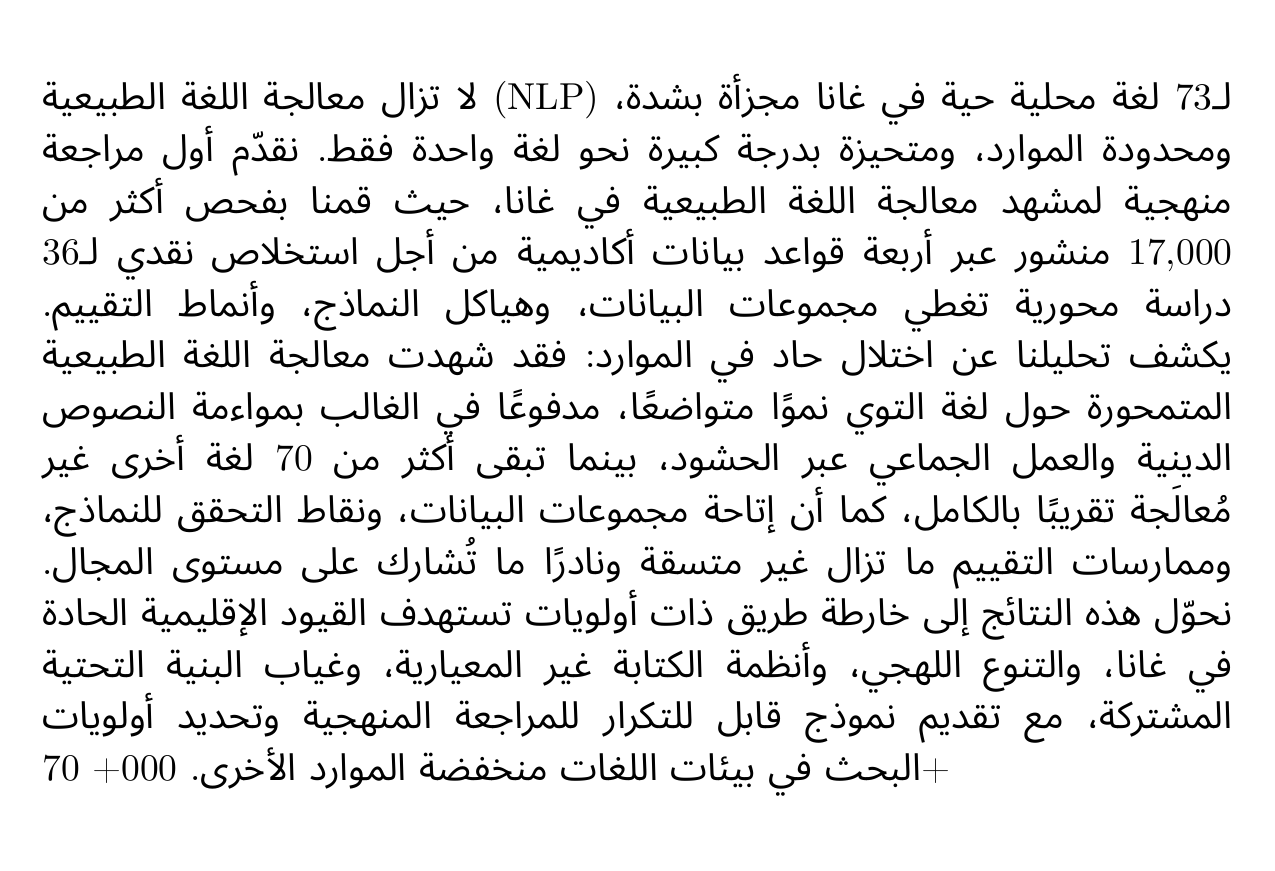}
    \caption{Arabic Translation}
\end{figure}

\begin{figure}[htbp]
    \centering
    \includegraphics[width=\linewidth]{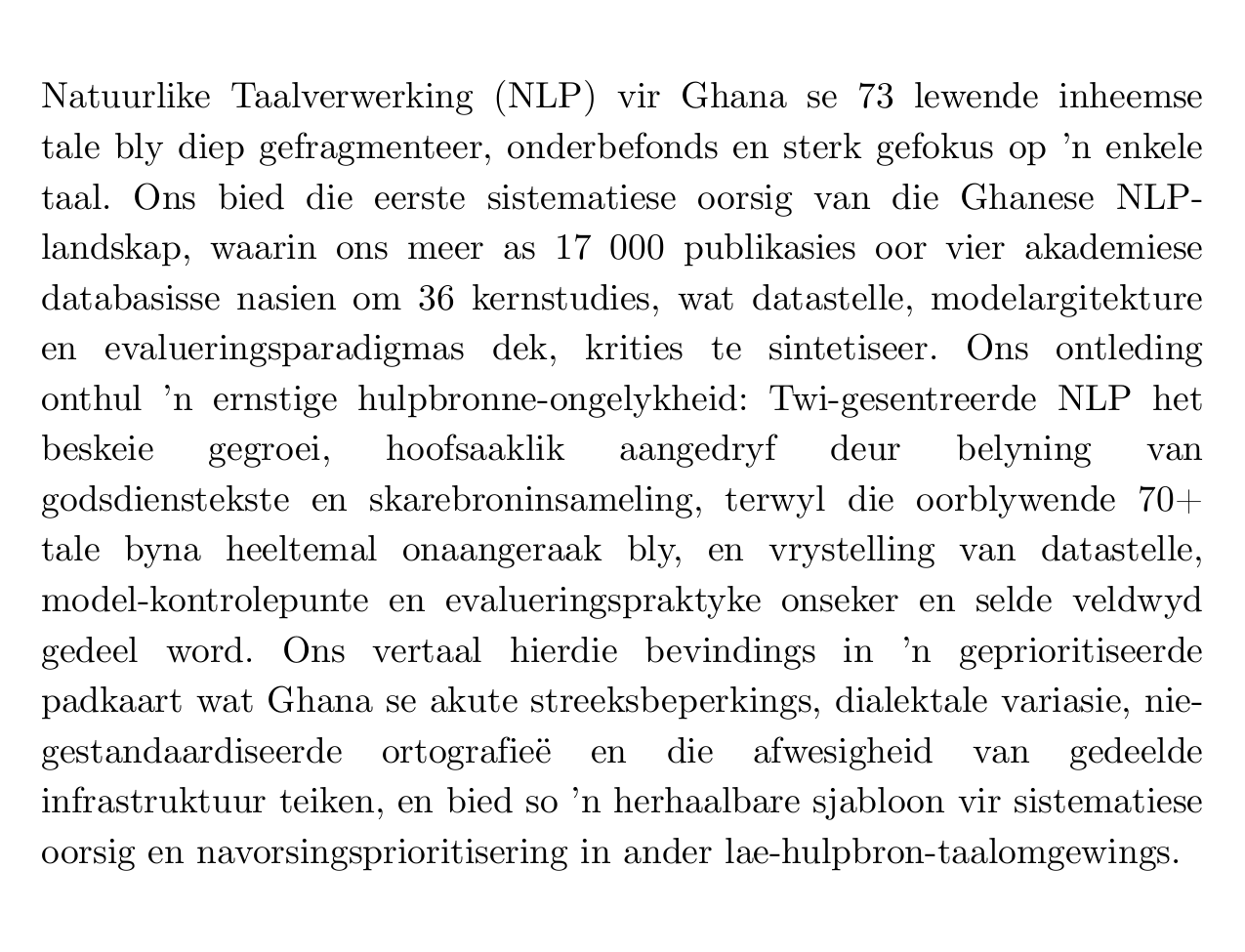}
    \caption{Afrikaans Translation}
\end{figure}

\begin{figure}[htbp]
    \centering
    \includegraphics[width=\linewidth]{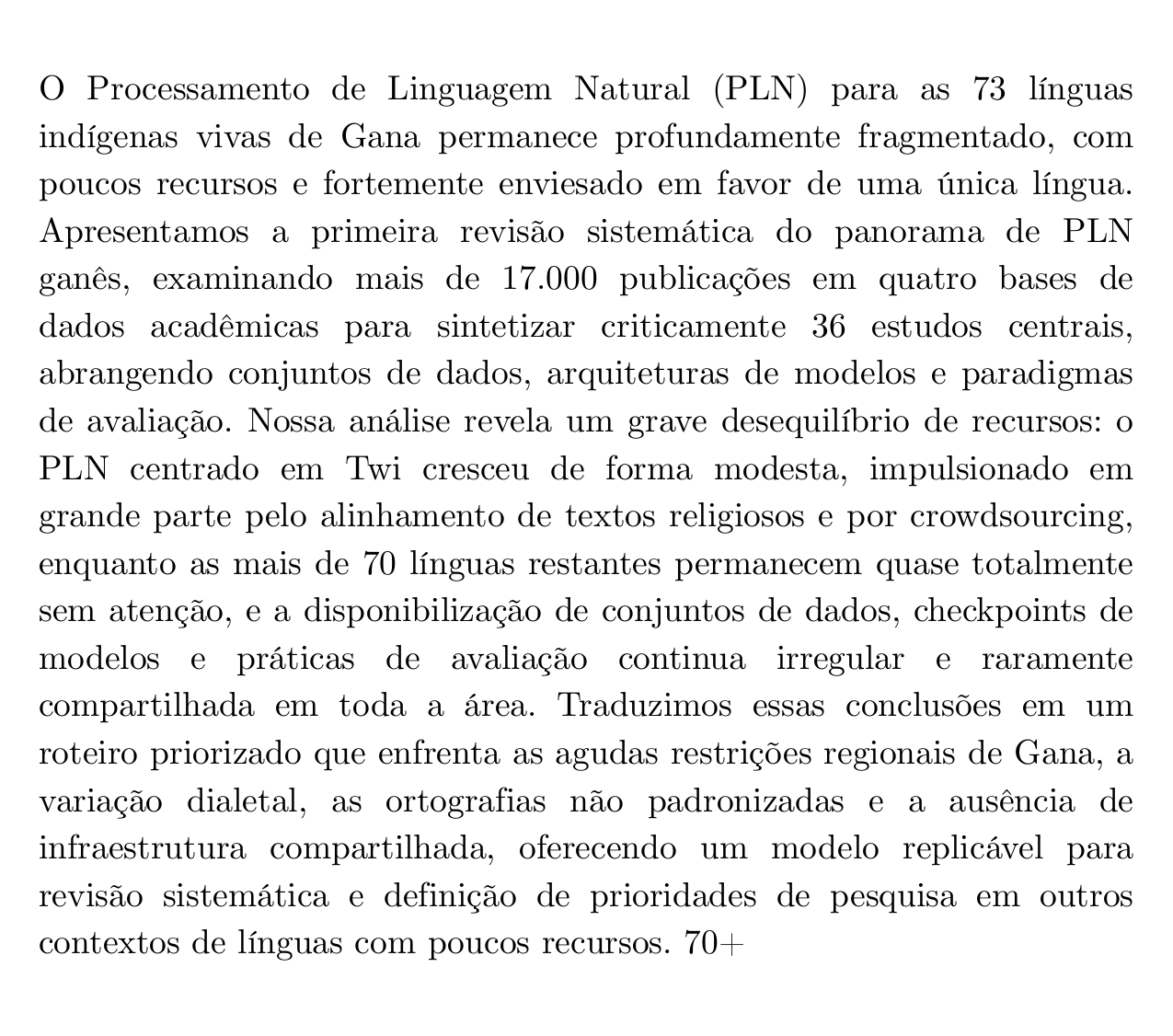}
    \caption{Portuguese Translation}
\end{figure}

\begin{figure}[htbp]
    \centering
    \includegraphics[width=\linewidth]{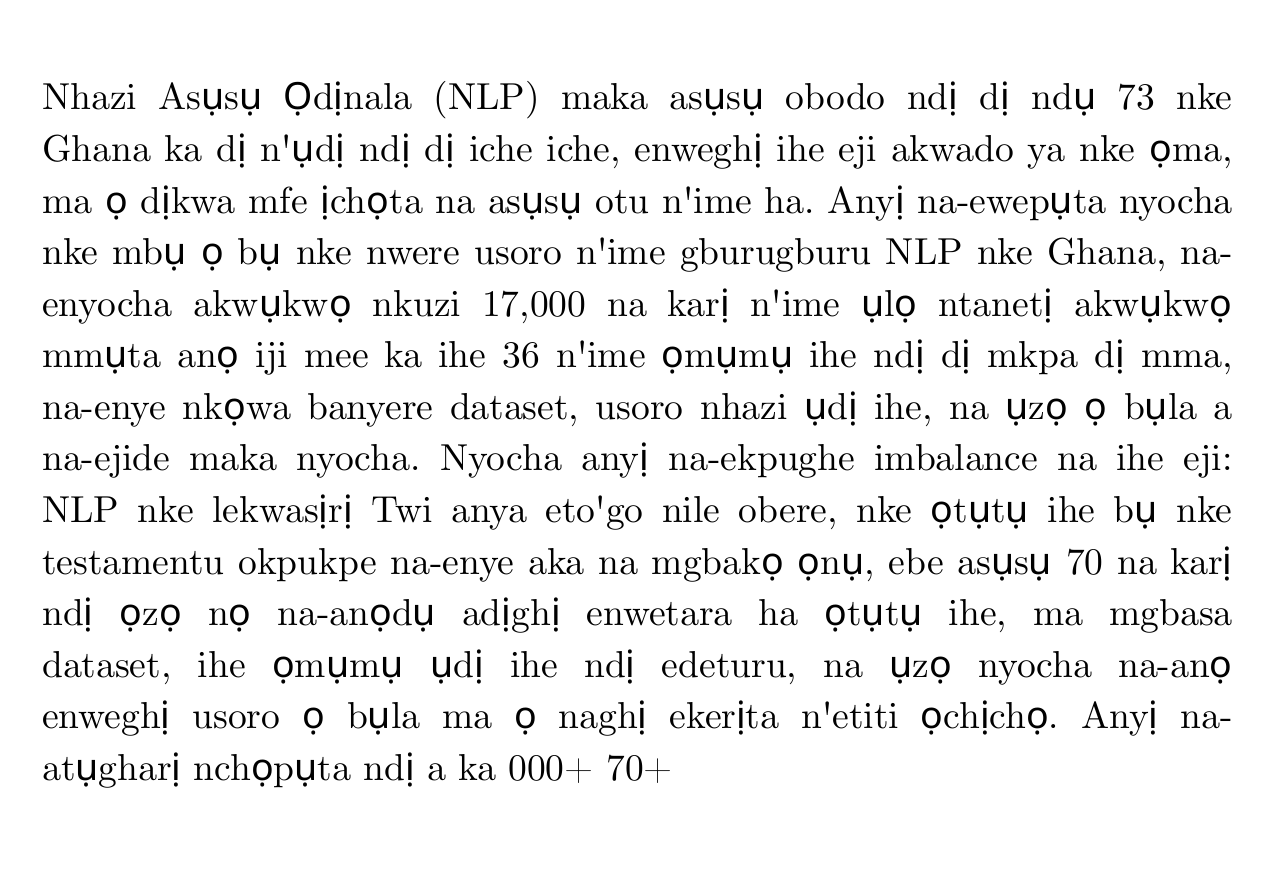}
    \caption{Igbo Translation}
\end{figure}

\begin{figure}[htbp]
    \centering
    \includegraphics[width=\linewidth]{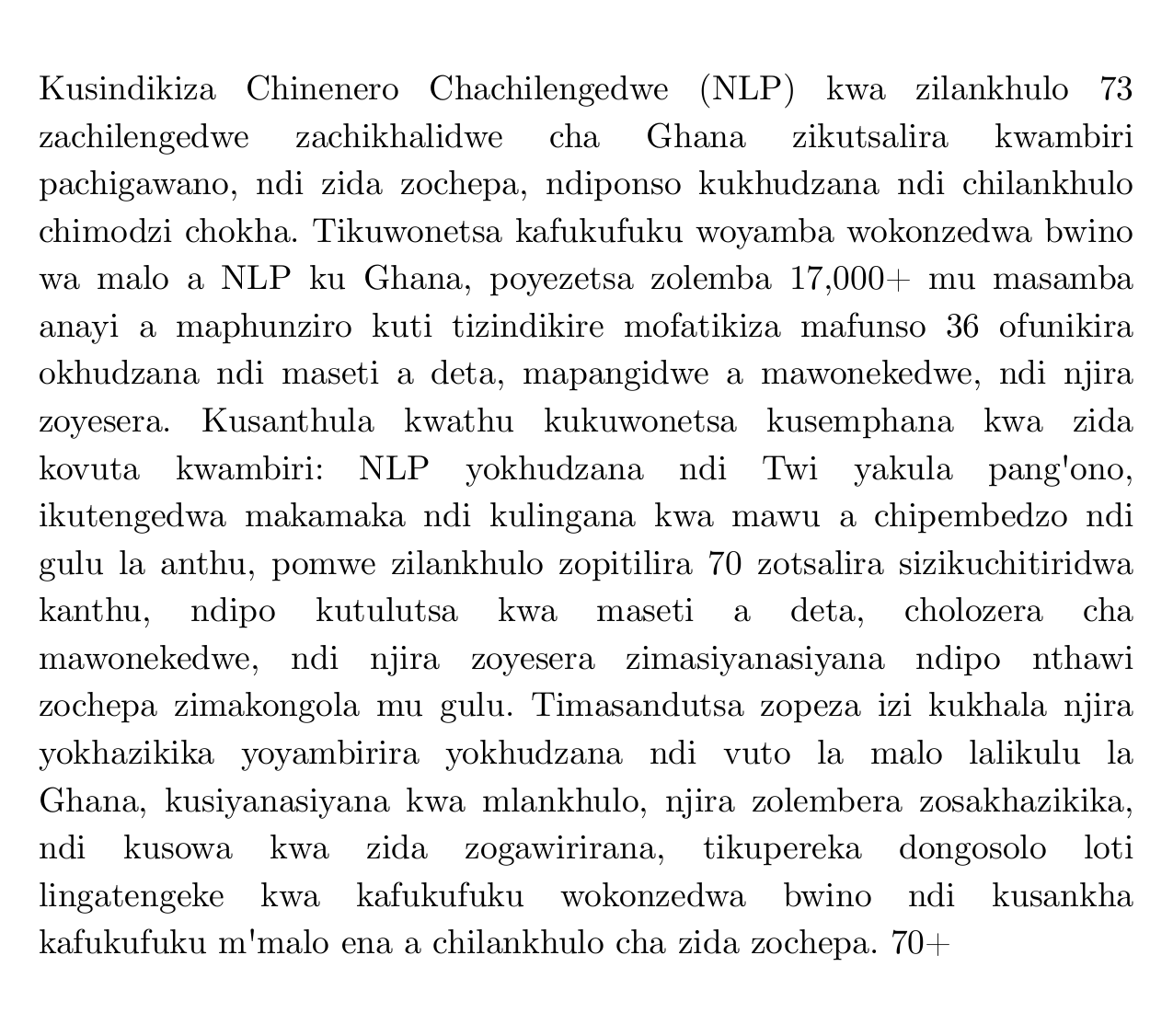}
    \caption{Chichewa Translation}
\end{figure}

\end{document}